\documentclass[pdflatex,sn-mathphys-num]{sn-jnl}%

\usepackage{rotating}
\usepackage{graphicx}%
\usepackage{multirow}%
\usepackage{amsmath,amssymb,amsfonts}%
\usepackage{amsthm}%
\usepackage{mathrsfs}%
\usepackage[title]{appendix}%
\usepackage{xcolor}%
\usepackage{textcomp}%
\usepackage{manyfoot}%
\usepackage{booktabs}%
\usepackage{algorithm}%
\usepackage{algorithmicx}%
\usepackage{algpseudocode}%
\usepackage{listings}%

\usepackage{latexsym}
\usepackage{booktabs}
\usepackage{enumitem}
\usepackage{color}
\let\orcidlogo\relax
\usepackage{orcidlink}

\newcommand{\E}{\mathrm{E}}

\DeclareMathOperator{\sign}{sign}

\newcommand{\calX}{\mathcal{X}}
\newcommand{\calY}{\mathcal{Y}}
\newcommand{\calW}{\mathcal{W}}
\renewcommand{\P}{\mathbb{P}}

\newcommand{\chartscale}{0.32}

\theoremstyle{thmstyleone}%

\theoremstyle{thmstyletwo}%

\theoremstyle{thmstylethree}%

\begin{document}

\title[Class flipping for uplift modeling and Heterogeneous Treatment Effect estimation on imbalanced RCT data]{Class flipping for uplift modeling and Heterogeneous Treatment Effect estimation on imbalanced RCT data}

\author*[1,2]{\fnm{Krzysztof} \sur{Ruda\'{s}}~\orcidlink{0000-0002-8309-9952}}\email{krzysztof.rudas@ipipan.waw.pl}
\equalcont{These authors contributed equally to this work.}

\author[1,2]{\fnm{Szymon} \sur{Jaroszewicz}~\orcidlink{0000-0001-9327-5019}}\email{szymon.jaroszewicz@ipipan.waw.pl}
\equalcont{These authors contributed equally to this work.}

\affil*[1]{\orgdiv{Institute of Computer Science}, \orgname{Polish Academy of Sciences}, \orgaddress{\city{Warsaw}, \country{Poland}}}

\affil[2]{\orgdiv{Faculty of Mathematics and Information Science}, \orgname{Warsaw University of Technology}, \orgaddress{\city{Warsaw}, \country{Poland}}}

\abstract{Uplift modeling and Heterogeneous Treatment Effect (HTE)
  estimation aim to predict the causal effect of an action, such as a
  medical treatment or a marketing campaign on a specific
  individual. In this paper, we focus on data from Randomized
  Controlled Experiments which guarantee a causal interpretation of
  the outcomes. Class imbalance is an important problem in uplift
  modeling/HTE because it occurs in many domains where it is applied,
  such as online advertising.  Unfortunately classical undersampling
  or oversampling based approaches are difficult to apply in this
  context because they distort the predicted effect. Calibration
  methods have been proposed in the past, however, they do not
  guarantee correct predictions. In this work, we propose an approach
  alternative to undersampling, based on flipping the class value of
  selected records. We show that the proposed approach does not
  distort the predicted effect and does not require calibration. The
  method is especially useful for models based on class variable
  transformation (modified outcome models). We address those models
  separately, designing a transformation scheme which guarantees
  correct predictions and simultaneously resolves the problem of
  spurious correlations with treatment variable which degrade
  performance of those models. Experiments fully confirm our
  theoretical results. Additionally, we demonstrate that our method
  is a viable alternative also for other types of uplift models
  and for standard imbalanced classification problems.}

\keywords{class flipping, Heterogenous Treatment Effect, uplift modeling, class imbalance}

\maketitle

\section{Introduction}\label{sec:intro}
Selecting observations that should be targets of an action such as a
marketing campaign or a medical treatment is becoming an important
problem in machine learning. Non-causal approaches focus on predicting
responses within a treated group of observations, e.g., customers who
received an offer. A new observation is deemed suitable for
action if the predicted outcome is positive (i.e., the customer made a
purchase). Unfortunately, such approaches ignore the responses which
would have been observed had the action not been taken.

Randomized Controlled Trials (RCTs) offer a solution to this problem.
In an RCT, the population is randomly divided into two groups: the
{\em treatment group} which receives the intervention, and the {\em
  control group}, which does not.
This allows us to construct a model which calculates the true effect of
an action on an individual described by a set of features.

There are two main lines of research on applying machine learning
models to experimental data.  The first is uplift modeling which is
most popular in business
applications~\citep{RadcliffeTechRep,Rzepakowski2011,olaya2020survey}
with the main focus on building best the possible models on RCT data.
Another line of research comes under the name Heterogeneous Treatment
Effect (HTE) estimation~\citep{athey2016recursive,Kunzel} and is mainly
concerned with nonrandomized experiments, but the methods proposed
there are also applicable to RCTs.  Learning from nonrandomized data
is attractive but relies on difficult to verify assumptions such as `no
unmeasured confounders'.  In practice, only random treatment
assignment guarantees that the true causal effect is correctly
identified.  The terms HTE estimation from RCT data and uplift
modeling will be used interchangeably in the remaining part of this
paper.

We address the problem of class imbalance in uplift modeling/HTE
estimation which has received very little attention in the literature.

\subsection{Notation}
Let us now introduce the notation used throughout the paper.  Scalars
and vectors will be denoted with lowercase letters $x,y$, and random
variables with script letters $\calX$, $\calY$.  Denote by $\P$ the
population distribution over $(\calX, \calY, \calW)$, where $\calX$ is
a $p$-dimensional vector of features, $\mathcal{Y}\in\{0,1\}$ a binary
response, and $\calW \in \{C,T\}$ is a random variable, which denotes
assignment to one of two groups: treatment ($T$) or control ($C$).
Using this notation we may formally define the {\em conditional
  average treatment effect} (CATE) also known as {\em uplift}
(i.e.~the net effect of an action on an individual described by
features $x$) as:
\begin{equation}\label{eq:cate}
\tau(x) = \mathbb{P}(\mathcal{Y}=1|x,\calW=T)-\mathbb{P}(\mathcal{Y}=1|x,\calW=C).
\end{equation}
Our goal is to build a model predicting $\tau(x)$ based on a training
set $\{(x_i, y_i, w_i)\}^n_{i=1}$ sampled at random from $\P$.

In this paper we assume the treatment assignment to be completely
random, so we are guaranteed that assumptions such as strong
ignorability~\citep{ImbensRubin} are satisfied and the model is causal
in nature~\citep{HernanRobbins}.  Moreover, we have $\calX\perp\calW$.
Note that without random assignment the definition of CATE shown above
would no longer have causal interpretation.

The paper is organized as follows. In the remaining part of
Section~\ref{sec:intro} we present existing works on uplift modeling
and HTE estimation, and position them in the wider context of causal
discovery. We also review works concerning class and treatment
imbalance problems in uplift modeling. In
Section~\ref{sec:imbal-classif} we introduce our class flipping method
and compare it to undersampling in the standard classification setting,
and in Section~\ref{sec:imbal-uplift} extend it to uplift
modeling/HTE estimation. In Section~\ref{sec:cvt} we adapt these
results to CVT models: an important class of uplift models very
sensitive to class imbalance.  In Section~\ref{sec:exper} we
experimentally compare class flipping and undersampling methods. In
Section~\ref{sec:conc} we conclude.
 
\subsection{Related Work}

Uplift modeling/HTE estimation are part of a broader problem of causal
discovery which uses experimental data where an intervention
has directly been applied to a subgroup of a population. This is the
scenario considered in this paper.

Most papers on uplift modeling concentrate on the classification
problem and most HTE estimation papers on regression.  Many early
methods were based on decision tree construction
algorithms~\citep{RadcliffeTechRep,Rzepakowski2011,athey2016recursive}
which use modified splitting criteria to take into account the
difference in responses between treatment groups. Later works extended
these results to ensemble
methods~\citep{GuelmanRandForest,Sol,athey-causal-forest}. Linear
models like logistic regression or Support Vector Machines have also
been
proposed~\citep{SVMDP14,my:upsvm,my:lp-upsvm,Imai,Chernozhukov14,RudasJaroszewicz,RudasJaroszewicz2,RudasJaroszewicz3}.
A learning to rank method for uplift modeling has been presented
in~\citep{Verbeke}.

A very important group of methods are so called
metamodels which adapt standard classifiers to uplift modeling
problem~\citep{Kunzel,Jas,Nie-quasi-oracle}.  The most popular metamodel is the Two
Model approach (called the T-learner in~\citep{Kunzel}) which builds
separate classifiers on the treatment and control groups and subtracts
their predictions.  Despite criticisms~\citep{RadcliffeTechRep} the
method often produces excellent results~\citep{RudasJaroszewicz}.

Another important uplift metamodel is the class variable
transformation (CVT)
approach~\citep{Lo2014,Jas,Lai,uplift-response-transf-revenue}.  The
method is based on reversing the class variable in the control group
and concatenating the resulting dataset with the treatment group.  It
can be shown~\citep{Jas} that any probabilistic classifier built on
such a dataset will predict uplift directly.  While appealing, the
method may have high variance~\citep{RudasJaroszewicz} and is very
susceptible to class imbalance~\citep{Nyberg}.  The CVT method will
thus be addressed separately (Section~\ref{sec:cvt-flip})
and in Section~\ref{sec:exper} we will confirm the results
of~\citep{NybergDMKD} which suggest that the method becomes highly
competitive if class imbalance is corrected.

A more complete overview of uplift modeling literature can be found
e.g.~in~\citep{DevriendtSurvey} and an overview of HTE literature
in~\citep{Lipkovich24}.  We now proceed to discuss literature on
imbalanced uplift modeling.

\paragraph{Class and treatment imbalance problems.}

Consider a classification problem with two classes $0$ and $1$.  The
class imbalance problem occurs when one of the classes (typically
class $1$) is significantly less frequent than the other.  The problem
has received significant attention in machine learning literature,
see~\citep{kaur2019systematic,Bontempi1,Bontempi2,Stefanowski} for a thorough overview.  The most
popular approach is undersampling the majority class such that the
imbalance is decreased.  Equivalently, one may assign lower weights to
records in the majority class which avoids data loss (an approach we
use in this paper).

Very little work has been devoted to class imbalance in uplift
modeling/HTE estimation.  In uplift modeling the main research focus
has been on treatment imbalance, where one of the treatments is
underrepresented; much less attention has been paid to the class
imbalance problem.  We begin with the review of literature for
treatment imbalance problem.  In~\citep{Betlei} the authors propose two
metamodels for uplift modeling and suggest their use in treatment
imbalanced scenarios.
Olaya et al.~\citep{olaya2020survey} discuss the
effect of imbalance in treatment group sizes on existing uplift
modeling methods.

In~\citep{ImbalanceAwareUplift} the authors combine two estimators
based on target variable transformations, such that the resulting
estimator is less affected by treatment imbalance (the authors also
discuss the case of non-randomized treatment assignment).
In~\citep{DeepEntireNetworkITE} a concept of Entire Space Neural
Networks was introduced which address the problem of treatment
imbalance by learning models for each treatment on the entire dataset.
In~\citep{vairetti} a method based on the SMOTE~\citep{SMOTE} algorithm
has been combined with propensity score matching to address the
problem of class imbalance in non-randomized experiments.  The paper
does not address the problems we describe in
Section~\ref{sec:imbal-uplift}, so the predicted treatment effects may be
distorted.

Most relevant to the current paper are works by Nyberg et
al.~\citep{Nyberg,NybergDMKD}, which are the only works we are aware of
addressing the problem of class imbalance in uplift modeling/HTE
estimation.  The authors were the first to notice an important problem
undersampling based methods encounter when applied to uplift modeling.

More specifically, when undersampling is applied, the original class
probabilities cannot easily be recovered.  When applied to
uplift modeling, treatment and control probabilities
become inconsistent and the final uplift estimate will not be
correct (more details are given in Section~\ref{sec:imbal-uplift}).
In~\citep{Nyberg,NybergDMKD} the authors note that in the special case
of $\mathbb{P}(\calY=1)\approx 0$ undersampling can be applied without the
problem occurring.  In other cases they suggest using cross-validation
to select undersampling factors which balance the benefits of
imbalance correction against the bias in estimating CATE.

In this work we propose an alternative class imbalance correction
method based on class flipping which can be applied to uplift modeling
directly.

\section{Class flipping for standard classification}\label{sec:imbal-classif}

In this section we introduce the main contribution of this paper which
is to correct class imbalance by flipping class values in the majority
class.  We will begin by showing this concept for ordinary
classifiers.  We will thus ignore the treatment indicator $\calW$.
Moreover, it will be more convenient to fix the predictors vector at
$x$ and work with the conditional distribution
$\P(\mathcal{Y}|\calX=x)$.

\subsection{Majority class undersampling}
Under and oversampling are the most widely used methods for correcting
the class imbalance in the classification problem~\citep{kaur2019systematic}.  Most
commonly, cases in the majority are undersampled such that the
disproportion between classes decreases.

A drawback of this method is that the relationship between the
proportion of observations removed and the predicted class
probabilities is not straightforward. For example, assume that the
majority class is $0$ (a typical case) and we randomly remove
$(1-k)\cdot 100\%$ of observations from that class, where $0\leq k\leq 1$ is
the undersampling factor (in other words, $k$ is the proportion of
majority class records kept).  Denote by $\mathbb{P}^*$ the
distribution which would be obtained from the original distribution
$\P$ if undersampling was applied at the population level.

In the remaining part of the paper we will follow the convention that
distributions obtained from $\P$ through undersampling/record
weighting will be denoted with $\P^*$.

Class probabilities before and after undersampling are related by the
following formula~\citep{NybergDMKD,Bontempi1,Bontempi2}:
\begin{align}\label{eq:under}
\mathbb{P}^*(\mathcal{Y}=0|x)=\frac{k\mathbb{P}(\mathcal{Y}=0|x)}{\mathbb{P}(\mathcal{Y}=1|x)+k\mathbb{P}(\mathcal{Y}=0|x)}.
\end{align}
Notice that the relationship is not linear and depends on the
(unknown) conditional class probabilities.

Instead of actually undersampling the data one can assign weights $q_i$ to data records with
\begin{equation}\label{eq:under-weights}
q_i=
\begin{cases}
  1 & \mbox{if }y_i=1,\cr
  k & \mbox{if }y_i=0.
\end{cases}
\end{equation}
Weighting typically gives slightly better results, especially when data is scarce.

\subsection{Class flipping}
As an alternative to undersampling, we propose a class flipping based
method where, instead of removing samples, their classes are reversed.
Suppose that we randomly select $(1-k)\cdot 100\%$ observations
from class $0$ and change their response to $1$.  At the population
level this results in a modified response variable:
\begin{align*}
& \breve{\mathcal{Y}}=
\begin{cases}
0 \mbox{ if } \mathcal{Y}=0 \mbox{ and } \mathcal{U}\leqslant k,\\
1 \mbox{ if } \mathcal{Y}=0 \mbox{ and } \mathcal{U}>k,\\
1 \mbox{ if } \mathcal{Y}=1,
\end{cases}
\end{align*}
where $\mathcal{U}$ is a random variable uniformly distributed on the
interval $[0,1]$ independent of $\mathcal{Y}$ and $\calX$. Then:
\begin{align}
\mathbb{P}(\breve{\mathcal{Y}}=0|x)& =\mathbb{P}(\mathcal{Y}=0,\mathcal{U}\leqslant k|x)=\mathbb{P}(\mathcal{Y}=0|x)\mathbb{P}(\mathcal{U}\leqslant k)\nonumber\\ 
& = k\mathbb{P}(\mathcal{Y}=0|x).\label{eq:row1}
\end{align}
We can see that after flipping, the probability of class $0$ changes
by the factor $k$ regardless of its original value so, unlike in the
case of undersampling, we have a {\em simple linear relationship}
between the proportion of flipped observations and the change in the
probability of class 0. The relationship holds for both conditional
and unconditional probabilities.

It can thus be seen that flipping offers an alternative to
undersampling for handling class imbalance, with the advantage of
easier recovery of original probabilities (Equation~\ref{eq:row1}
vs.~Equation~\ref{eq:under}).  In the next section we will see that the
advantage becomes much more important in the case of uplift
modeling/CATE estimation, where it allows for achieving model
identifiability.

\paragraph{Additional notation.}

In the following sections we will need to flip both classes ($0$ and
$1$) and we will now introduce such possibility.  The two classes will
be flipped using coefficients $k_{0}$ and $k_{1}$ respectively, and
$\breve{\mathcal{Y}}$ becomes
\begin{align}
& \breve{\mathcal{Y}}=
\begin{cases}
0 \mbox{ if } \mathcal{Y}=0\mbox{ and } \mathcal{U}\leqslant k_{0},\\
1 \mbox{ if } \mathcal{Y}=0\mbox{ and } \mathcal{U}>k_{0},\\
0 \mbox{ if } \mathcal{Y}=1 \mbox{ and } \mathcal{U}>k_{1},\\
1 \mbox{ if } \mathcal{Y}=1 \mbox{ and } \mathcal{U}\leqslant k_{1},
\end{cases}\label{eq:breve-y}
\end{align}
where $k_{0},k_1\in[0,1]$.  Additionally we require that at least one
of $k_{0}$, $k_1$ be equal to one, i.e.~we flip cases in only one of
the classes.\footnote{In full generality the method could
simultaneously flip classes $0$ and $1$ in different proportions, however
it is easy to see that the desired relationship in
Equation~\ref{eq:row1} would no longer hold.}  In the remaining part
of the article we will clearly state which class is flipped,
i.e.~whether $k_0<1$ or $k_1<1$.

\subsubsection{Deterministic version of the transform}\label{sec:flip-weighted}
Recall (Equation~\ref{eq:under-weights}) that the effects of
undersampling can be achieved by record weighting.  We now show that
the effect of random class flips can be also be obtained by
appropriate weighting and duplicating of records, which makes the
procedure fully deterministic.

Assume that we are flipping class $0$ with factor $k_0$.  Take the
following conditional expectation of some function
$f(x,\breve{\mathcal{Y}})$ (e.g.~a loss function):
\begin{align}
\E_{\breve{\mathcal{Y}}}[f(x,\breve{\mathcal{Y}})|x]
& = f(x,1)\mathbb{P}(\breve{\mathcal{Y}}=1|x)+f(x,0)\mathbb{P}(\breve{\mathcal{Y}}=0|x)\nonumber\\
&=f(x,1)(1-k_0\mathbb{P}(\mathcal{Y}=0|x))+f(x,0)k_0\mathbb{P}(\mathcal{Y}=0|x).\label{eq:flip-deterministic}
\end{align}
The equation suggests the following scheme: each class $0$ record
$(x_i,0)$ is replaced with two records $(x_i,0)$ and $(x_i,1)$ with
weights respectively $k_0$ and $1-k_0$.  Records in class $1$ are
unchanged and all get weights equal to $1$.

Let us illustrate  the procedure on a simple example with two data
records:
\begin{align*}
\begin{array}{l}
(x_1, y=1) \\
(x_2, y=0) 
\end{array}\quad\mbox{become}\quad\begin{array}{l}
(x_1, y=1),\, q_1=1 \qquad\\
(x_2, y=0),\, q_2=k_0 \qquad\\
(x_2, y=1),\, q_2'=1-k_0, 
\end{array}
\end{align*} 
where $q_i$'s denote record weights.

Notice that the change in probability of $\mathcal{Y}$ given in
Equation~\ref{eq:row1} will hold even on the sample level.  This is
easy to see since every record with class zero will now receive a
weight $k_0$ and, due to adding duplicate records with class 1 and
weight $1-k_0$, the total sum of weights in the database will not
change.  As a result, for every $x$, the probability
$\mathbb{P}(\breve{\mathcal{Y}}=0|x)$ in the weighted sample will be
equal to $k_0$ times the probability $\mathbb{P}({\mathcal{Y}}=0|x)$ in
the original sample.

\section{Class flipping for uplift modeling/CATE estimation}\label{sec:imbal-uplift}

Now we return to the main setting of the paper (see
Section~\ref{sec:intro}), where the population is randomly divided
into treatment and control groups.  Recall that the random variable
$\calW\in\{C,T\}$ denotes the group an individual was assigned to and
our aim is estimating uplift or CATE as defined in
Equation~\ref{eq:cate}.  We will again fix the predictors
vector at $x$, and work with triples $(x,\calY,\calW)$, where
$(\calY,\calW)\sim\P(\mathcal{Y},\calW|x)$.

Dealing with class imbalance in uplift modeling is more complicated
than in standard classification since the imbalance may need to be
corrected differently in the treatment and control groups,
i.e.~coefficients used in Equation~\ref{eq:breve-y} will differ across
groups.  We denote coefficients used for controls with $k_0^C$, $k_1^C$,
and for treatment cases with $k_0^T$, $k_1^T$.

Let us first notice that undersampling generally results in a loss of
model identifiability\footnote{i.e.~CATE will not be predicted
correctly even when sample size tends to infinity}.  Indeed, if class $0$
is undersampled, applying Equation~\ref{eq:under} to the definition of
CATE (reusing coefficient naming introduced for flipping) we obtain
\begin{align*}
\mathbb{P}^*&(\mathcal{Y}=1|x,\calW=T)-\mathbb{P}^*(\mathcal{Y}=1|x,\calW=C)\\
&=\frac{\mathbb{P}(\mathcal{Y}=1|x,\calW=T)}{\mathbb{P}(\mathcal{Y}=1|x,\calW=T)+k^T_0\mathbb{P}(\mathcal{Y}=0|x,\calW=T)}\\
&\quad-\frac{\mathbb{P}(\mathcal{Y}=1|x,\calW=C)}{\mathbb{P}(\mathcal{Y}=1|x,\calW=C)+k^C_0\mathbb{P}(\mathcal{Y}=0|x,\calW=C)},
\end{align*}
from which the original CATE, $\tau(x)$ cannot be recovered even when
$k^T_0=k^C_0$ since it depends on the unknown response probabilities.
More details can be found in~\citep{NybergDMKD}, where it is also shown
that the estimation becomes correct only for $\P(\calY=1|x)\approx 0$.

We will now demonstrate that class flipping is free from the above
problem and true uplift/CATE, $\tau(x)$ can be recovered from
predictions of an uplift model trained on the flipped class variable
$\breve{\calY}$.  Instead of the original $\tau(x)$ such a model
predicts $\breve{\tau}(x)$ given by
\begin{equation}\label{eq:breve-tau}
   \breve{\tau}(x) = \mathbb{P}(\breve{\mathcal{Y}}=1|x,\calW=T)-\mathbb{P}(\breve{\mathcal{Y}}=1|x,\calW=C).
\end{equation}
The first case we consider is when class $0$ is the majority class in
both groups.  Assume $k^C_1=k^T_1=1$, $k^C_0=k^T_0=k<1$, that is we
flip randomly cases from class $0$ into class $1$ in both groups with
equal probabilities.  Substituting Equation~\ref{eq:row1} into
Equation~\ref{eq:breve-tau} we get
\begin{align}
 \breve{\tau}(x)&  = 1-\mathbb{P}(\breve{\mathcal{Y}}=0|x,\calW=T)-1+\mathbb{P}(\breve{\mathcal{Y}}=0|x,\calW=C)\nonumber\\
& = k^C_0\mathbb{P}(\mathcal{Y}=0|x,\calW=C)-k^T_0\mathbb{P}(\mathcal{Y}=0|x,\calW=T)\nonumber\\
& = k\left(1-\mathbb{P}(\mathcal{Y}=1|x,\calW=C)-1+\mathbb{P}(\mathcal{Y}=1|x,\calW=T)\right)\nonumber\\
&=k\tau(x).\label{eq:imbal-0-0}
\end{align}
The correct $\tau(x)$ can be recovered after multiplying
$\breve{\tau}$ by $\frac{1}{k}$.  An analogous result easily follows
for the case when $1$ is the majority class in both groups and
$k^C_0=k^T_0=1$, $k^C_1=k^T_1=k<1$.

For completeness let us consider an unusual situation with different
majority classes in treatment and control groups.  Without loss of
generality consider the case when we flip class $1$ observations in
the treatment group and class $0$ in controls, i.e.~$k^T_0=k^C_1=1$
and $k^T_1=k^C_0=k<1$.  We have
\begin{align}
\breve{\tau}(x) &= \mathbb{P}(\breve{\mathcal{Y}}=1|x,\calW=T)-\mathbb{P}(\breve{\mathcal{Y}}=1|x,\calW=C) \nonumber\\
& = \mathbb{P}(\breve{\mathcal{Y}}=1|x,\calW=T)-1+\mathbb{P}(\breve{\mathcal{Y}}=0|x,\calW=C)\nonumber\\
& = k^T_1\mathbb{P}(\mathcal{Y}=1|x,\calW=T)-1+k^C_0\mathbb{P}(\mathcal{Y}=0|x,\calW=C))\nonumber\\
& = k\mathbb{P}(\mathcal{Y}=1|x,\calW=T)-1+k-k\mathbb{P}(\mathcal{Y}=1|x,\calW=C))\nonumber\\
& = k\tau(x) +k-1.\label{eq:imbal-0-1}
\end{align}
Here, $\tau(x)$ can again be recovered from model predictions using a
simple linear transformation.  

One remaining question is how to select the factor $k$.  An obvious
solution is using cross-validation to select the best value of $k$.
This solution requires large enough data to correctly select the best
model and may require significant computational resources.

In addition to cross-validation we propose a simple strategy to bring
class probabilities in both groups as close as possible to
$\frac{1}{2}$.  For ordinary classification the factor
$k=\frac{1}{2\P(\calY=0)}$ achieves this (see Equation~\ref{eq:row1}).
In uplift modeling/HTE estimation class probabilities are different in
both groups, and we will choose the simplest solution: compute the
flipping factor $k$ based on the average of probabilities of majority
classes in both groups:
\begin{align}
  k&=\frac{1}{2\frac{\P(\calY=m^C|\calW=C)+\P(\calY=m^T|\calW=T)}{2} }\nonumber\\
  &=\frac{1}{\P(\calY=m^C|\calW=C)+\P(\calY=m^T|\calW=T)},\label{eq:balancing}
\end{align}
where $m^C,m^T\in\{0,1\}$ are the majority classes in, respectively,
treatment and control groups.

\section{Class flipping for the CVT uplift model}\label{sec:cvt}

In this section we present a class flipping scheme designed
specifically for the Class Variable Transformation (CVT)
model~\citep{Jas}.  Its advantage is that it allows for converting an
arbitrary classifier into an uplift model which is trained once on a
modified dataset.  It is thus simpler than other metamodels,
e.g.~those found in~\citep{Kunzel}.

This model is addressed separately since it is especially susceptible
to class imbalance~\citep{NybergDMKD}, and has a specific form
requiring additional factors to be taken into account.  We begin with
a short overview of the model, details can be found in~\citep{Jas}.

\subsection{The CVT model and its properties} \label{sec:cvt-intro}

The CVT model is obtained by flipping the class in the control group,
concatenating it with the treatment group and building a single
classifier on the combined dataset.  It is equivalent to fitting a
classifier on a modified class variable defined as
\begin{align}
& \tilde{\mathcal{Y}}=
\begin{cases}
\mathcal{Y},& \mbox{ if } \calW=T,\\
1-\mathcal{Y},& \mbox{ if } \calW=C.
\end{cases}\label{eq:cvt}
\end{align}
To justify the model assume first that
$\mathbb{P}(\calW=T)=\mathbb{P}(\calW=C)=\frac{1}{2}$, i.e.~treatment
and control groups have equal sizes.  We have
\begin{align}
 \P(\tilde{\mathcal{Y}}=1|x) &= \P(\tilde{\mathcal{Y}}=1|x,\calW=T)\mathbb{P}(\calW=T) +\P(\tilde{\mathcal{Y}}=1|x,\calW=C)\mathbb{P}(\calW=C)\nonumber\\
 & = \frac{1}{2}\left(\mathbb{P}(\mathcal{Y}=1|x,\calW=T)+1-\mathbb{P}(\mathcal{Y}=1|x,\calW=C)\right)=\frac{\tau(x)+1}{2},\label{eq:z-deriv}
\end{align}
and $\tau(x)$ can be recovered from model predictions $\P(\tilde{\mathcal{Y}}=1|x)$ using linear transformation
\begin{equation}\label{eq:z-pred}
\tau(x) =  2\P(\tilde{\mathcal{Y}}=1|x)-1.
\end{equation}
The model can also be intuitively understood by looking
at~\citep[Table~9.1]{Haughton25}, adapted below to our notation:
\begin{center}
  \begin{tabular}{|c|c|c|}
    \hline
    & $\mathcal{Y}=0$ & $\mathcal{Y}=1$ \\
    \hline
Treatment & $\tilde{\mathcal{Y}}=0$ & $\tilde{\mathcal{Y}}=1$ \\\hline
Control & $\tilde{\mathcal{Y}}=1$ & $\tilde{\mathcal{Y}}=0$ \\\hline
  \end{tabular}
\end{center}
Let us look at whether the treatment effect was positive or negative
in each case.  For a treated case with a positive response
($\mathcal{Y}=1$) the treatment effect has to be either positive or
neutral depending on what would have happened, had the case been a
control.  On the other hand, for treated cases with negative response
the effect cannot be positive, it can only be negative or neutral.
For control cases, the situation is reversed, e.g.~the treatment
effect can be positive only if the response was negative.  This
provides an intuitive interpretation for the modified response
$\tilde{\mathcal{Y}}$: it is equal to $1$ if the treatment effect is
either positive or neutral, and equal to $0$ if the effect can only be
neutral or negative.  More detailed explanations can be found
in~\citep{Haughton25,Lo2014,Lai,Jas}.

When $\mathbb{P}(\calW=T)\neq\mathbb{P}(\calW=C)$ the
derivation~(\ref{eq:z-deriv}) is no longer valid.
In~\cite[Chapter~9]{Haughton25} in Equation~9.1 a modified version of
the estimator is given where conditional outcome probabilities are
divided by probabilities of treatment assignment.  It is proved that
the new estimator is correct regardless of treatment proportions.  The
authors of~\citep{Jas} suggest, without formal justification, an
equivalent approach based on undersampling the more frequent treatment
group such that the equality is satisfied.  Of course a weighting
scheme from Equation~\ref{eq:under-weights} can be used instead.
Below we use the weighting based version of the approach
from~\cite{Jas}, since it will be easier for us to combine weights
with those given in Section~\ref{sec:flip-weighted}.

We begin by justifying the use of the procedure from~\cite{Jas}.
Denote by $\mathbb{P}^*$ the distribution which would be obtained from
the original distribution $\P$ if undersampling (or reweighting) was
applied at the population level such that
$\mathbb{P}^*(\calW=T)=\mathbb{P}^*(\calW=C)=\frac{1}{2}$. We can
repeat the above reasoning to obtain
\begin{align}
 \P^*(\tilde{\mathcal{Y}}=1|x)  & = \frac{1}{2}\left(\mathbb{P}^*(\mathcal{Y}=1|x,\calW=T)+1-\mathbb{P}^*(\mathcal{Y}=1|x,\calW=C)\right)\nonumber\\
 & = \frac{1}{2}\left(\mathbb{P}(\mathcal{Y}=1|x,\calW=T)+1-\mathbb{P}(\mathcal{Y}=1|x,\calW=C)\right)=\frac{\tau(x)+1}{2},\label{eq:weighted-cvt}
\end{align}
where  equalities
$\mathbb{P}^*(\mathcal{Y}=1|x,\calW=T)=\mathbb{P}(\mathcal{Y}=1|x,\calW=T)$
and
$\mathbb{P}^*(\mathcal{Y}=1|x,\calW=C)=\mathbb{P}(\mathcal{Y}=1|x,\calW=C)$
follow from the fact that undersampling is applied uniformly and
independently of $\calX$ and $\calY$ within each group.

Let us conclude by providing the undersampling proportion $l$ for the
more frequent group.  Assume that we have more treatment observations
and we randomly remove $(1-l)$ proportion of them to balance the
groups.  Using Equation~\ref{eq:under} we get
\begin{equation}\label{eq:trt-balance}
\mathbb{P}^*(\calW=T) = \frac{l\mathbb{P}(\calW=T)}{l\mathbb{P}(\calW=T)+\mathbb{P}(\calW=C)},
\end{equation}
and solving $\mathbb{P}^*(\calW=T) =\frac{1}{2}$ for $l$ we get
$l=\frac{\mathbb{P}(\calW=C)}{\mathbb{P}(\calW=T)}$.  If the size of
the control group is higher, we need to resample it with
$l=\frac{\mathbb{P}(\calW=T)}{\mathbb{P}(\calW=C)}$.

\subsection{Correcting class imbalance in the CVT model}\label{sec:cvt-flip}

Let us first explain why the CVT model is so sensitive to class
imbalance~\citep{NybergDMKD}.  Assume that we have only a small number
of class $1$ observations in the treatment and control groups. Then
the transformed response $\tilde{\mathcal{Y}}$ is equal to $0$ for
most observations in the treatment group and equal to $1$ for most
observations in the control group.  More formally, suppose that
$\mathbb{P}(\mathcal{Y}=1|\calW=C)=\mathbb{P}(\mathcal{Y}=1|\calW=T)=\epsilon\approx
0$.  Noting that the class is reversed only in the control group we
get
\[
\mathbb{P}(\tilde{\mathcal{Y}}=1|\calW) = 
\begin{cases}
  \epsilon\approx 0 & \mbox{ if }\calW=T,\\
  1-\epsilon\approx 1 & \mbox{ if }\calW=C.
\end{cases}
\]
This causes
$\tilde{\mathcal{Y}}$ to become highly correlated with $\calW$, and
creates a strong incentive for the classifier built on the combined
dataset to focus on random differences between treatment and control
groups, not on predicting the true causal effect.

Correcting class imbalance in the CVT model should thus be done in a
way which leads to independence between $\tilde{\mathcal{Y}}$ and
$\calW$, and also balances $\tilde{\mathcal{Y}}$ as much as possible.

First, class flipping will be applied to correct for class imbalance,
transforming the original response $\calY$ into $\breve{\mathcal{Y}}$
(Equation~\ref{eq:breve-y}), then Equation~\ref{eq:cvt} is applied to
$\breve{\mathcal{Y}}$ resulting in a doubly-transformed class variable
which we denote $\tilde{\breve{\mathcal{Y}}}$ (the transformed,
flipped response).  Moreover, the more frequent group (treatment
or control) is undersampled (or weighted) according to
Equation~\ref{eq:trt-balance} so all probabilities will be computed in
a modified distribution $\P^*$.

Let us now justify that the model gives correct uplift predictions.
Suppose $0$ is the majority class and we use $k_0$ flipping
coefficient for that class in both treatment and control groups.

First, note that after class flipping transform given in
Equation~\ref{eq:imbal-0-0} we obtain a new dataset in which the CATE
$\tau(x)$ is $k_0$ times the uplift in the original dataset.  This is
true at the dataset level, so {\em any} uplift model trained on this
dataset will predict CATE scaled by $k_0$.  This is true also for the
weighted CVT model given in Equations~\ref{eq:weighted-cvt}
and~\ref{eq:z-pred}.  As a result we obtain a model which correctly
predicts $k_0\tau(x)$ and after multiplying by $\frac{1}{k_0}$ recover
the original CATE.  Note that this is true for any $0<k_0<1$.

Now we will find class flipping factors $k_0^C,k_1^C,k_0^T,k_1^T$
leading to independence between $\tilde{\breve{\mathcal{Y}}}$ and
$\calW$, which is equivalent to
\[
\mathbb{P}^*(\tilde{\breve{\mathcal{Y}}}=1|\calW=T) = \mathbb{P}^*(\tilde{\breve{\mathcal{Y}}}=1|\calW=C),
\]
and (Equation~\ref{eq:cvt}) to
\begin{align}\label{eq:indep-cvt}
\mathbb{P}^*(\breve{\mathcal{Y}}=1|\calW=T) &= \mathbb{P}^*(\breve{\mathcal{Y}}=0|\calW=C).
\end{align}
The first case we consider is when class $0$ is the majority class in
both groups: $k^C_1=k^T_1=1$, $k^C_0=k^T_0=k<1$.  This set of
coefficients guarantees model identifiability based on
Equations~\ref{eq:weighted-cvt} and~\ref{eq:imbal-0-0} as discussed above.
Applying Equation~\ref{eq:row1} we transform
Equation~\ref{eq:indep-cvt} as follows
\begin{align*}
1-\mathbb{P}^*(\breve{\mathcal{Y}}=0|\calW=T) &=\mathbb{P}^*(\breve{\mathcal{Y}}=0|\calW=C),\\
1-k\mathbb{P}^*(\mathcal{Y}=0|\calW=T) &= k\mathbb{P}^*(\mathcal{Y}=0|\calW=C).
\end{align*}
Solving for $k$ we obtain
\begin{equation}\label{eq:cvt-balancing}
k=\frac{1}{\mathbb{P}^*(\mathcal{Y}=0|\calW=T)+\mathbb{P}^*(\mathcal{Y}=0|\calW=C)}.
\end{equation}
Notice that this is the same formula as the one given in
Equation~\ref{eq:balancing}.  This equality is however just a
coincidence since both formulas were derived to satisfy different
conditions.

When class $1$ is the majority class in both groups then analogous
reasoning leads to $k^C_0=k^T_0=1$, $k^C_1=k^T_1=k<1$,
$k=\frac{1}{\mathbb{P}^*(\mathcal{Y}=1|\calW=T)+\mathbb{P}^*(\mathcal{Y}=1|\calW=C)}$.

The unusual situation with different majority classes in treatment and
control groups is described in 
Section~\ref{sec:cvt-different-maj} in the Appendix.

\subsubsection{Deterministic imbalance correction for the CVT model}\label{sec:cvt-deterministic}

Note that if random flipping given in Equation~\ref{eq:breve-y} is
used directly, the process is not deterministic and we cannot
guarantee that the transformed class variable will be independent of
treatment assignment in a given training sample.

However, recall from Section~\ref{sec:flip-weighted} that when the
deterministic version of the transform given in this section is used,
no sampling based randomness is introduced and exact probability
scaling is achieved also at the dataset level.  In this case exact
independence between $\calW$ and $\tilde{\breve{\mathcal{Y}}}$ is
achieved in the transformed dataset.

\section{Experiments}\label{sec:exper}

In this section we will demonstrate the effectiveness of the proposed
class flipping method on real data\footnote{Code to run the
experiments is publicly available at
\url{https://github.com/RudasKAP/uplift_imbalance}}.

\subsection{Experimental methodology}

We begin by describing the experimental methodology, as well as the
algorithms and datasets used in the evaluation.

\paragraph{Datasets.}
Unfortunately, despite numerous applications of RCTs, there are still
relatively few publicly available datasets.  We use five datasets:
three popular uplift modeling benchmarks, a publicly available A/B
testing dataset, and data from a randomized medical trial.  We use two
different targets in the Hillstrom and Criteo datasets to obtain a
total of seven benchmark datasets.

The first dataset is the well known Hillstrom
dataset~\citep{Hillstrom2008}, which contains the results of an e-mail
campaign for an Internet retailer.  The dataset contains information
about 64,000 customers who have been randomly split into three groups:
the first received an e-mail campaign advertising men's merchandise,
the second a campaign advertising women's merchandise, and the third
was kept as a control.  Two target classification variables are
available: `visit' (whether a person visited the website) and
`conversion' (whether a purchase was made).  The first target is
moderately imbalanced ($\mathbb{P}(\mathcal{Y}=1)\approx 0.147$), the
second is highly imbalanced ($\mathbb{P}(\mathcal{Y}=1)\approx
0.009$). In this paper we use only women's campaign group and both
visit and conversion as response variables, resulting in two datasets
{\tt Hillstrom-visit} and {\tt Hillstrom-conversion}.

The second dataset used in this work is a large dataset of online
advertisements shown to randomly targeted subgroup of users provided
by Criteo~\citep{Diemert2018}, which is rapidly becoming the standard
dataset for uplift modeling.  We use the second, corrected version of
the data.  The dataset contains 14 million records and 12 features.
We use the `visit' and `conversion' target variables which indicate,
respectively, whether the ad was clicked and whether a purchase was
made.  The minority class ($1$ indicating an ad click or purchase)
constitutes $4.7\%$ of cases for the `visit' target and $0.29\%$ for
the `conversion' target.  The dataset also features treatment
imbalance with control cases constituting only about 15\% of the
dataset.  In the remaining part of the paper we refer to those datasets
as {\tt Criteo-visit} and {\tt Criteo-conversion}.

The third dataset is simulated data provided by Starbucks which
mimics customer behavior during a real rewards
campaign.\footnote{There are many versions of the dataset, we use a
pre-aggregated one
from~\url{https://github.com/01KAT1/Marketing-Promotion-Campaign-Uplift-Modelling-Starbucks-Dataset}}
The dataset contains 84534 records and 7 predictors.  About $1.2$\% of
responses are positive, making the dataset highly imbalanced.
In the remaining part of
the paper we refer to this dataset as {\tt Starbucks}.

Another dataset used in the experiments is the {\tt MarketingAB}
dataset from Kaggle
(\url{https://www.kaggle.com/datasets/faviovaz/marketing-ab-testing}).
It contains data from an online A/B testing campaign with four
variables describing the time the ad was displayed and the total number of
ads shown to a user.  The dataset is imbalanced with the probability
of the ad being clicked around $2.5\%$.

Finally we include data from the GUSTO-I clinical
trial~\citep{Gusto-trial}.  The trial tested two strategies of
treating myocardial infractions (there is also a mixed strategy
excluded from our experiments).  Twenty-one pre-treatment variables
are included.  We used the version of the dataset available in the
{\tt predtools} R package
{\url{https://cran.r-project.org/web/packages/predtools/index.html}.
  Three variables describing study location ({\tt regl}, {\tt grpl},
  {\tt grps}) were dropped since they include a very large number of
  categories.  The {\tt ttr} variable was also dropped as it was
  correlated with treatment assignment.  The minority class (patient's
  death within 30 days) prevalence is about 7\%.

All datasets are summarized in Table~\ref{tab:upl-datasets}.
  
\begin{table}
  \begin{tabular}{|l|c|c|c|c|c|}
    \hline
    Dataset & \# records & features & \%treated & \% minority  & \% minority \\
     &  &  &  &  control &  treatment\\
    \hline
    Criteo-visit         & 13979592 & 12 & 85\%    & 3.82\%  & 4.85\%  \\\hline
    Criteo-conversion    & 13979592 & 12 & 85\%    & 0.19\%  & 0.31\%  \\\hline
    Hillstrom-visit      & 42693    & 8  & 50\%    & 10.62\% & 15.14\% \\\hline
    Hillstrom-conversion & 42693    & 8  & 50\%    & 0.57\%  & 0.88\%  \\\hline
    Starbucks            & 84534    & 7  & 50.11\% & 0.76\%  & 1.7\%   \\\hline
    MarketingAB          & 588101   & 4  & 96\%    & 1.79\%  & 2.55\%  \\\hline
    Gusto                & 40830    & 24 & 75.96\% & 7.32\%  & 7.01\%  \\\hline
  \end{tabular}
  \caption{Uplift dataset used in experiments}\label{tab:upl-datasets}
\end{table}

\paragraph{Uplift modeling algorithms.}

One of the most important classes of uplift modeling algorithms are
metamodels, i.e.~approaches based on converting standard probabilistic
classifiers into uplift models~\citep{Kunzel,Jas}.

The first metamodel used in our experiments is the {\tt T-learner}
(also called the two model approach)~\citep{Kunzel,RadcliffeTechRep}.
This is a natural reference, since it is the most popular uplift
modeling technique which usually performs very
well~\citep{RudasJaroszewicz}.

Another meta-learner used in the experiments is the CVT algorithm (see
Section~\ref{sec:cvt}) which is important since it is most affected by
class imbalance~\citep{Nyberg}.  We use the version which corrects for
imbalance between treatment group sizes, as suggested in~\citep{Jas}
and described in Section~\ref{sec:cvt-intro}.  Since we designed
special class imbalanced correction for this model, the results will
be described in a separate section.

Another algorithm we use are uplift trees
from~\citep{Rzepakowski2011}.  The trees are constructed similarly to
standard decision trees but with splitting criteria adapted to uplift
modeling.  Full details of uplift tree construction are beyond the
scope of this paper and can be found e.g.~in~\citep{Rzepakowski2011}.
There were several splitting criteria proposed, the most important
being the $E$ criterion based on squared Euclidean distance between
treatment and control class probability distributions, and the $KL$
criterion based on the Kullback-Leibler divergence.  In our
experiments we use the $KL$ criterion since the $E$ criterion is
unaffected by class flipping, leading to the trees built on the
flipped and original data being identical (up to numerical precision).
The proof of this fact can be found in
Section~\ref{sec:trees-E-unchanged} in the Appendix.

We limit tree height to 5 and impose a limit of a minimum of 100
samples (weighted count) per leaf node.  The reason for limiting
maximum tree height was performance on the Criteo datasets.  We also
found that having too few cases in a leaf of the tree makes
probability estimates within that leaf quite poor, and since we assess
models based on ranking measures, this leads to poor uplift curves.

\paragraph{Imbalance correction algorithms.}

The experiments will be divided into two parts.  First we will test
the general uplift imbalance correction given in
Section~\ref{sec:imbal-uplift}, then we will separately test imbalance
correction for the CVT algorithm as described in
Section~\ref{sec:cvt}.  The algorithms used in those two cases will
differ slightly.  First we describe the generic class imbalance
correction from Section~\ref{sec:imbal-uplift} which can work with any
uplift model.

Besides the uncorrected models, we evaluate versions adjusted with
various class imbalance correction methods.  The most important is of
course our class flipping method.  We use this method with $k$
selected both using Equations~\ref{eq:balancing}
or~\ref{eq:cvt-balancing}, and using cross-validation.

The most relevant competing method is the Stratified Undersampling
proposed in~\citep{NybergDMKD}.  The method applies the same
oversampling factor $k$ in both treatment and control groups.  We use
a weighting based version of the method given in
Equation~\ref{eq:under-weights}.  Since undersampling distorts the
predicted uplift,~\cite{NybergDMKD} suggested using cross-validation
to select an optimal factor $k$.  We include this method, as well as
the simpler method for selecting the value of $k$ using
Equations~\ref{eq:balancing} and~\ref{eq:cvt-balancing} which do not
require cross-validation.

Experiments for the CVT model are conducted separately since the model
is particularly strongly affected by class imbalance.  Also, it uses a
custom class flipping approach proposed in
Section~\ref{sec:cvt-flip} which corrects the dependence between
target variable and the treatment assignment and is one of the main
contributions of the paper.  We denote the method as {\tt FlippedCVT}
in the experiments.

In case of the CVT model we found cross-validation to perform poorly:
in every experiment we ran, using Equation~\ref{eq:cvt-balancing} gave
better results.  This was true also for Stratified Undersampling.  We
therefore do not include results for the cross-validated CVT method to
avoid cluttering the paper.

\paragraph{Base classifiers.}
Since most uplift models we use are metamodels, they can be applied
together with different base classifiers.  We use logistic regression,
decision trees and gradient boosted forests.  We use implementations
from the {\tt scikit-learn} package~\citep{scikit-learn} using default
parameters with the following exceptions.  For logistic regression
(denoted {\tt Logistic} in the rest of the text) we use a variable
standarization preprocessing step.  For decision trees (denoted {\tt
  DT} in the rest of the text) maximum tree height of 100 was set and
we required each leaf to contain a minimum of 5\% of training cases to
ensure good probability estimates (the parameter {\tt
  min\_samples\_leaf} in the {\tt scikit-learn} implementation).  As
was the case with uplift trees, we found that allowing tree leaves
with smaller numbers of samples falling into them leads to poor
performance since methods used in the paper rely on good probability
estimates.

For gradient boosting we used the {\tt
  HistGradientboostingClassifier} which uses histograms to speed up
the training process.  We included two versions of the method, with and
without early stopping, denoted, respectively, {\tt HistGB\_ES}, {\tt
  HistGB\_noES} in the rest of the text.

\paragraph{Evaluation measures.}

Evaluating uplift models is more challenging than evaluating
classifiers due to the Fundamental Problem of Causal Inference: for
single records we do not know whether the treatment was really
effective~\citep{Holland86}.  As a result, comparisons can only be
made at the level of groups of records.  A typical approach to
visualize uplift model performance are {\em uplift curves}.  The $x$
axis of the curve corresponds to the percentage of the population
assigned the treatment based on model predictions, and the $y$ axis to
the corresponding estimated net gain (i.e.~gain with respect to
estimated control group outcome) per individual.  Details can be found
in~\citep{Rzepakowski2011,RadcliffeTechRep,Verbeke}.

For example, in the upper left chart in Figure~\ref{fig:criteo} we can
see that targeting the whole population gives, on average, 0.01 increase
in click probability with respect to not targeting anyone.  However,
we can use uplift models to select 20\% of population where 80-90\% of
those gains are achieved.  Gains on the remaining 80\% of the population
are relatively small and not targeting them will not cause a
significant loss in profits.

To numerically compare models, Area Under the Uplift Curve (AUUC) will
be used.  Unfortunately, quite often the differences between AUUCs for
different models are small.  To make results easier to compare we
first compute the AUUC, then subtract the area under the diagonal of
the curve, and finally multiply the result by 1000.  This amplifies
small differences between models making them easier to interpret.
This measure was proposed in~\citep{NybergDMKD} and called mAUUC.

Each curve was computed by splitting the data into training ($50\%$)
and test sets ($50\%$) on which curves were computed.  For stability,
the process was repeated 100 times and the curves averaged (due to
performance only $20$ iterations were performed on the Criteo
datasets).  The legends in the figures show the average mAUUC for each
model and tables additionally show standard deviations.

\paragraph{Other uplift modeling metrics.}
Another popular type of curves used for assessing uplift models are
Qini curves~\citep{RadcliffeTechRep} which are identical to uplift
curves but re-scaled by the size of treated population.  An
interesting class of evaluation metrics for uplift models includes
various fairness criteria~\citep{Lo2025-fairness}.  Investigating how
class flipping combines with such measures is left as a future
research topic.

\subsection{Standard classification problems}\label{sec:standard-classif}

While the main topic of the paper is imbalance in uplift modeling/HTE
estimation, the proposed class flipping method is also applicable to
standard classification.  We begin by experimentally comparing it to
majority class undersampling in the classification setting.  This
experiment will help us explain the behavior of uplift models below.

We conjecture that flipping will have different properties than
undersampling.  On the one hand, if the data is
linearly separable, flipping will destroy this property; on the other hand,
undersampling changes the overall distribution of predictor variables (since
only one class is sampled and predictor distributions in both classes
differ) while flipping does not.  Moreover flipping requires a much
simpler correction to restore predicted probabilities.

Full experimental results and methodology are described in
Appendix~\ref{sec:a-classif}.  We used a total of 33 datasets detailed
in Table~\ref{tab:classif-datasets}, cross-validated Area Under the
ROC Curve was used to compare models\footnote{Note that AUROC is equivalent to the Mann–Whitney U test statistic.}.  Detailed results are shown
in Table~\ref{tab:classif-imbal}, here we only provide a summary in
Tables~\ref{tab:results_all_logistic},~\ref{tab:results_all_DT},~\ref{tab:results_all_hist_ES},
and~\ref{tab:results_all_hist_noES}, which give the number of times
(over all datasets) each undersampling correction method was best.
The results are split by classifier type.  If more than one model gave
identical best results on a given dataset a win was counted for all
such models.

\begin{table}
    \begin{minipage}{0.48\textwidth}
    \centering
  \begin{tabular}{|l|l|}
    \hline
    Method        & Number of wins \\ \hline
    None          & 9                                     \\ \hline
    Flipping      & 14                                     \\ \hline
    Undersampling & 12                                     \\ \hline
  \end{tabular}
  \caption{Number of wins over all datasets for logistic regression}\label{tab:results_all_logistic}
    \end{minipage}
    \hfill %
    \begin{minipage}{0.48\textwidth}
     \centering
  \begin{tabular}{|l|l|}
    \hline
    Method        & Number of wins \\ \hline
    None          & 5                                     \\ \hline
    Flipping      & 8                                     \\ \hline
    Undersampling & 24                                     \\ \hline
  \end{tabular}
  \caption{Number of wins over all datasets for decision trees}\label{tab:results_all_DT}
\end{minipage}

 \vspace{1.5em}

\begin{minipage}{0.48\textwidth}
\centering
  \begin{tabular}{|l|l|}
    \hline
    Method        & Number of wins \\ \hline
    None          & 29                                    \\ \hline
    Flipping      & 1                                    \\ \hline
    Undersampling & 5                                     \\ \hline
  \end{tabular}
  \caption{Number of wins over all datasets for HistGradientboosting with early stopping}\label{tab:results_all_hist_ES}
\end{minipage}
 \hfill %
\begin{minipage}{0.48\textwidth}
\centering
  \begin{tabular}{|l|l|}
    \hline
    Method        & Number of wins \\ \hline
    None          & 27                                     \\ \hline
    Flipping      & 4                                     \\ \hline
    Undersampling & 7                                     \\ \hline
  \end{tabular}
\caption{Number of wins over all datasets for HistGradientboosting without early stopping}\label{tab:results_all_hist_noES}
\end{minipage}
\end{table}

It can be seen that the benefits of imbalance correction are not
uniform.  Improvement (often insignificant) may be observed in many
cases but large deterioration is also possible.  One can conclude that
correcting imbalance for standard classification problem should not be
an automatic decision.

Table~\ref{tab:results_all_logistic} shows that the proposed class
flipping method outperforms undersampling for logistic regression. On
the other hand, Table~\ref{tab:results_all_DT} indicates that the
classic undersampling method achieves by far the best results for
decision trees. The experiments also suggest that for gradient
boosting no method for correcting class imbalance is needed as
boosting models can handle class imbalance on their own and
corrections are usually harmful.

Similar results have been observed in some of the more recent
papers on class imbalance.  For example in~\cite{galli-imbal} an
extensive analysis shows that imbalance correction decreases
performance of boosting-based methods.  Similar conclusions were
reached in~\cite{imbal-xgboost} (small improvements where observed
after parameter tuning).  In~\cite{imbal-convnn} it has been found
that for two class problems correcting class imbalance does not
improve neural network performance.

The poor performance of class flipping with decision trees
requires an explanation.  We believe that the reason is a follows.
All decision tree learning algorithms stop the tree construction once
a pure (i.e.~containing only one class) leaf is reached.  Once class
flipping is used, leaves containing only instances of the original
majority class are no longer pure.  This makes the tree construction
process continue leading to a large random tree.  We found this
problem hard to circumvent in the current decision tree
implementations and feel that the effect of class flipping and
undersampling on decision trees merits a separate research paper.

\subsection{Experimental evaluation of class imbalance correction for general uplift models}

We now evaluate the general uplift imbalance correction introduced in
Section~\ref{sec:imbal-uplift}.  We begin with the well known
T-learner meta model which we evaluate on several datasets in its
original form and with four types of class imbalance correction
summarized in the table below:%
{\small
\begin{center}
  \begin{tabular}{|l|l|}
\hline
Name & Description \\
\hline
{\tt T-learner} & Original T-learner \\\hline
{\tt Flipped T-learner} & Class flipping, $k$ chosen using~(\ref{eq:balancing}) \\\hline
{\tt Flipped T-learner CV} & Class flipping, $k$ chosen using cross-validation\\\hline
{\tt Stratified T-learner} & Stratified Undersampling~\citep{NybergDMKD}, $k$ chosen using~(\ref{eq:balancing})\\\hline
{\tt Stratified T-learner CV} & Stratified Undersampling, $k$ cross-validated\\\hline
  \end{tabular}
\end{center}
}%
Each meta-model is tested with four base learners described above.

The results of experiments are summarized in
Table~\ref{tab:meta-learners}, and example uplift curves are shown in
Figures~\ref{fig:criteo_T} and~\ref{fig:Criteo_C_T}.  Winning models
are denoted with bold; models whose AUUCs were closer than $0.001$
were considered equivalent.  The uncorrected T-learner handled some
cases well and produced a best result in 9 out of 28 cases.  However,
in the remaining 19 cases some form of class imbalance correction was
beneficial.  It can be seen that class flipping performed slightly
better than undersampling.  The proposed method with the fraction of
flipped cases selected using Equation~\ref{eq:balancing} ({\tt Flipped
  T-learner}) performed best in 11 cases.  The version based on
cross-validation ({\tt Flipped T-learner CV}) performed slightly worse
winning in 10 cases.  Cross-validation based stratified undersampling
({\tt Stratified T-learner CV}) was beneficial in 9 cases: slightly
less than class flipping.  When undersampling thresholds was selected
using Equation~\ref{eq:balancing} ({\tt Stratified T-learner}) the
method performed poorly.

Looking at specific base learners we see clear differences.  For
logistic regression using class flipping resulted in improvements in
four cases.  The original T-learner and stratified undersampling each
won in only two cases.  When decision trees were used as base
learners, class flipping did match the original T-learner's
performance in 6 cases but never improved on it.  Stratified
undersampling generally decreased performance, except for the {\tt
  MarketingAB} and {\tt Criteo-conversion} datasets.

Let us now move to the case of the gradient boosting base learner.  We
have tried two variants of this method: with and without early
stopping.  When early stopping was used flipping based imbalance
correction brought improvements and the original T-learner performed
poorly.  When no early stopping was used stratified undersampling was
advantageous.

To conclude, the T-learner shows some robustness to class imbalance
especially for decision trees, but imbalance correction may bring
visible improvements in many cases.  The proposed class flipping
method was the most beneficial for T-learners based on logistic
regression and gradient boosting with early stopping.  Stratified
undersampling (cross-validation based) is beneficial mostly for
gradient boosting without early stopping.

We believe that the experiments demonstrate that class flipping is a
valuable tool for addressing class imbalance, as it may bring
significant improvements and dominates undersampling based methods in
many cases.  Another advantage of class flipping is that the flipping
threshold selection based on Equation~\ref{eq:balancing} works well,
but for undersampling using cross-validation seems necessary, which
can be prohibitively expensive for larger datasets.

\begin{table}
\begin{tabular}{|l|c|c|c|c|c|}
\hline
\multicolumn{6}{|c|}{base learner: {\tt logistic}}\\\hline
Dataset & T-learner & Flipped & Flipped CV& Stratified & Stratified CV \\\hline
Criteo-conversion & 0.505$\pm$0.0109 & \textbf{0.5081$\pm$0.0061} & \textbf{0.5089$\pm$0.005} & -0.1167$\pm$0.0175 & \textbf{0.509$\pm$0.0049}\\\hline
Criteo-visit & \textbf{4.1467$\pm$0.0485} & 4.0792$\pm$0.0164 & 4.1219$\pm$0.03 & 0.7351$\pm$0.0764 & 4.1041$\pm$0.0486\\\hline
Gusto & 0.4925$\pm$1.2548 & \textbf{0.5191$\pm$1.0958} & 0.4789$\pm$1.1683 & 0.3353$\pm$1.018 & 0.3865$\pm$1.1286\\\hline
Hillstrom-conversion & 0.5489$\pm$0.2704 & \textbf{0.5718$\pm$0.239} & 0.5651$\pm$0.2492 & 0.5393$\pm$0.2666 & 0.5482$\pm$0.2666\\\hline
Hillstrom-visit & {7.669$\pm$0.9789} & 7.4292$\pm$0.8619 & 7.565$\pm$0.9515 & 7.3912$\pm$0.8377 & \textbf{7.6715$\pm$0.8699}\\\hline
MarketingAB & 2.8148$\pm$0.3516 & 2.895$\pm$0.2127 & \textbf{3.0948$\pm$0.2199} & 1.846$\pm$0.302 & 2.9593$\pm$0.3036\\\hline
Starbucks & \textbf{2.659$\pm$0.1998} & 2.6158$\pm$0.2069 & 2.63$\pm$0.2047 & 2.4968$\pm$0.1978 & {2.654$\pm$0.2014}\\\hline
\hline
\multicolumn{6}{|c|}{base learner: {\tt decision tree}}\\\hline
Criteo-conversion & {0.4006$\pm$0.0257} & {0.4006$\pm$0.0257} & {0.4006$\pm$0.0257} & 0.0391$\pm$0.0189 & \textbf{0.4741$\pm$0.0074}\\\hline
Criteo-visit & \textbf{3.8098$\pm$0.0631} & \textbf{3.8098$\pm$0.0631} & \textbf{3.8098$\pm$0.0631} & 0.4456$\pm$0.1486 & 2.2602$\pm$0.2526\\\hline
Gusto & \textbf{0.593$\pm$1.4828}& \textbf{0.5926$\pm$1.4822} & \textbf{0.5925$\pm$1.4791} & 0.2604$\pm$1.1304 & 0.5292$\pm$1.3372\\\hline
Hillstrom-conversion & \textbf{0.443$\pm$0.329} & \textbf{0.4428$\pm$0.3291} & \textbf{0.4431$\pm$0.3289} & 0.398$\pm$0.3011 & 0.405$\pm$0.329\\\hline
Hillstrom-visit & \textbf{5.6928$\pm$0.9434} & \textbf{5.6928$\pm$0.9434} & \textbf{5.6928$\pm$0.9434} & 5.0418$\pm$0.9294 & 5.5142$\pm$ 0.9213\\\hline
MarketingAB & 2.87$\pm$0.3772 & 2.87$\pm$0.3772 & 2.8674$\pm$0.3917 & 1.9651$\pm$0.4713 & \textbf{2.954$\pm$0.3733}\\\hline
Starbucks & \textbf{2.4558$\pm$0.2093} & \textbf{2.4556$\pm$0.2097} & \textbf{2.4558$\pm$0.2093} & 2.3153$\pm$0.2173 & 2.4461$\pm$0.2117\\\hline
\hline
\multicolumn{6}{|c|}{base learner: {\tt Hist Gradient Boosting Early Stopping}}\\\hline
Criteo-conversion & 0.2401$\pm$0.0425 & \textbf{0.4574$\pm$0.0141} & 0.4492$\pm$0.0159 & 0.1017$\pm$0.0299 & 0.4336$\pm$0.0135\\\hline
Criteo-visit & 4.0955$\pm$0.0605 & 4.1697$\pm$0.062 & \textbf{4.1772$\pm$0.0664} & 2.2504$\pm$0.0665 & 4.1322$\pm$0.0737\\\hline
Gusto & 0.9018$\pm$1.3323 & 0.4555$\pm$1.6543 & 0.6812$\pm$1.356 & \textbf{1.0749$\pm$1.1326} & 0.8908$\pm$1.3307\\\hline
Hillstrom-conversion & \textbf{0.3173$\pm$0.3298} & 0.0529$\pm$0.3764 & 0.1987$\pm$0.388 & 0.2857$\pm$0.3085 & 0.2982$\pm$0.3087\\\hline
Hillstrom-visit & 5.0576$\pm$1.0168 & 5.4325$\pm$1.1536 & \textbf{5.4855$\pm$1.2023} & 4.3992$\pm$1.0231 & 4.9229$\pm$1.1679\\\hline
MarketingAB & 2.4422$\pm$0.3301 & \textbf{2.6739$\pm$0.3785} & 2.5632$\pm$0.3705 & 1.8186$\pm$0.368 & 2.6416$\pm$0.3101\\\hline
Starbucks & 2.2262$\pm$0.2126 & 1.3356$\pm$0.4476 & \textbf{2.3421$\pm$0.225} & 2.2157$\pm$0.2234 & 2.3011$\pm$0.1926\\\hline
\hline 
\multicolumn{6}{|c|}{base learner: {\tt Hist Gradient Boosting No Early Stopping}}\\\hline
Criteo-conversion & 0.3109$\pm$0.0274 & 0.4177$\pm$0.0209 & 0.4155$\pm$0.0137 & 0.1148$\pm$0.0218 & \textbf{0.4238$\pm$0.0142}\\\hline
Criteo-visit & 4.0664$\pm$0.0906 & 4.0886$\pm$0.0694 & 4.1097$\pm$0.0601 & 2.2388$\pm$0.0707 & \textbf{4.128$\pm$0.0617}\\\hline
Gusto & 1.712$\pm$1.5288 & 0.9387$\pm$1.4979 & 1.0468$\pm$1.3254 & 1.7373$\pm$1.2438 & \textbf{1.7517$\pm$1.4964}\\\hline
Hillstrom-conversion & \textbf{0.281$\pm$0.3052} & 0.2728$\pm$0.3169 & 0.2736$\pm$0.3103 & 0.1974$\pm$0.3017 & 0.2609$\pm$0.2788\\\hline
Hillstrom-visit & 3.7638$\pm$1.0139 & \textbf{3.9437$\pm$1.0779} & 3.8311$\pm$1.174 & 3.3769$\pm$1.0328 & 3.4637$\pm$1.304\\\hline
MarketingAB & 2.1427$\pm$0.3462 & 1.604$\pm$0.3893 & 1.8216$\pm$0.3701 & 1.5084$\pm$0.3066 & \textbf{2.3323$\pm$0.3413}\\\hline
Starbucks & 1.9295$\pm$0.218 & 1.5473$\pm$0.2226 & 1.6192$\pm$0.2083 & 1.766$\pm$0.2257 & \textbf{1.9761$\pm$0.2225}\\\hline
\hline
Wins & 9 & 11 & 10 & 1 & 9 \\\hline
\end{tabular}
  \caption{mAUUC for T-learners with various base learners and class imbalance correction methods}\label{tab:meta-learners}
\end{table}

\begin{figure}[t]
\begin{center}
\includegraphics[scale=\chartscale]{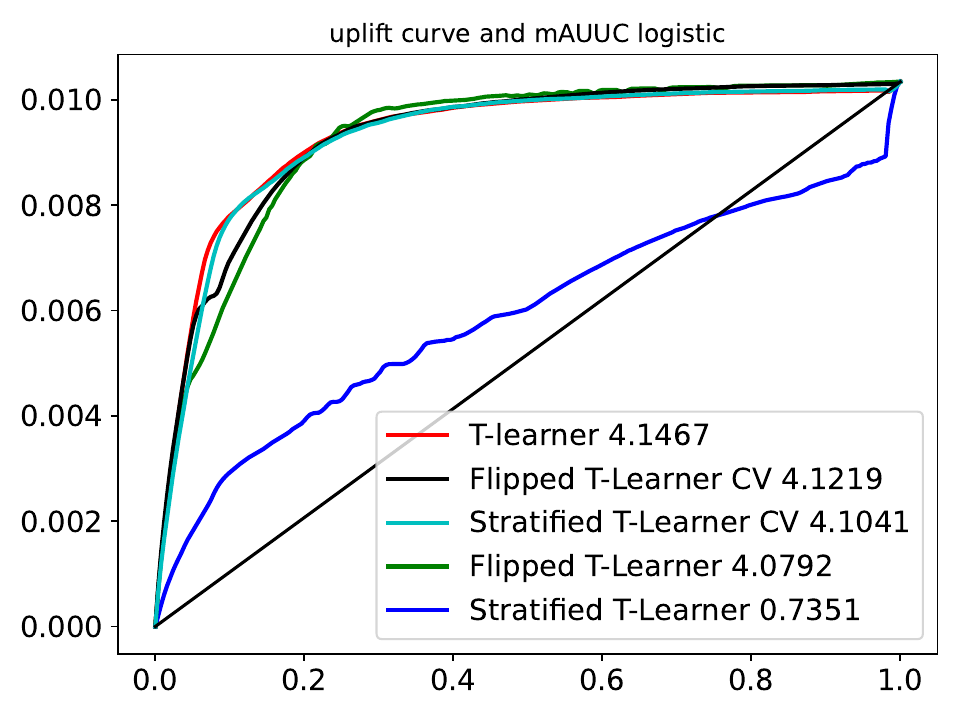}
\includegraphics[scale=\chartscale]{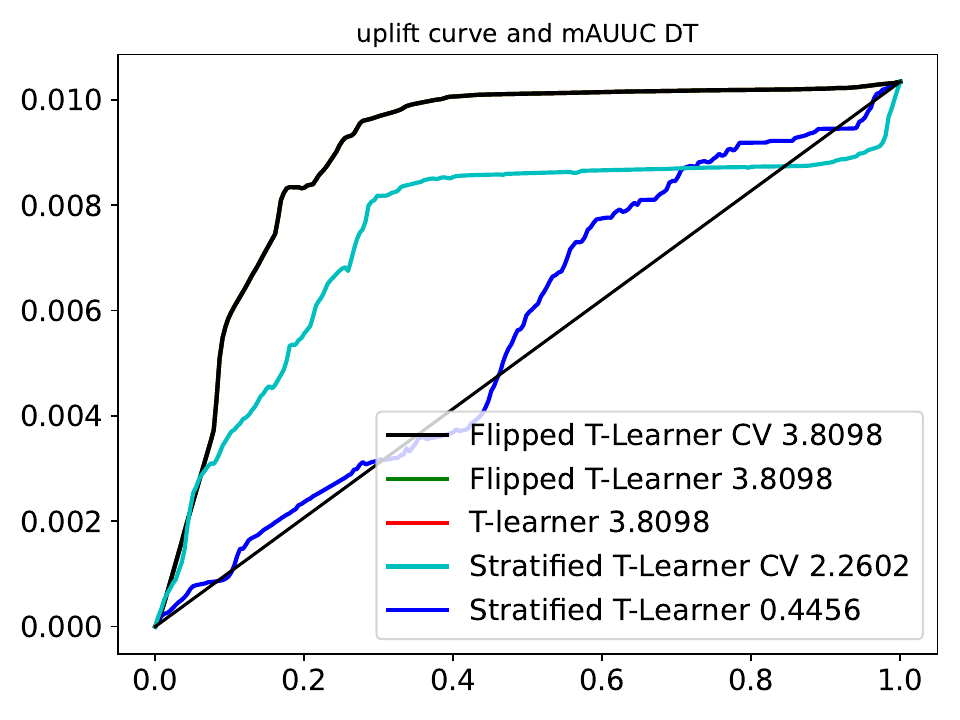}
\includegraphics[scale=\chartscale]{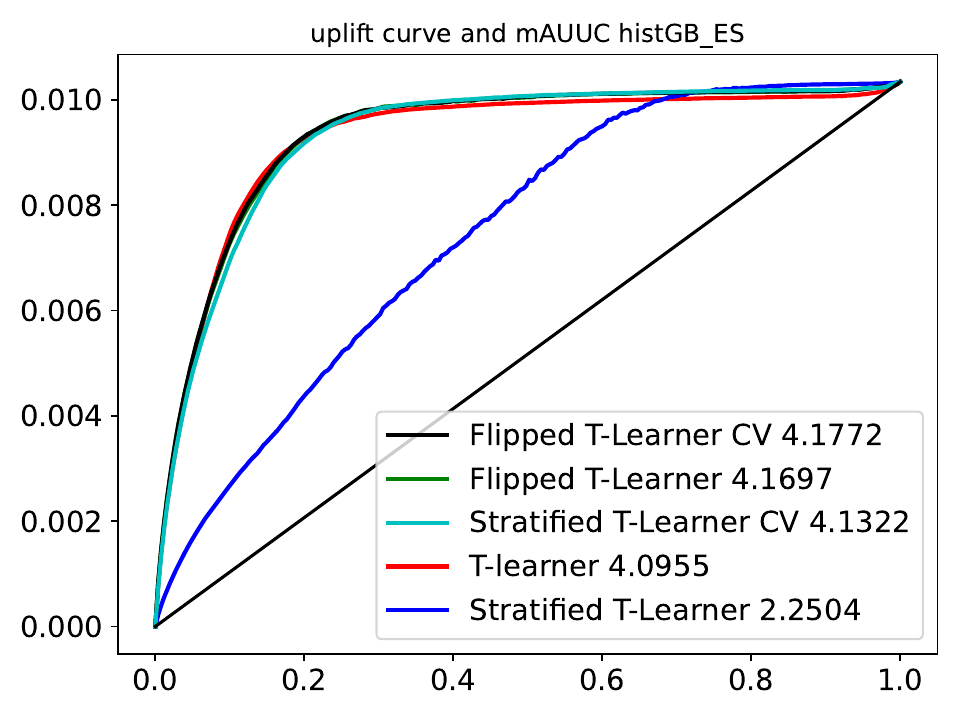}
\includegraphics[scale=\chartscale]{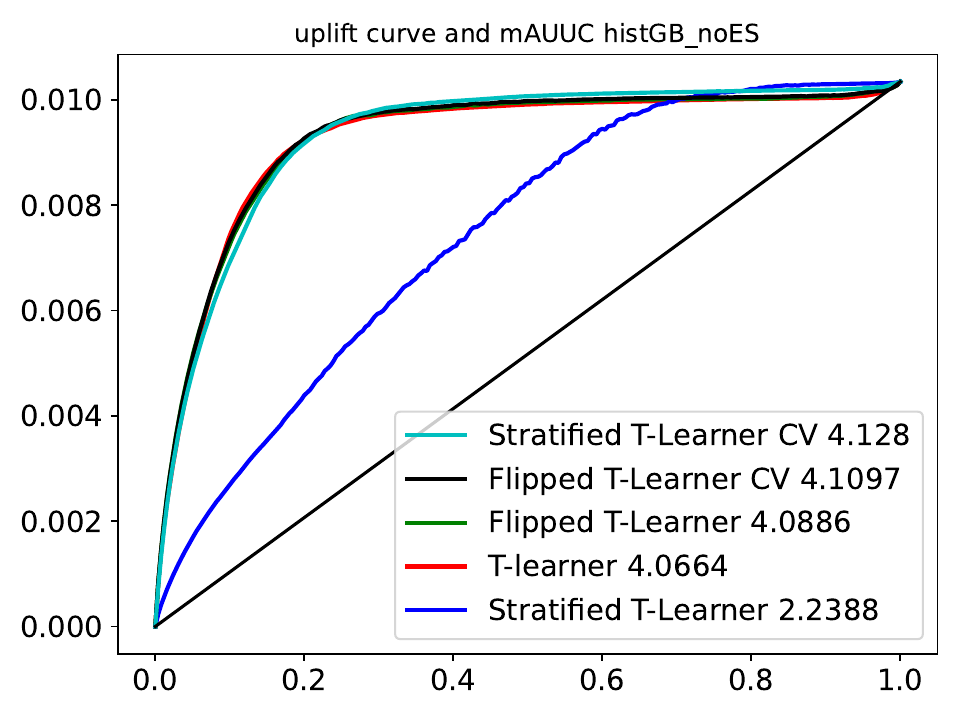}
\end{center}
\caption{Uplift curves and mAUUC for the T-learner on the {\tt Criteo-visit} dataset for logistic regression, decision tree and HistGradientboosting used as base learners}\label{fig:criteo_T} 
\end{figure}

\begin{figure}[t]
\begin{center}
\includegraphics[scale=\chartscale]{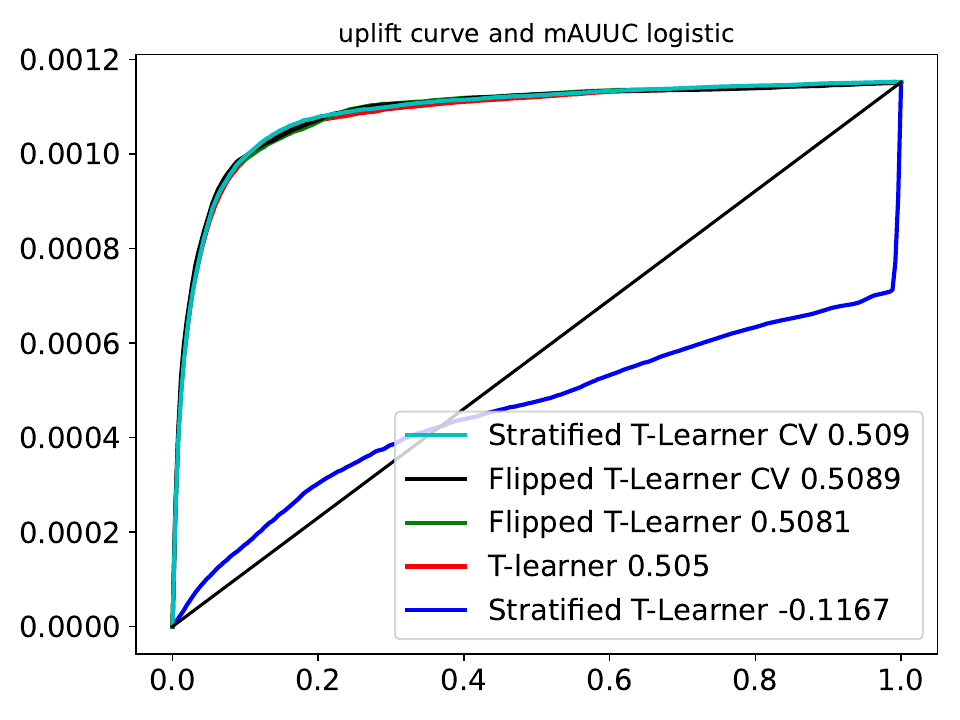}
\includegraphics[scale=\chartscale]{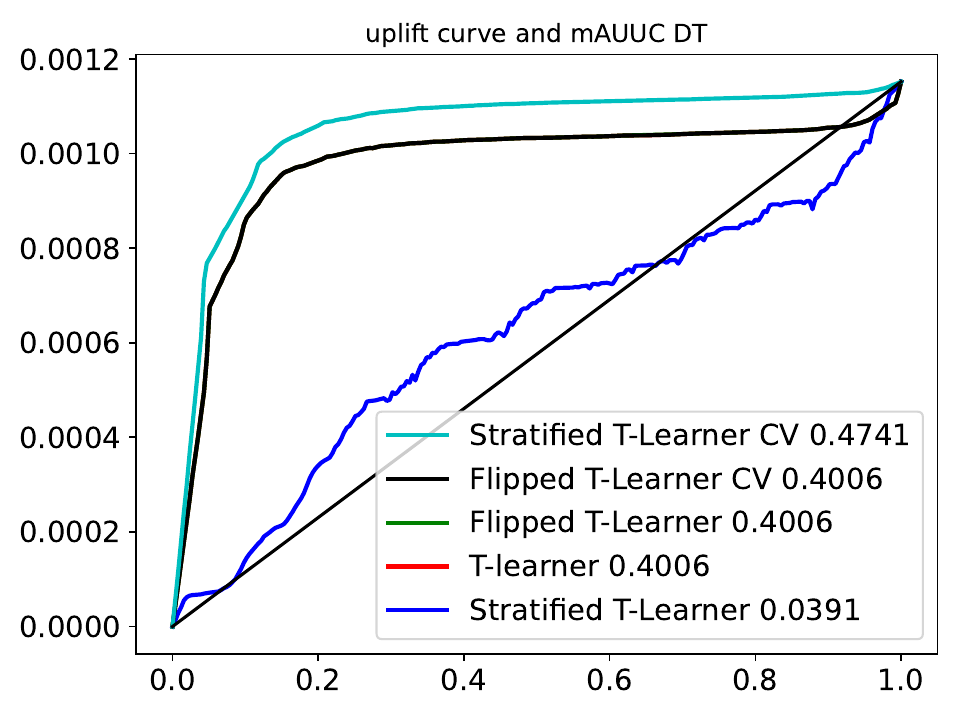}
\includegraphics[scale=\chartscale]{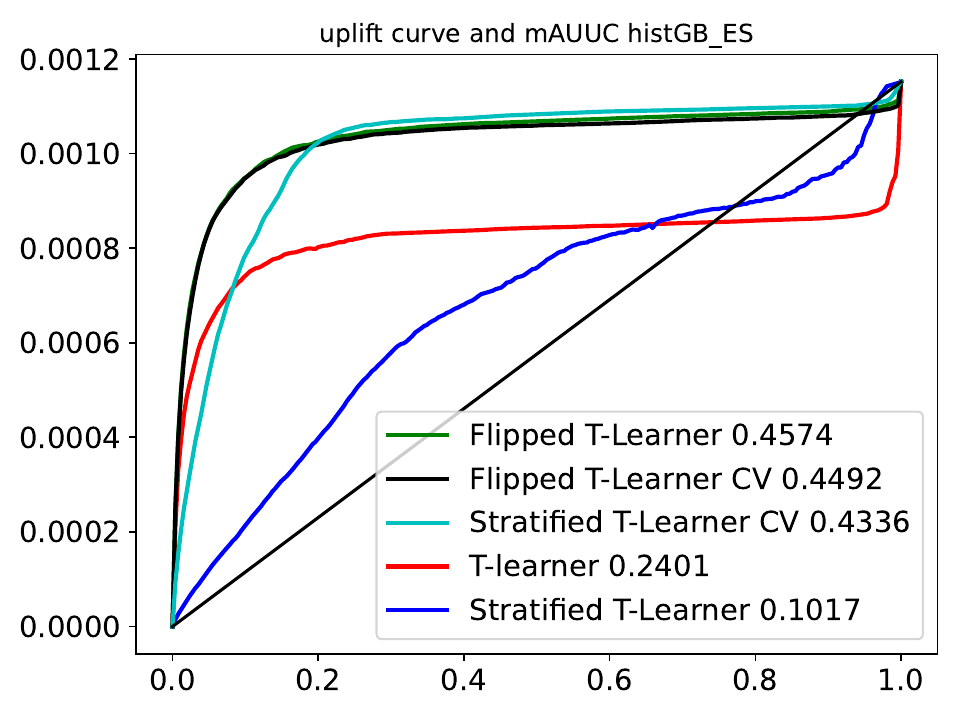}
\includegraphics[scale=\chartscale]{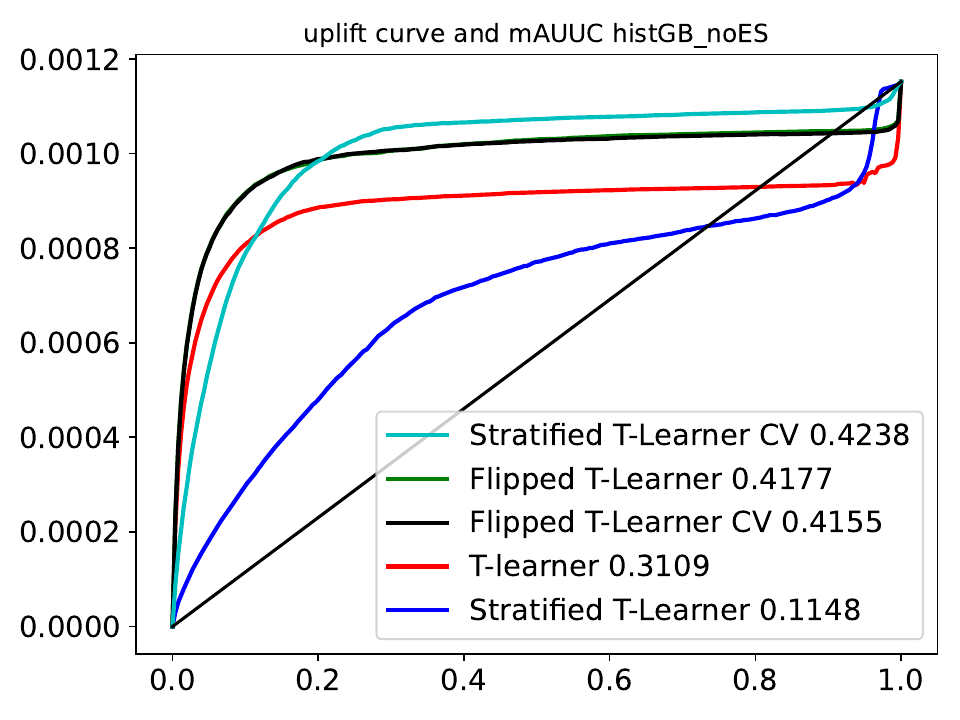}
\end{center}
\caption{Uplift curves and mAUUC for the T-learner on the {\tt Criteo-conversion} dataset for logistic regression, decision tree and HistGradientboosting used as base learners}\label{fig:Criteo_C_T} 
\end{figure}

\subsection{Uplift trees}

Another generic uplift model we use in our experiments are uplift
trees from~\cite{Rzepakowski2011} with the $KL$ splitting criterion.
The results are shown in Table~\ref{tab:utree}, and example curves in
Figures~\ref{fig:Criteo_U} and~\ref{fig:Criteo_C_U}.  It can be seen
that the model does benefit from class imbalance correction in about
half of the cases.  Stratified undersampling consistently resulted in
performance degradation.  Example uplift curves can be seen in
Figures~\ref{fig:Criteo_U} and~\ref{fig:Criteo_C_U}.

\begin{table}
\begin{tabular}{|l|c|c|c|c|c|}
\hline
Dataset & Uplift Tree & Flipped & Stratified  \\\hline
Criteo-conversion & 0.1708$\pm$0.078 & \textbf{0.4543$\pm$0.0171} & -0.1044$\pm$0.0282\\\hline
Criteo-visit & 3.642$\pm$0.1187 & \textbf{4.1394$\pm$0.0348} & 1.1212$\pm$0.1805\\\hline
Gusto & \textbf{0.5736$\pm$1.3719} & 0.4218$\pm$1.501 & 0.4725$\pm$1.2391\\\hline
Hillstrom-conversion & \textbf{0.2856$\pm$0.3188} & 0.1957$\pm$0.3188 & 0.0919$\pm$0.3182\\\hline
Hillstrom-visit & 5.2907$\pm$1.3087 & \textbf{5.3717$\pm$1.2272} & 4.5469$\pm$1.1343\\\hline
MarketingAB & \textbf{2.8227$\pm$0.3841} & 2.8184$\pm$0.4171 & 1.9125$\pm$0.5345\\\hline
Starbucks & \textbf{2.1961$\pm$0.3896} & 2.137$\pm$0.2387 & 0.6406$\pm$0.4143\\\hline
\hline
Wins    & 4  & 3 & 0\\\hline
\end{tabular}
  \caption{mAUUC for uplift trees with class imblance corrections}\label{tab:utree}
\end{table}

\begin{figure}[t]
  \begin{minipage}[b]{0.5\linewidth}
 \begin{center}
 \includegraphics[scale=\chartscale]{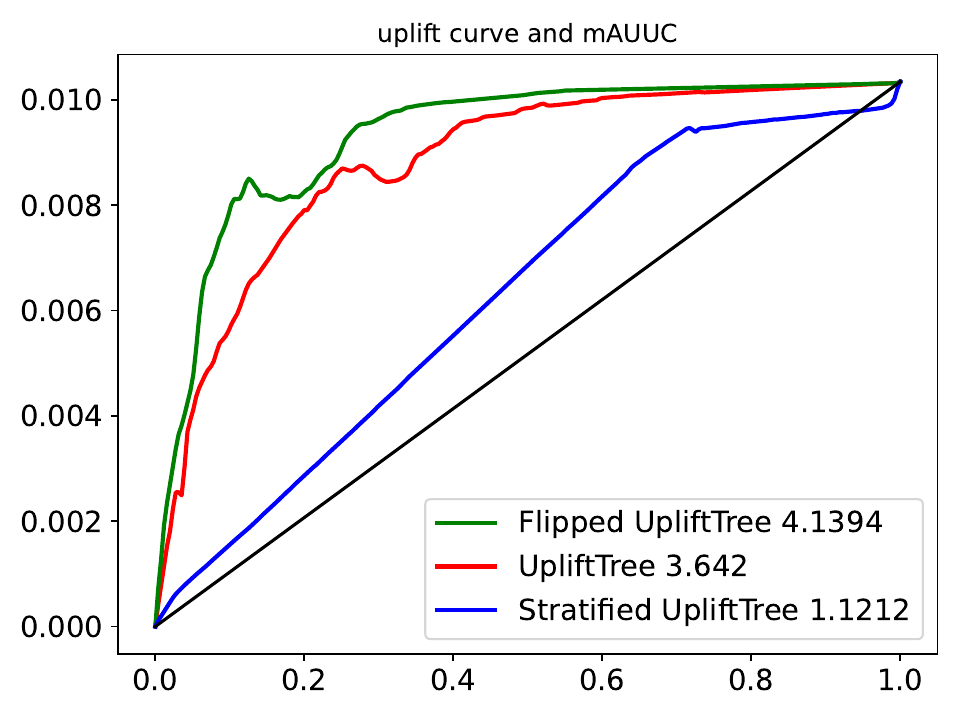}
    \end{center}
\caption{Uplift curves and mAUUC for uplift tree with different class imbalance correction methods on the {\tt Criteo-visit} dataset}\label{fig:Criteo_U}
  \end{minipage}
~~\begin{minipage}[b]{0.5\linewidth}
    \begin{center}
\includegraphics[scale=\chartscale]{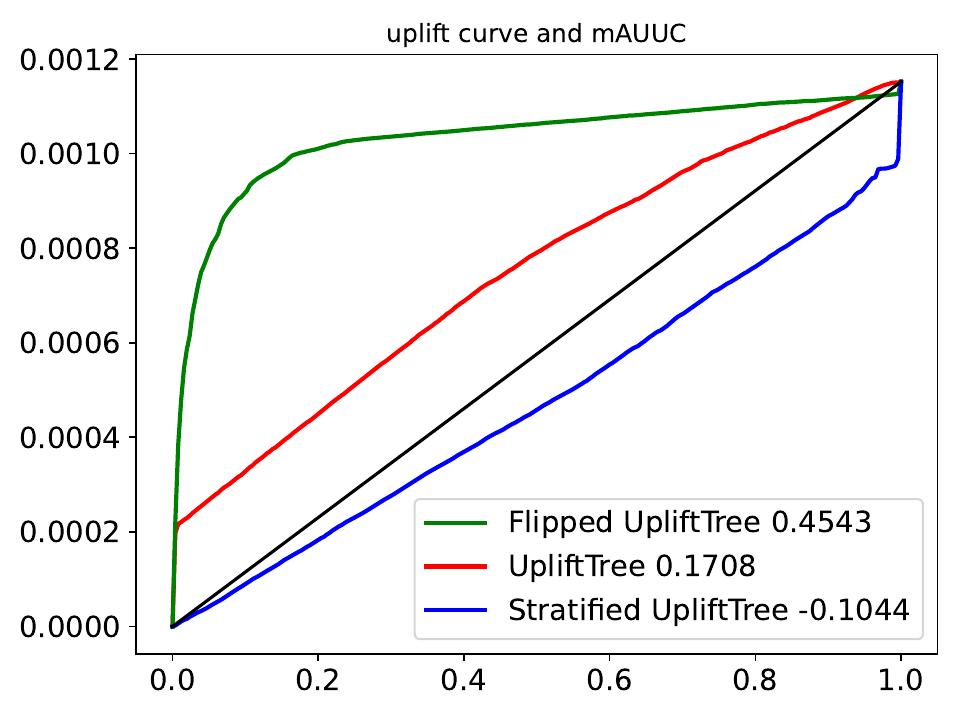}
    \end{center}
\caption{Uplift curves and mAUUC for uplift tree with different class imbalance correction methods on the {\tt Criteo-conversion} dataset}\label{fig:Criteo_C_U}  
  \end{minipage}
\end{figure}

\subsection{Experimental evaluation of class imbalance correction for the CVT model}

In this section we present experimental evaluation of the proposed
class flipping method combined with the CVT meta-learner presented in
Section~\ref{sec:cvt}.  This scenario is where class flipping is most
beneficial and outperforms other imbalance correction models as well
as uncorrected CVT approach.  The reason was given
in~Section~\ref{sec:cvt-flip}: the transformation on which the model
is based creates a spurious correlation between target and treatment
assignment which is especially strong when class imbalance is present.

For each dataset we used all four base learners and four different
metamodels: the original CVT model ({\tt CVT}), the CVT model with
class flipping ({\tt Flipped CVT}), the CVT model with stratified
imbalance correction of~\citep{NybergDMKD} ({\tt Stratified CVT}), and
the two model ({\tt T-learner}) approach  without imbalance
correction.  We used the last model as a reference, since it
is the most popular model for uplift modeling and often works well
in practice.

In this section we do not use cross-validation to select imbalance
correction ratio, since the method performed poorly for both class
flipping and undersampling.  The CVT model seems to benefit most from
the independence between target variable and the treatment indicator
which requires careful parameter selection.

Table~\ref{tab:cvt-flip} shows the results for all datasets, and
Figures~\ref{fig:criteo} and~\ref{fig:Criteo_C} show uplift curves for
two of them.  Figures for the remaining five datasets can be found in
Section~\ref{sec:additional-exper} in the Appendix.

\begin{table}
\begin{tabular}{|l|c|c|c|c|c|}
\hline
\multicolumn{5}{|c|}{base learner: {\tt logistic}}\\\hline
Dataset & T-learner & CVT & Flipped CVT & Stratified CVT \\\hline
Criteo-conversion & 0.505$\pm$ 0.0109 & -0.2732$\pm$ 0.0136 & \textbf{0.5088$\pm$ 0.0052} & 0.256$\pm$ 0.0176 \\\hline
Criteo-visit & \textbf{4.1467$\pm$ 0.0485} & 3.5776$\pm$ 0.1185 & 4.0815$\pm$ 0.0195 & 3.5746$\pm$ 0.0519 \\\hline
Gusto & \textbf{0.4925$\pm$ 1.2548} & -0.5514$\pm$ 1.2595 & 0.411$\pm$ 1.0289 & -0.5839$\pm$ 1.0367 \\\hline
Hillstrom-conversion & 0.5489$\pm$ 0.2704 & 0.3602$\pm$ 0.2874 & \textbf{0.5815$\pm$ 0.2496} & 0.5039$\pm$ 0.2573 \\\hline
Hillstrom-visit & \textbf{7.669$\pm$ 0.9789} & 6.8102$\pm$ 1.0392 & 7.3782$\pm$ 0.85 & 6.8905$\pm$ 0.835 \\\hline
MarketingAB & \textbf{2.8148$\pm$ 0.3516} & -0.6765$\pm$ 0.2432 & 2.7911$\pm$ 0.2102 & 0.9857$\pm$ 0.3465 \\\hline
Starbucks & \textbf{2.659$\pm$ 0.1998} & 1.474$\pm$ 0.497 & 2.6084$\pm$ 0.2076 & 2.4157$\pm$ 0.2165 \\\hline
\hline
\multicolumn{5}{|c|}{base learner: {\tt decision tree}}\\\hline
Criteo-conversion & 0.4009$\pm$ 0.0257 & -0.3772$\pm$ 0.013 & \textbf{0.4764$\pm$ 0.0078} & 0.1162$\pm$ 0.0421 \\\hline
Criteo-visit & 3.8094$\pm$ 0.065 & 2.2053$\pm$ 0.6018 & \textbf{4.0325$\pm$ 0.1905} & 3.1559$\pm$ 0.2362\\\hline
Gusto & 0.5869$\pm$ 1.4834 & -0.0125$\pm$ 1.2149 &  \textbf{1.6852$\pm$ 1.6737} & 0.9674$\pm$ 1.3885 \\\hline
Hillstrom-conversion & \textbf{0.4419$\pm$ 0.3317} & -0.0251$\pm$ 0.3423 & 0.3907$\pm$ 0.3105 & 0.2905$\pm$ 0.3083 \\\hline
Hillstrom-visit & 5.6965$\pm$ 0.9454 & 4.456$\pm$ 1.3977 & \textbf{5.8165$\pm$ 1.0234} & 5.149$\pm$ 1.233 \\\hline
MarketingAB & 2.8722$\pm$ 0.3811 & 1.5097$\pm$ 0.4485 & \textbf{3.0976$\pm$ 0.3053} & 2.1234$\pm$ 0.3618\\\hline
Starbucks & \textbf{2.4548$\pm$ 0.2107} &  0.513$\pm$ 0.4167 & 2.4322$\pm$ 0.2137 & 2.0238$\pm$ 0.3524 \\\hline
\hline
\multicolumn{5}{|c|}{base learner: {\tt Hist Gradient Boosting Early Stopping}}\\\hline
Criteo-conversion & 0.2592$\pm$ 0.0428 &  -0.2$\pm$ 0.0199 & \textbf{0.4427$\pm$ 0.0212} & 0.2354$\pm$ 0.0216 \\\hline
Criteo-visit & 4.0993$\pm$ 0.068 & 2.9145$\pm$ 0.0772 & \textbf{4.3858$\pm$ 0.1071} & 3.3815$\pm$ 0.0844 \\\hline
Gusto & 1.0092$\pm$ 1.4251 & 0.1266$\pm$ 1.2442 & 1.0557$\pm$ 1.5888 & \textbf{1.6075$\pm$ 1.3952} \\\hline
Hillstrom-conversion & \textbf{0.3496$\pm$ 0.3146} & -0.0623$\pm$ 0.3262 & 0.1507$\pm$ 0.3453 &  0.2745$\pm$ 0.3361 \\\hline
Hillstrom-visit & 5.1295$\pm$ 1.1342 & 4.3624$\pm$ 1.5339 & \textbf{6.0095$\pm$ 1.0147} &  5.3434$\pm$ 1.1081 \\\hline
MarketingAB & 2.4708$\pm$ 0.3361 & 1.1929$\pm$ 0.3217 & \textbf{2.6503$\pm$ 0.4384} & 2.2457$\pm$ 0.3857 \\\hline
Starbucks & \textbf{2.2514$\pm$ 0.2181} & 0.5101$\pm$ 0.3576 & 1.3151$\pm$ 0.4229 & 1.8336$\pm$ 0.2934 \\\hline
\hline 
\multicolumn{5}{|c|}{base learner: {\tt Hist Gradient Boosting No Early Stopping}}\\\hline
Criteo-conversion & 0.3063$\pm$ 0.0246 & -0.1987$\pm$ 0.0168 & \textbf{0.4879$\pm$ 0.0109} & 0.278$\pm$ 0.0192 \\\hline
Criteo-visit & 4.0689$\pm$ 0.0548 & 2.8272$\pm$ 0.0697 & \textbf{4.4337$\pm$ 0.0482} & 3.3922$\pm$ 0.0903\\\hline
Gusto & 1.712$\pm$ 1.5289 & 0.1642$\pm$ 1.2305 &  1.6413$\pm$1.4773 & \textbf{1.6522$\pm$ 1.2718} \\\hline
Hillstrom-conversion & 0.2809$\pm$ 0.3052 & -0.0497$\pm$ 0.3365 & \textbf{0.3021$\pm$ 0.3153} &  0.2351$\pm$ 0.3088\\\hline
Hillstrom-visit & 3.7636$\pm$ 1.0139 & 2.3898$\pm$ 1.1387 & \textbf{4.1455$\pm$ 1.0997} & 3.3925$\pm$ 1.143\\\hline
MarketingAB & 2.1508$\pm$ 0.3503 & 0.8929$\pm$ 0.3073 & \textbf{2.8312$\pm$ 0.3053} & 2.4695$\pm$ 0.3239 \\\hline
Starbucks & \textbf{1.9295$\pm$0.2181} & 0.416$\pm$0.2708 & 1.8661$\pm$ 0.2045 & 1.402$\pm$ 0.2726 \\\hline
\hline
Wins & 10 & 0 & 16 & 2 \\\hline
\end{tabular}
  \caption{mAUUC for different CVT-learner methods and T-learner}\label{tab:cvt-flip}
\end{table}

\begin{figure}[t]
\begin{center}
\includegraphics[scale=\chartscale]{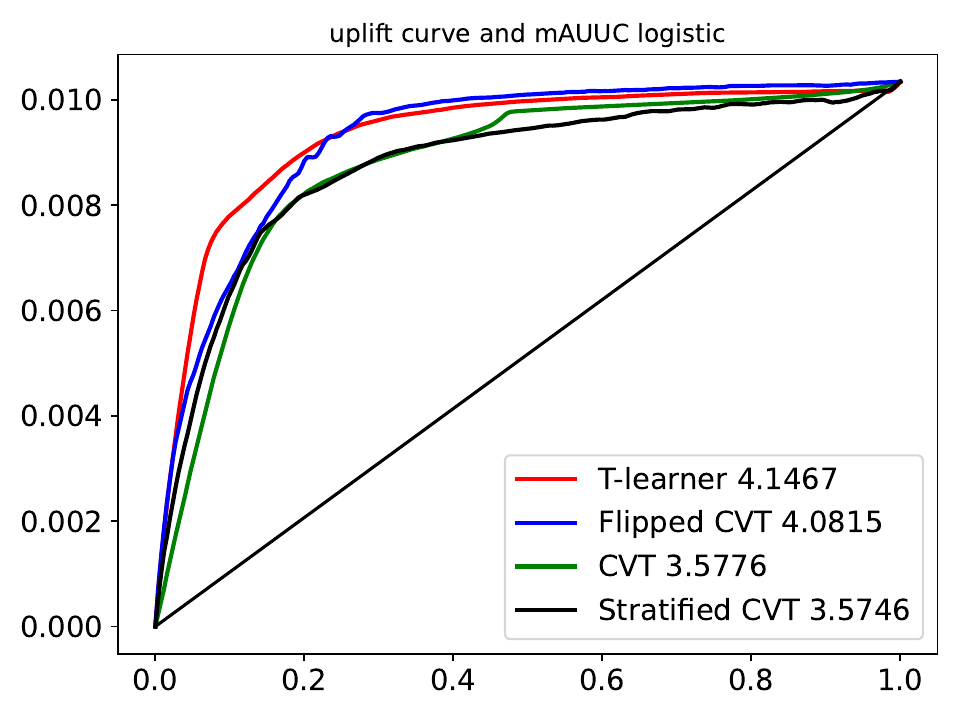}
\includegraphics[scale=\chartscale]{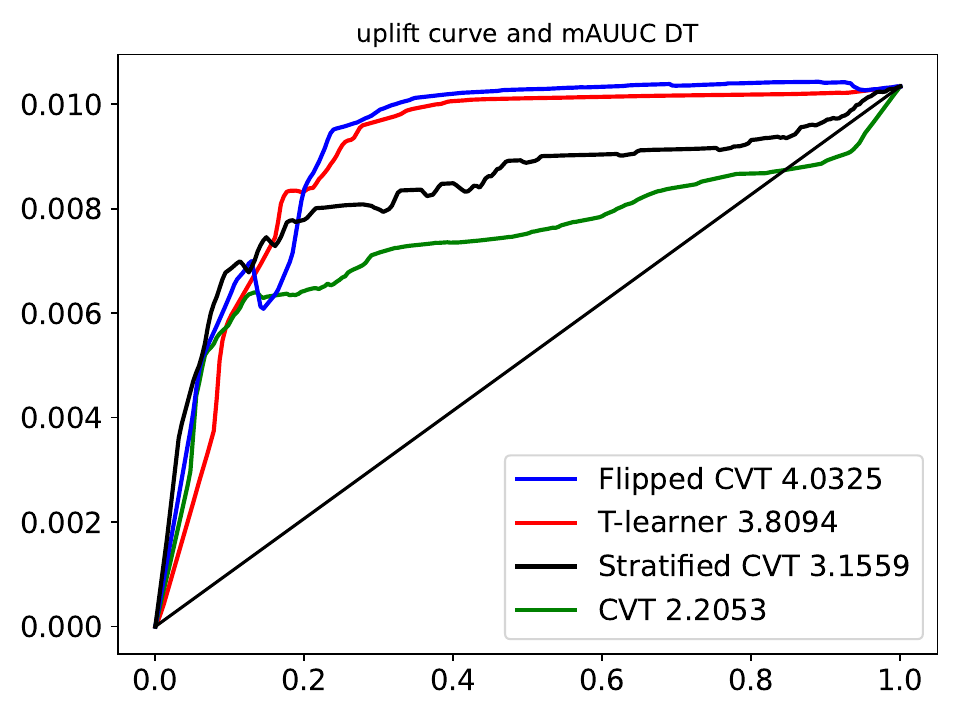}
\includegraphics[scale=\chartscale]{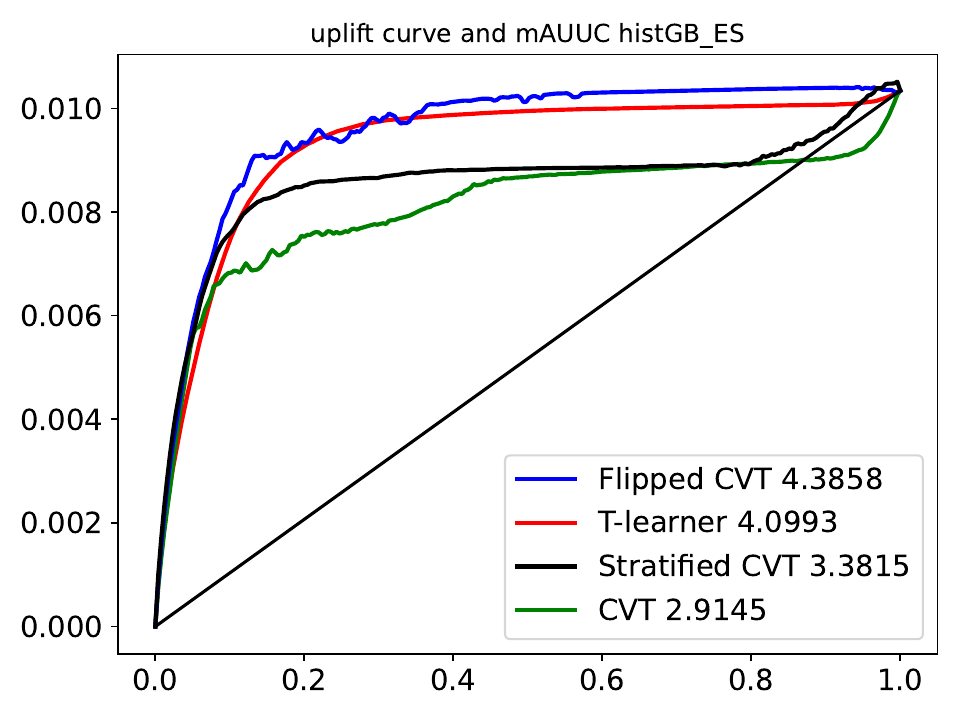}
\includegraphics[scale=\chartscale]{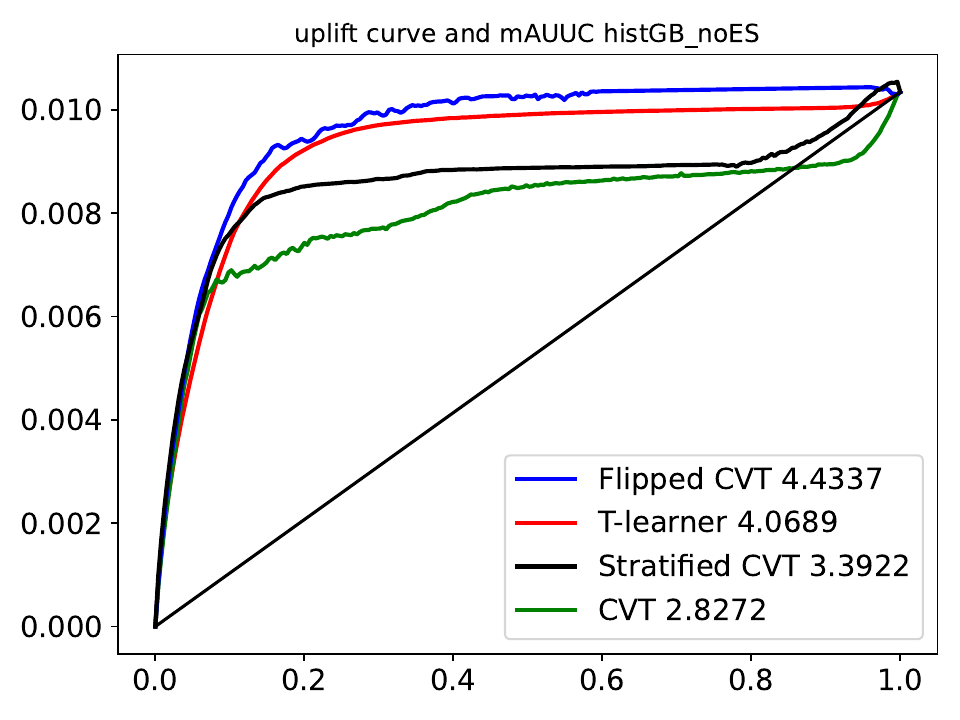}
\end{center}
\caption{Uplift curves and mAUUC for different CVT-learner methods and T-learner on {\tt Criteo-visit} dataset for logistic regression, decision tree and HistGradientboosting used as base learners}\label{fig:criteo} 
\end{figure}

\begin{figure}[H]
\begin{center}
\includegraphics[scale=\chartscale]{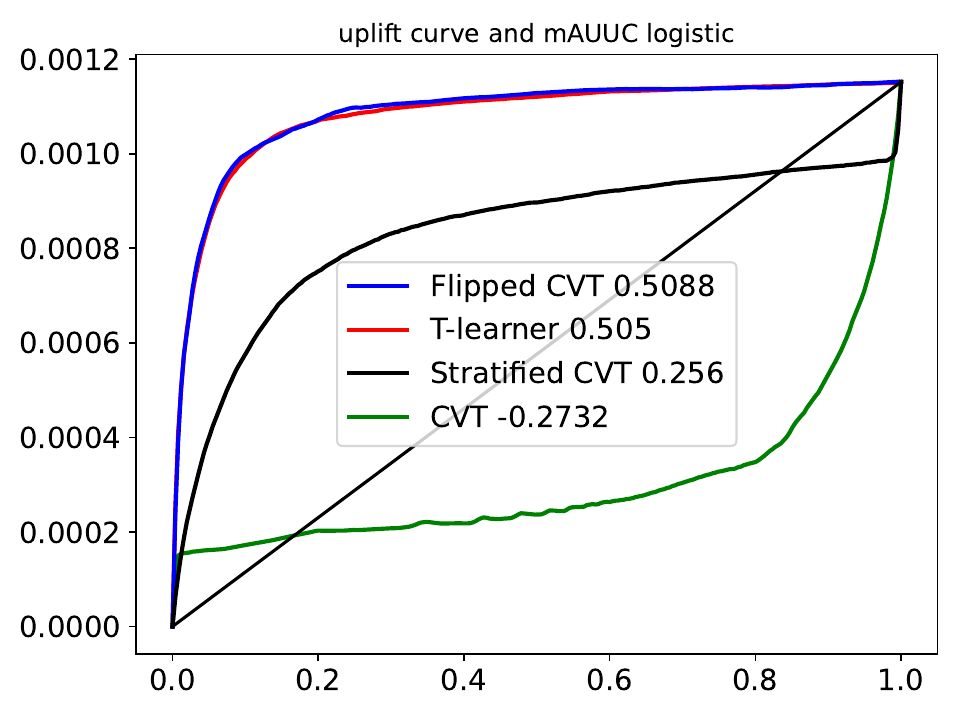}
\includegraphics[scale=\chartscale]{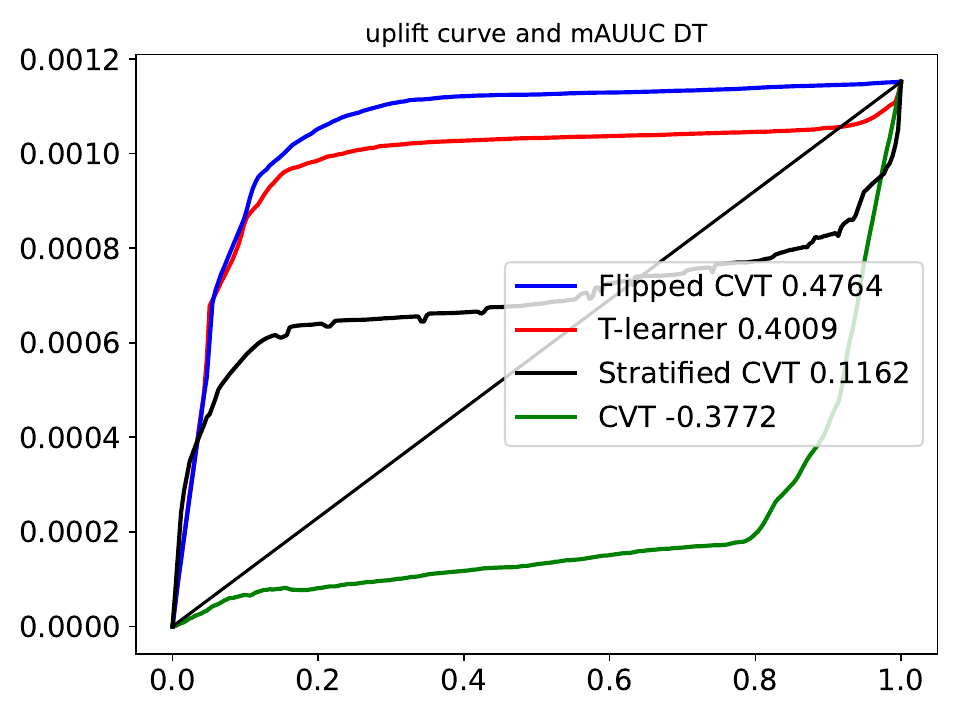}
\includegraphics[scale=\chartscale]{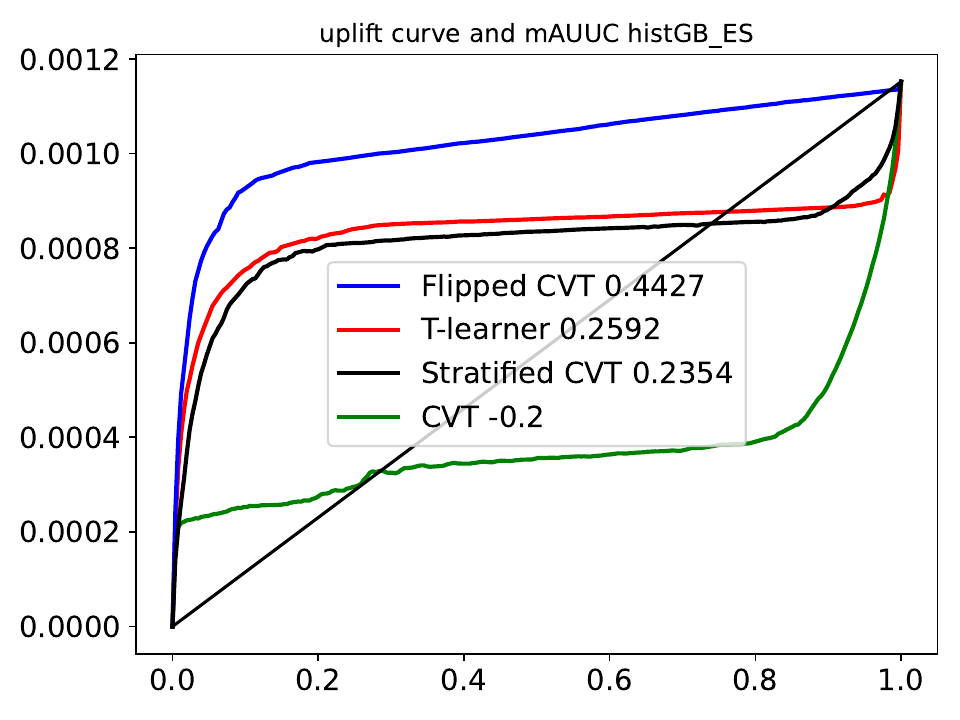}
\includegraphics[scale=\chartscale]{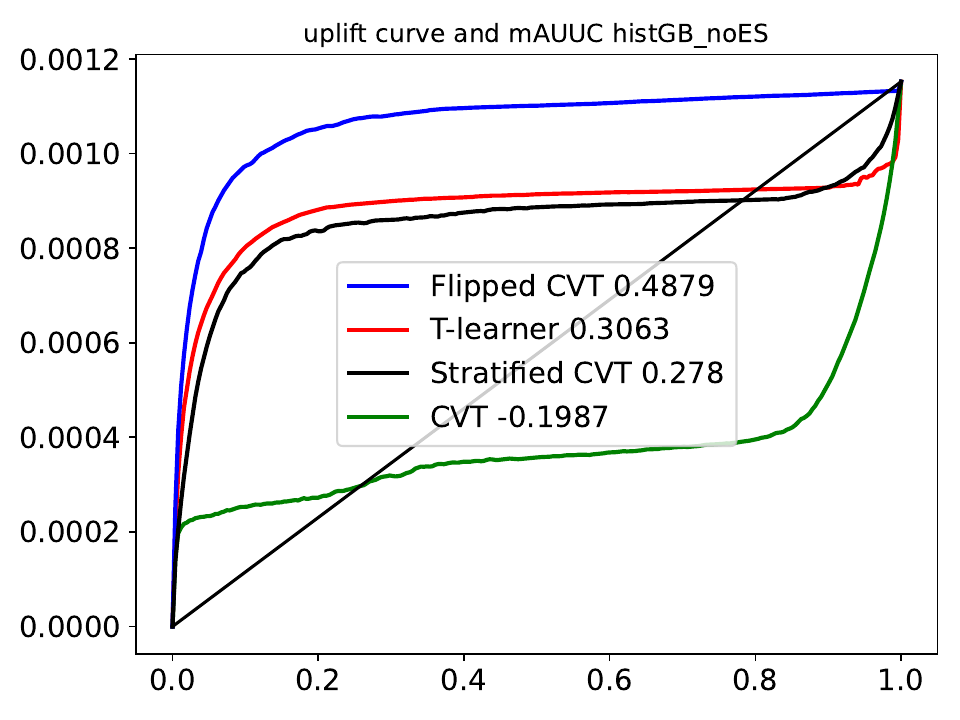}
\end{center}
\caption{Uplift curves and mAUUC for different CVT methods and T-learner on {\tt Criteo-conversion} dataset for logistic regression, decision tree and HistGradientboosting used as base learners}\label{fig:Criteo_C} 
\end{figure}

Based on the table and the figures, we may conclude that the vanilla
{\tt CVT} model performs poorly and was {\em always} outperformed by the
{\tt T-learner}.  This is most apparent for the {\tt Criteo-conversion}
dataset where the method fails to correctly predict treatment
heterogeneity regardless of the choice of the base learner.  However,
both class imbalance correction methods visibly improved its
performance, demonstrating that it is significantly affected by class
imbalance.

The experiments clearly demonstrate the advantages of the class
flipping ({\tt Flipped CVT}) method introduced in this paper.  It {\em
  always} improves the performance of the CVT method and outperforms
the Stratified Undersampling of~\cite{NybergDMKD} in all cases except
four (see Figures~\ref{fig:hillstrom_W_C},~\ref{fig:Gusto}
and~\ref{fig:starbucks} in the Appendix).  The T-learner remains a
strong contender, but is often outperformed by the proposed {\tt
  Flipped CVT} method.

Notice also that those performance gains were observed even for base
learners such as decision trees for which class flipping didn't work
well in classification experiments shown in
Section~\ref{sec:standard-classif}.

\section{Conclusions}\label{sec:conc}

The paper introduced a novel class imbalance correction method for
uplift modeling/HTE estimation based on flipping the class label in
the majority class.  The approach has the advantage that conditional
class probabilities can be recovered through linear transformations
which guarantees causal model identifiability.

We have tested the approach on uplift meta-learners, where we found it
to help when logistic regression or gradient boosting was used as the
base learner.  It was less helpful for decision trees.  Overall it
performed comparably or better than approaches based on undersampling.
Results for the standard classification problem we mixed with the
proposed method performing best for logistic regression.

We separately addressed the CVT uplift algorithm where class flipping
leads to great improvements in performance and allows the {\tt CVT}
approach to compete with other meta-models.  Those performance gains
were observed even for base learners such as decision trees for which
class flipping didn't work well in classification experiments.
The CVT model is different from other models since the class
transformation it uses may lead to severe bias for unbalanced data.

Future research will include investigating the effects of class
flipping on evaluation criteria involving
fairness~\citep{Lo2025-fairness}.  We also plan deeper investigations
into performance of decision trees under class flipping.

\bibliography{sn-bibliography}%

\begin{appendices}

\section{Class flipping for the CVT model for different treatment/control majority classes}\label{sec:cvt-different-maj}

We now examine the possibility of achieving independence between
$\tilde{\breve{\mathcal{Y}}}$ and $\calW$ in the CVT model in the
unusual case when the majority classes in treatment and control groups
differ.  Suppose we flip class $1$ observations in the treatment group
and class $0$ in controls, i.e.~$k^T_0=k^C_1=1$ and $k^T_1=k^C_0=k<1$.
The last equality is necessary for model identifiability, see
Equation~\ref{eq:imbal-0-1}.  Equation~\ref{eq:indep-cvt} becomes
\begin{align*}
k\mathbb{P}^*({\mathcal{Y}}=1|\calW=T) &= k\mathbb{P}^*({\mathcal{Y}}=0|\calW=C),
\end{align*}
which is satisfiable only in the special case when
$\mathbb{P}^*(\mathcal{Y}=1|\calW=T)=\mathbb{P}^*(\mathcal{Y}=0|\calW=C)$,
which means that independence of $\tilde{\breve{\mathcal{Y}}}$ and
$\calW$ cannot, in general, be achieved.

One may now ask which value of $k$ minimizes the dependence according
to some measure.  We chose the absolute value of the
log-odds-ratio~\citep{agresti:categorical} given by
\begin{align*}
L(k)=\left|\log\frac{\frac{\mathbb{P}^*(\tilde{\breve{\mathcal{Y}}}=1|\calW=T)}{\mathbb{P}^*(\tilde{\breve{\mathcal{Y}}}=0|\calW=T)}}{\frac{\mathbb{P}^*(\tilde{\breve{\mathcal{Y}}}=1|\calW=C)}{\mathbb{P}^*(\tilde{\breve{\mathcal{Y}}}=0|\calW=C)}}\right| = \left|\log\frac{\frac{\mathbb{P}^*(\breve{\mathcal{Y}}=1|\calW=T)}{\mathbb{P}^*(\breve{\mathcal{Y}}=0|\calW=T)}}{\frac{\mathbb{P}^*(\breve{\mathcal{Y}}=0|\calW=C)}{\mathbb{P}^*(\breve{\mathcal{Y}}=1|\calW=C)}}\right|,
\end{align*}
which is equal to zero iff independence is achieved.  We will now find
the minimum of the function for $0<k<1$.

Recall that we assume  $k=k^T_1=k^C_0$, and denote $a=\mathbb{P}^*(\mathcal{Y}=1|\calW=T)$, $b=\mathbb{P}^*(\mathcal{Y}=0|\calW=C)$.
Now\begin{align*}
 L(k) & = \left|\log\frac{\frac{ka}{1-ka}}{\frac{kb}{1-kb}}\right| = \left|\log\left(\frac{a}{1-ka}\right)-\log\left(\frac{b}{1-kb}\right) \right|,
\end{align*}
and its derivative:
\begin{align*}
\frac{d}{d k}L(k) & = \sign\left\{\log\frac{\frac{a}{1-ka}}{\frac{b}{1-kb}}\right\}\left(\frac{a}{1-ka}-\frac{b}{1-kb}\right).
\end{align*}
Notice that the signs of $\frac{a}{1-ka}-\frac{b}{1-kb}$ and
$\log\frac{\frac{a}{1-ka}}{\frac{b}{1-b}}$ are the same so $\frac{d}{d
  k}L(k)>0$ for $k\in(0,1)$.  The minimum is thus obtained when
$k\rightarrow 0$ which is not a usable solution since it would lead to
instances of only one class being present.

As a result, class flipping should not be used with the CVT method when
treatment and control majority classes differ.  Luckily in real life
these scenarios occur very rarely.

\section{Class flipping does not affect uplift trees using the $E$ criterion.}\label{sec:trees-E-unchanged}

The $E$ splitting criterion~\citep{Rzepakowski2011} is based on
squared Euclidean distance $d^2$ between treatment and control class
probability distributions.  Take two probability distributions
$P=(p,1-p)$, $Q=(q,1-q)$.  We have
\[
d^2(P, Q) = (p-q)^2 + ((1-p)-(1-q))^2
= 2(p-q)^2 
\]
so
\begin{align*}
d^2(\P(\breve{\mathcal{Y}}|\calW=T),\, &\P(\breve{\mathcal{Y}}|\calW=C)) = 2(\P(\breve{\mathcal{Y}}=0|\calW=T)- \P(\breve{\mathcal{Y}}=0|\calW=C))^2\\
&= 2(k\P({\mathcal{Y}}=0|\calW=T)- k\P({\mathcal{Y}}=0|\calW=C))^2\\
&=k^2d^2(\P({\mathcal{Y}}|\calW=T),\, \P({\mathcal{Y}}|\calW=C)).
\end{align*}
Since after deterministic class flipping
(Section~\ref{sec:flip-weighted}) the above equality holds exactly for
all conditional class distributions, all splitting criteria computed
during tree construction will be the same as for the original data,
just scaled by a constant $k^2$.  As a result, all selected tests will
be identical to those selected on unflipped data up to numerical
precision.

\section{Additional experiments}\label{sec:additional-exper}

In this section we show uplift curves not included in the main text.

\subsection{T-Learner.  Uplift trees}

This subsection shows uplift curves for T-learners and uplift trees for 
the {\tt Gusto} (Figures~\ref{fig:Gusto_T} and \ref{fig:Gusto_U}), {\tt Hillstrom-conversion} (Figures~\ref{fig:hillstrom_W_C_T} and \ref{fig:hillstrom_W_C_U}), {\tt Hillstrom-visit} (Figures~\ref{fig:hillstrom_W_V_T} and \ref{fig:hillstrom_W_V_U}), {\tt MarketingAB} (Figures~\ref{fig:marketingAB_T} and \ref{fig:marketingAB_U}), and {\tt Starbucks} (Figures~\ref{fig:starbucks_T} and \ref{fig:starbucks_U}) datasets.

\begin{figure}[H]
\begin{center}
\includegraphics[scale=\chartscale]{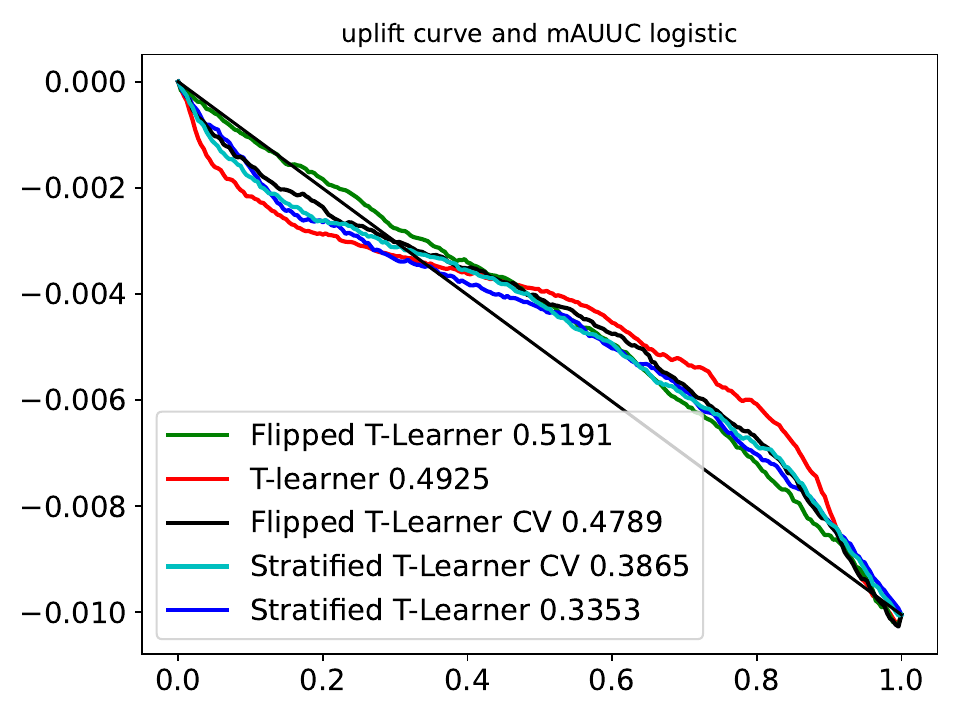}
\includegraphics[scale=\chartscale]{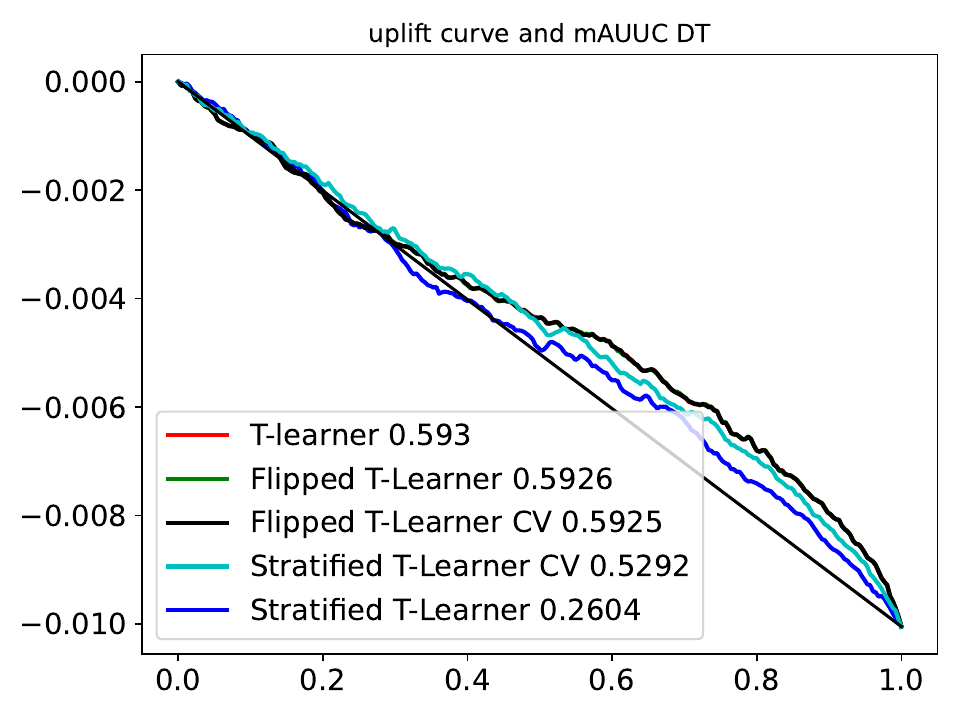}
\includegraphics[scale=\chartscale]{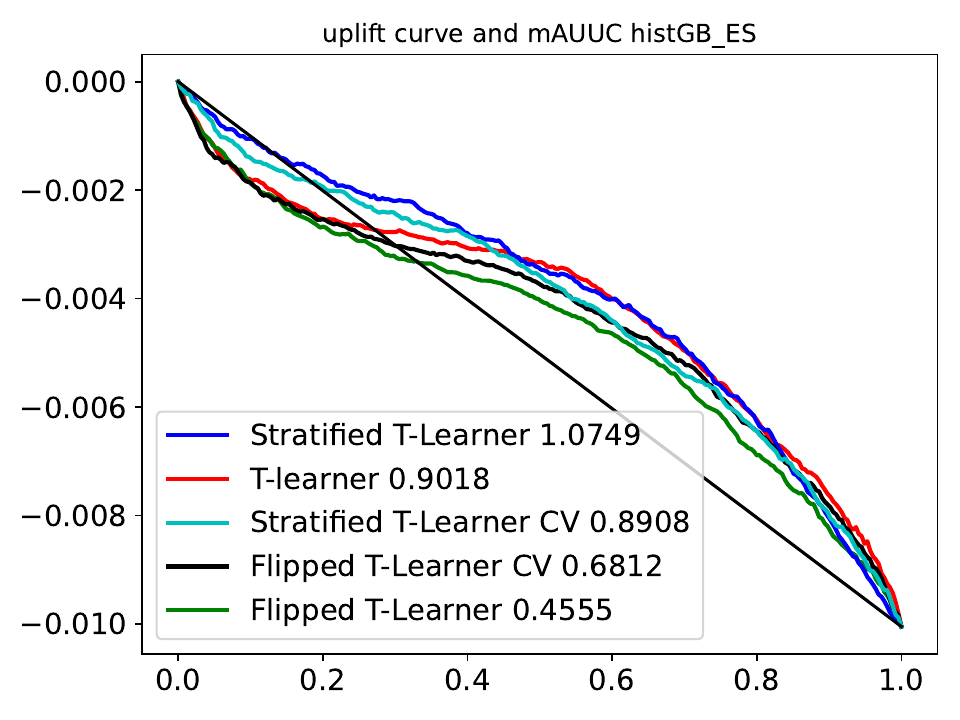}
\includegraphics[scale=\chartscale]{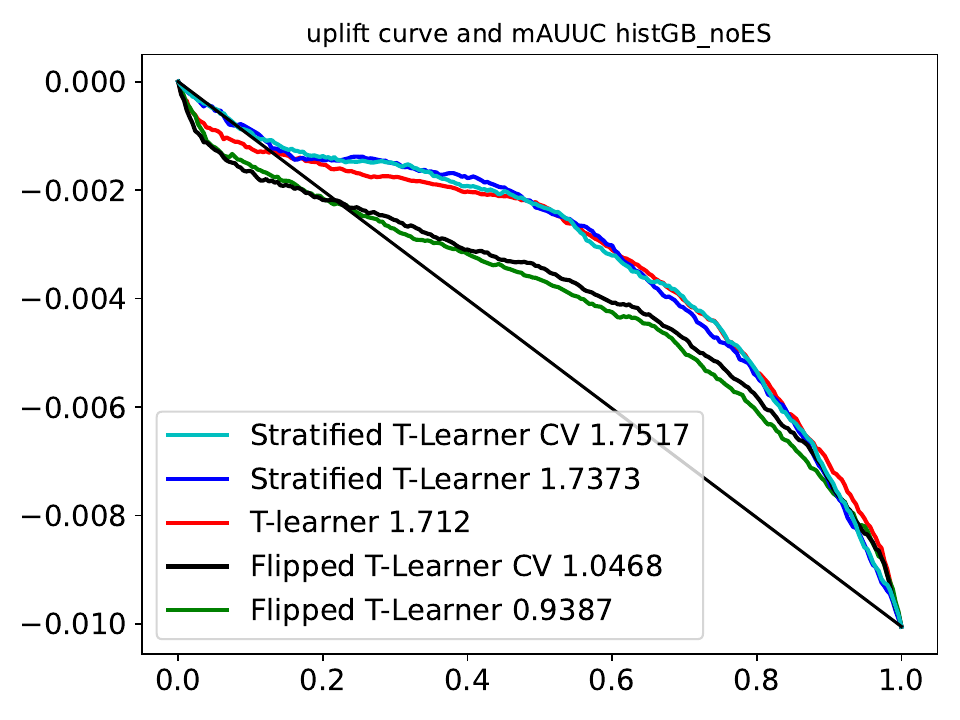}
\end{center}
\caption{Uplift curves and mAUUC for  T-learners on the {\tt Gusto} dataset for logistic regression, decision tree and gradient boosting used as base learners}\label{fig:Gusto_T} 
\end{figure}

\begin{figure}[H]
\begin{center}
\includegraphics[scale=\chartscale]{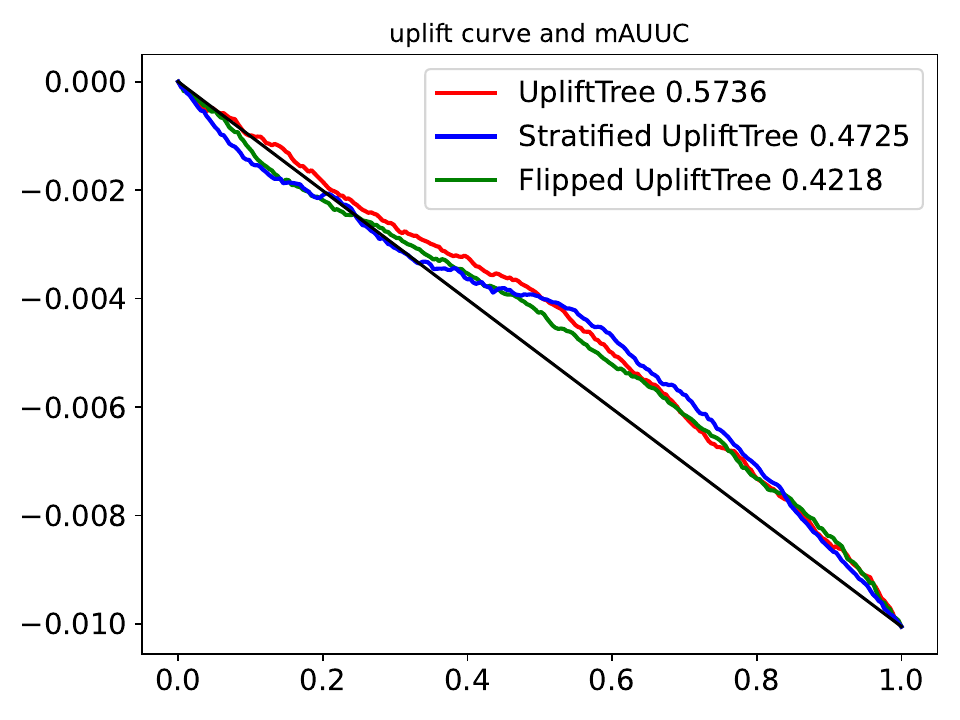}
\end{center}
\caption{Uplift curves and mAUUC for uplift trees on the {\tt Gusto} dataset}\label{fig:Gusto_U} 
\end{figure}

\begin{figure}[H]
\begin{center}
\includegraphics[scale=\chartscale]{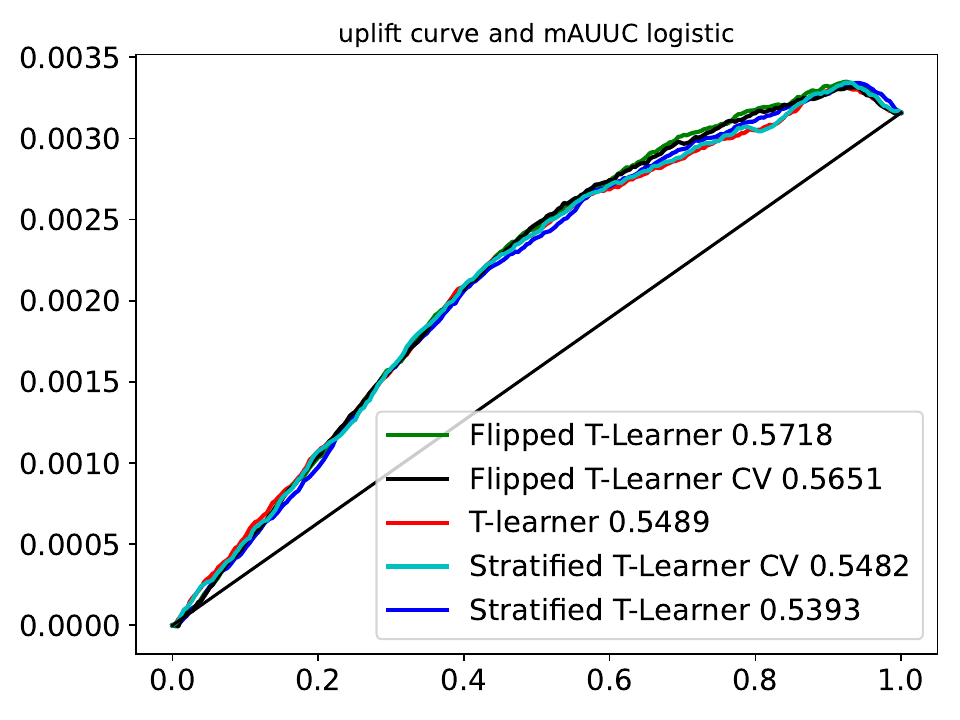}
\includegraphics[scale=\chartscale]{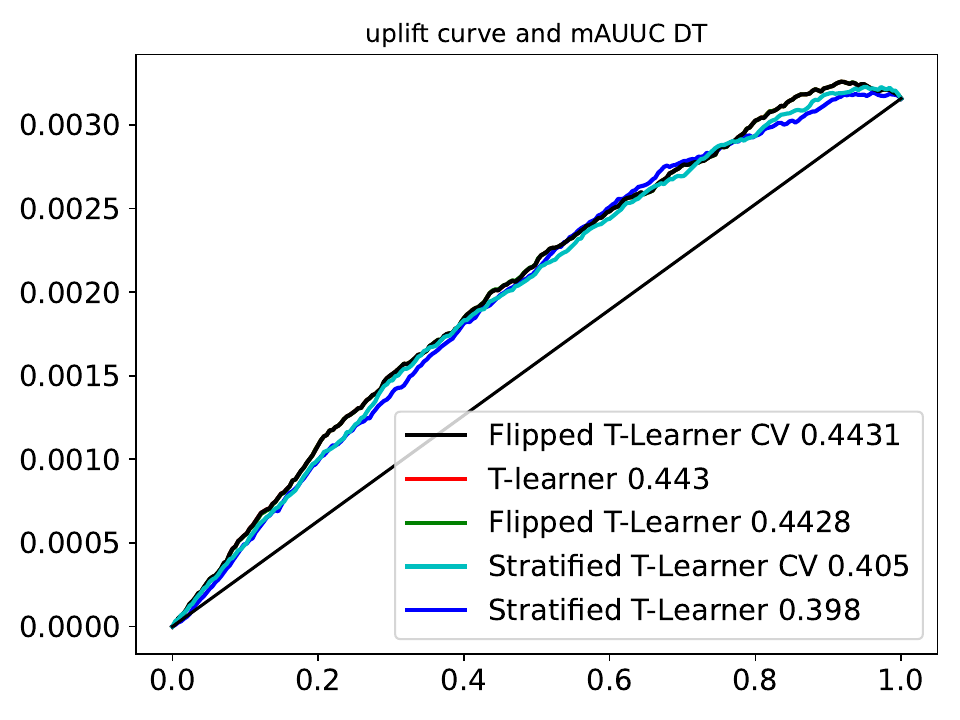}
\includegraphics[scale=\chartscale]{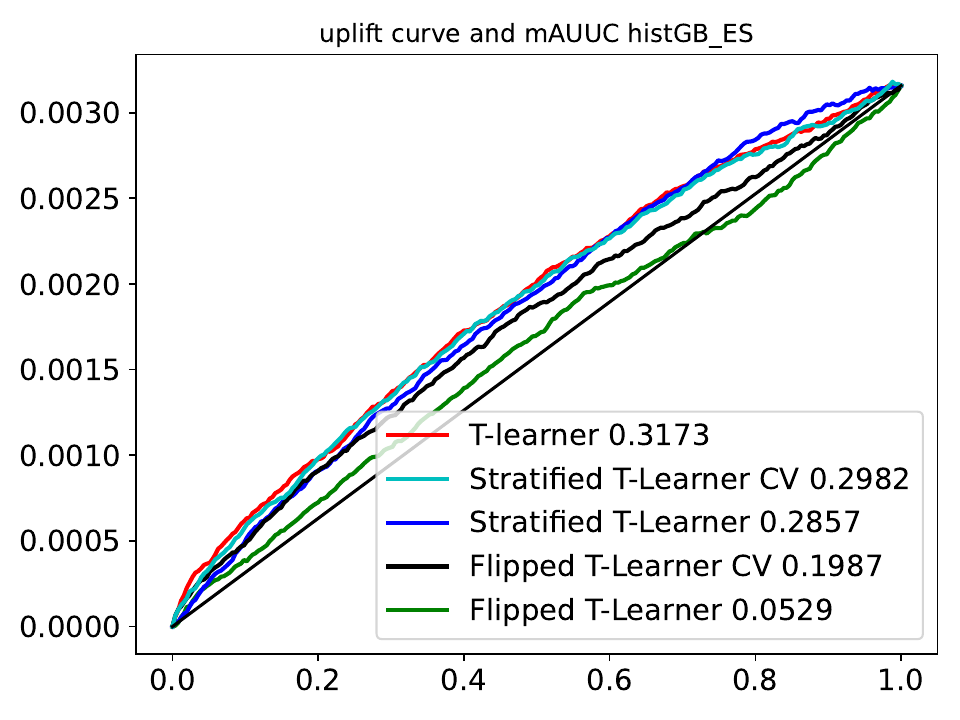}
\includegraphics[scale=\chartscale]{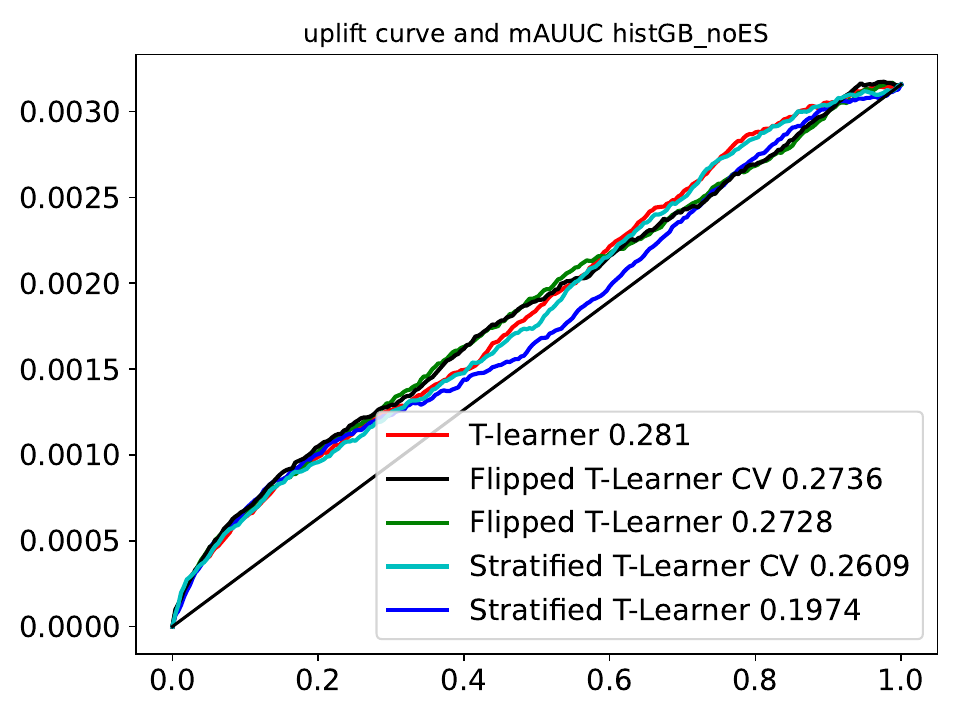}
\end{center}
\caption{Uplift curves and mAUUC for T-learners on the {\tt Hillstrom-conversion} dataset for logistic regression, decision tree and gradient boosting used as base learners}\label{fig:hillstrom_W_C_T} 
\end{figure}

\begin{figure}[H]
\begin{center}
\includegraphics[scale=\chartscale]{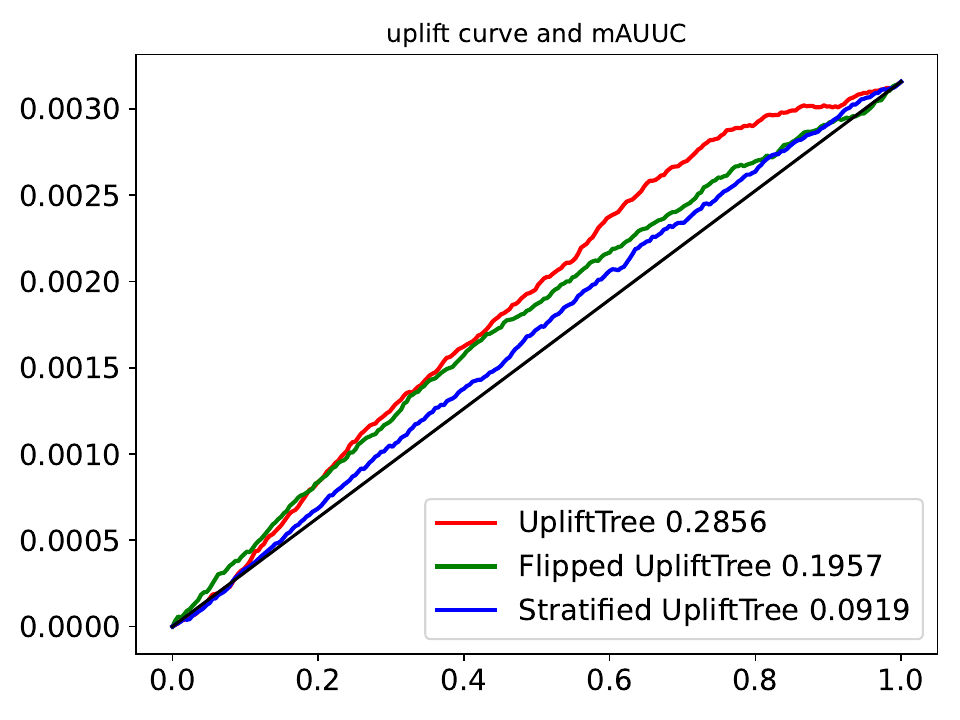}
\end{center}
\caption{Uplift curves and mAUUC for uplift trees on the {\tt Hillstrom-conversion} dataset}\label{fig:hillstrom_W_C_U} 
\end{figure}

\begin{figure}[H]
\begin{center}
\includegraphics[scale=\chartscale]{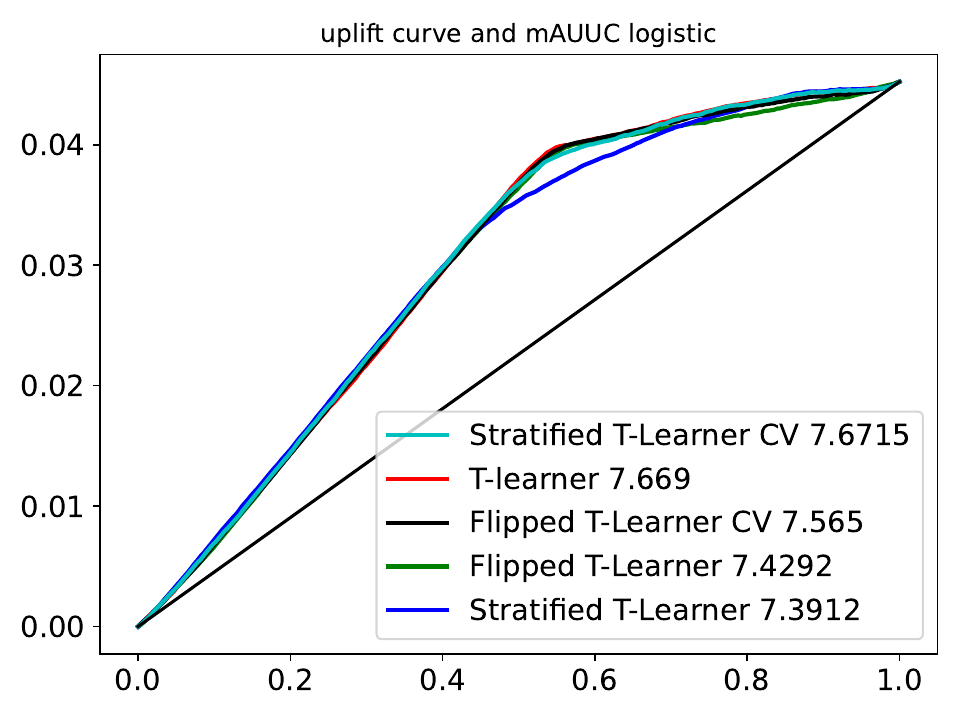}
\includegraphics[scale=\chartscale]{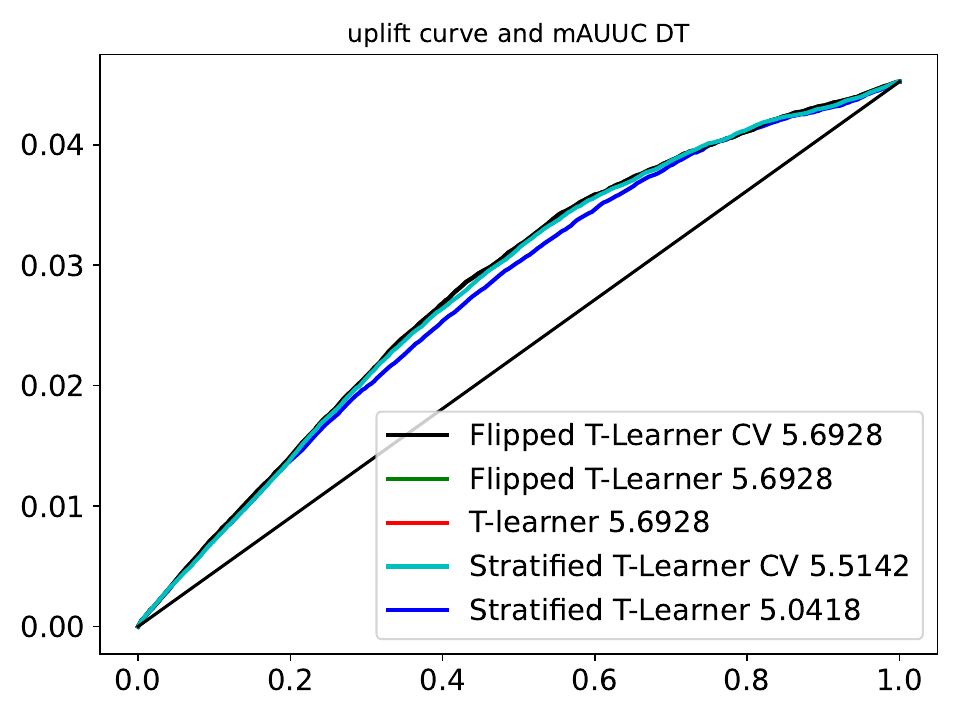}
\includegraphics[scale=\chartscale]{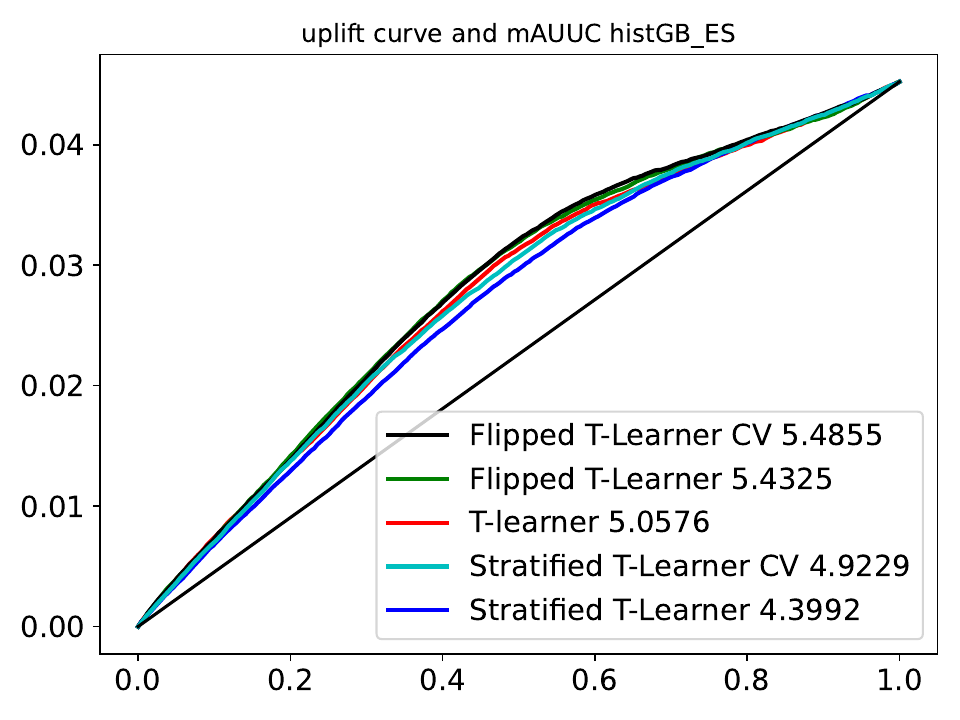}
\includegraphics[scale=\chartscale]{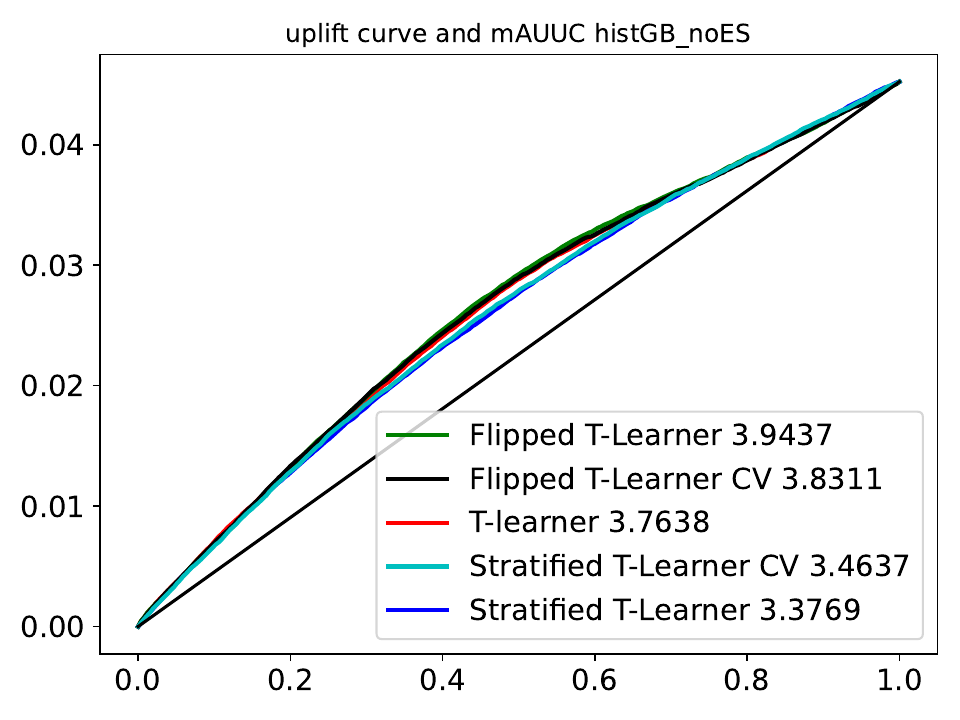}
\end{center}
\caption{Uplift curves and mAUUC for T-learners on the {\tt Hillstrom-visit} dataset for logistic regression, decision tree and gradient boosting used as base learners}\label{fig:hillstrom_W_V_T} 
\end{figure}

\begin{figure}[H]
\begin{center}
\includegraphics[scale=\chartscale]{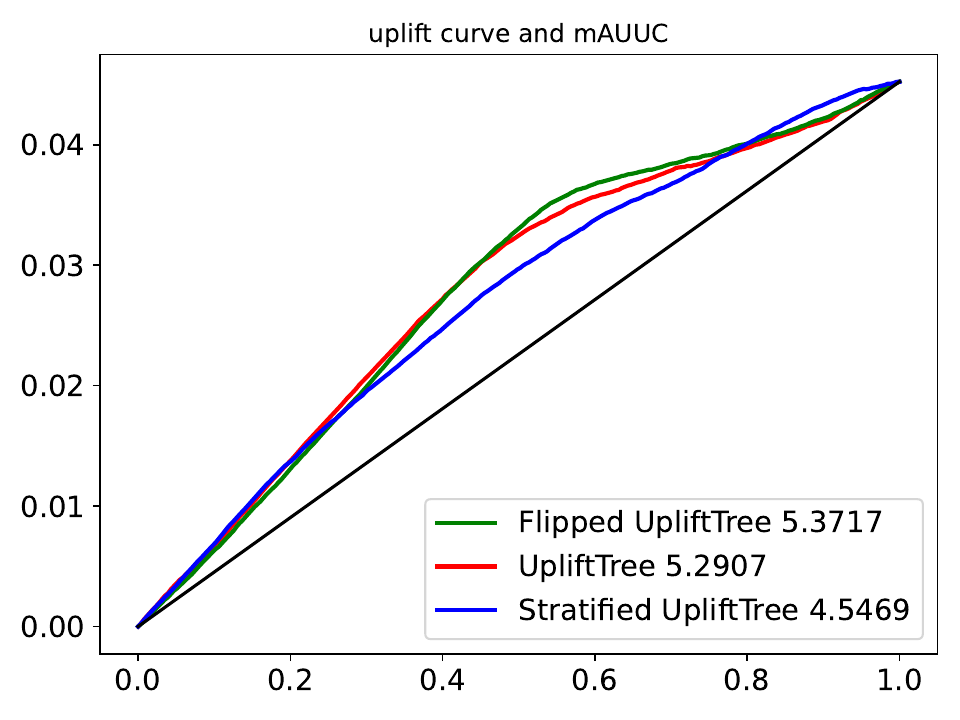}
\end{center}
\caption{Uplift curves and mAUUC for uplift trees on the {\tt Hillstrom-visit} dataset}\label{fig:hillstrom_W_V_U}
\end{figure}

\begin{figure}[H]
\begin{center}
\includegraphics[scale=\chartscale]{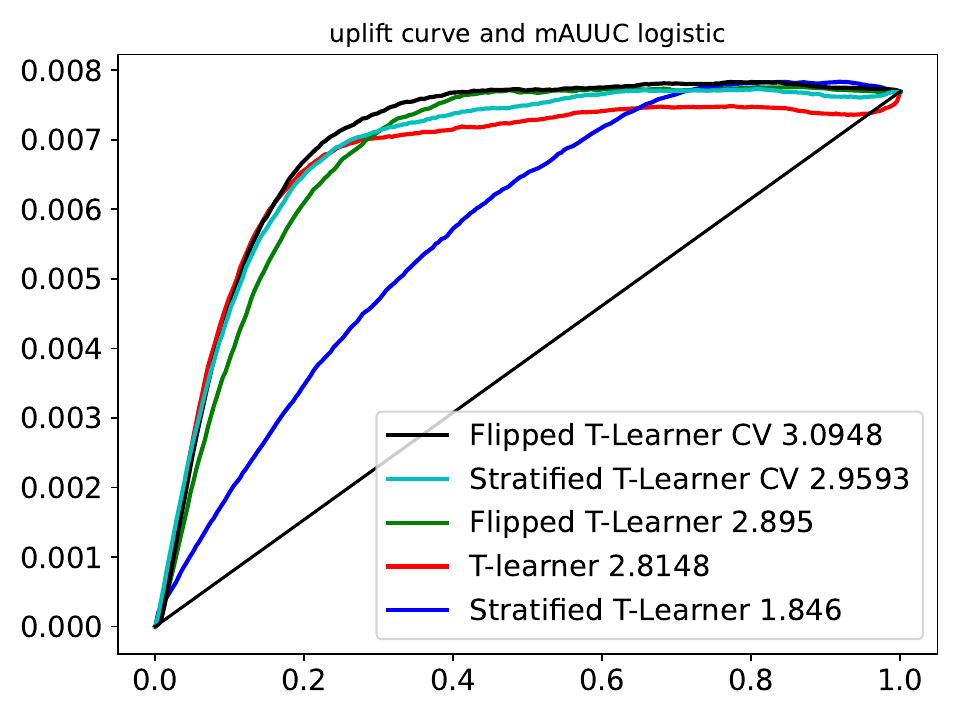}
\includegraphics[scale=\chartscale]{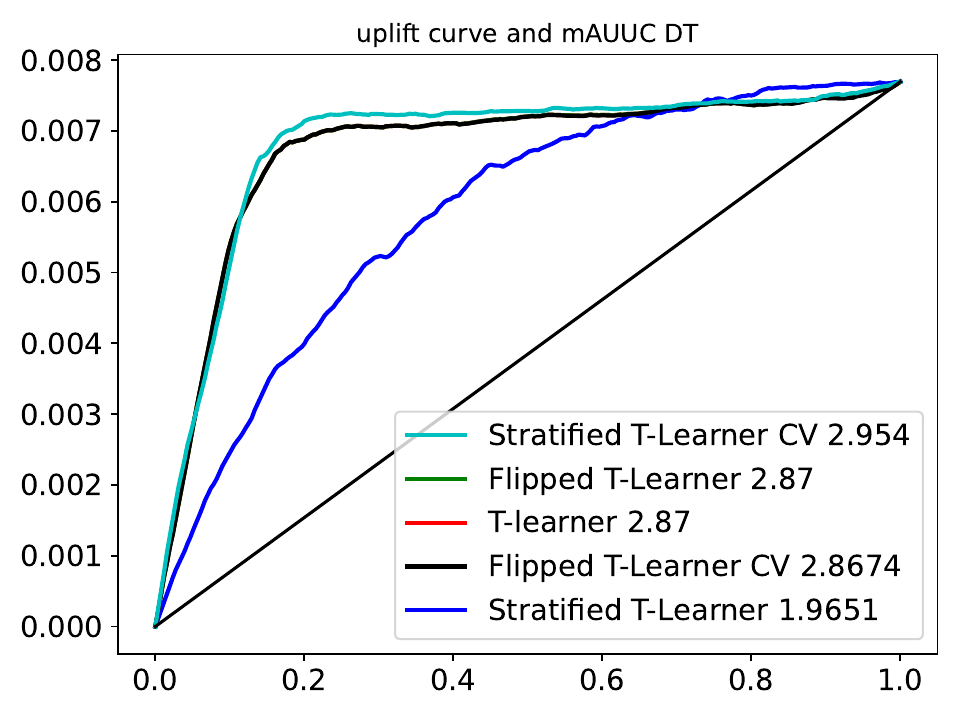}
\includegraphics[scale=\chartscale]{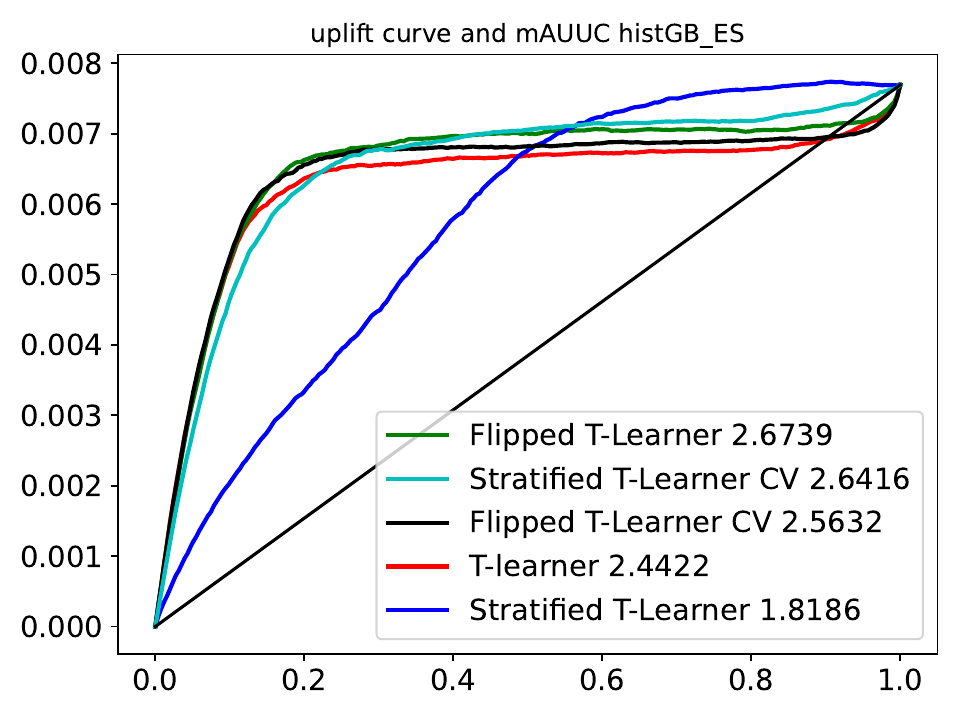}
\includegraphics[scale=\chartscale]{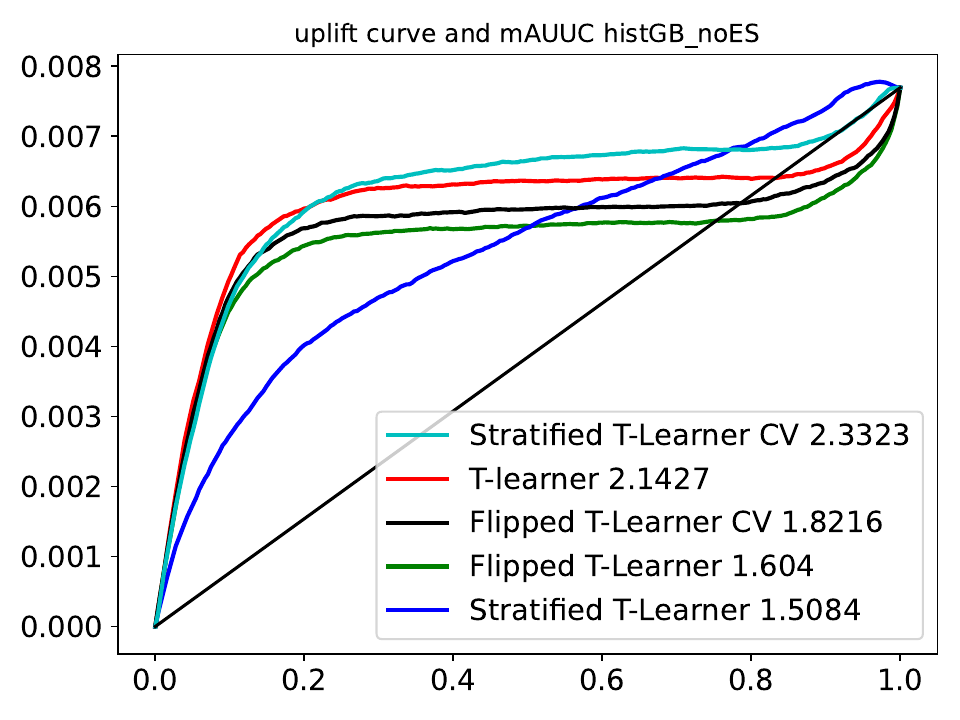}
\end{center}
\caption{Uplift curves and mAUUC for T-learners methods on the {\tt marketingAB} dataset for logistic regression, decision tree and gradient boosting used as base learners}\label{fig:marketingAB_T} 
\end{figure}

\begin{figure}[H]
\begin{center}
\includegraphics[scale=\chartscale]{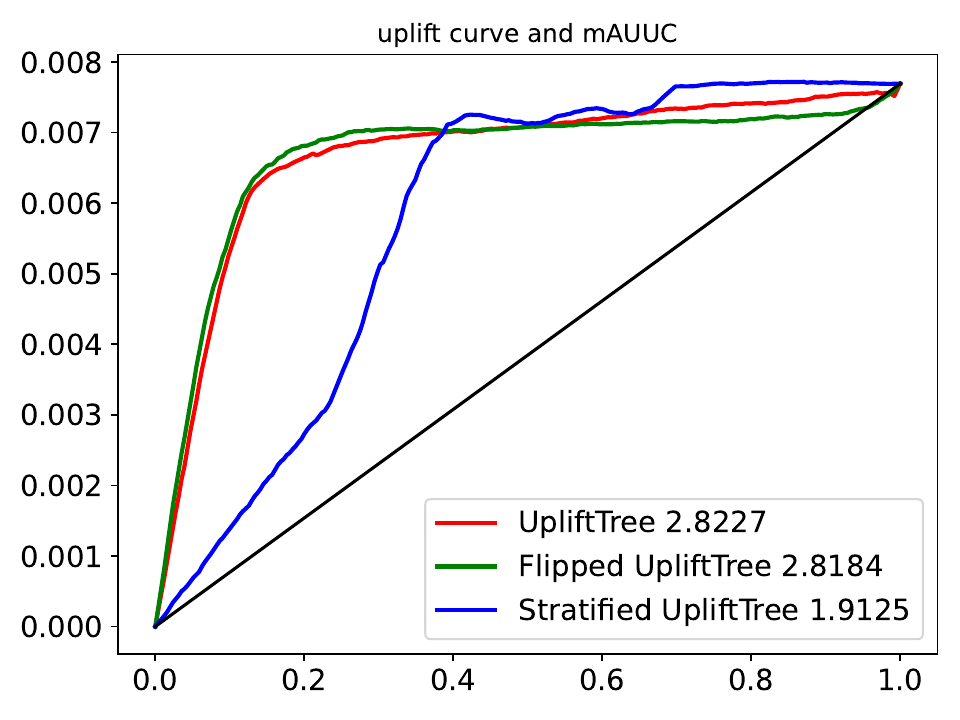}
\end{center}
\caption{Uplift curves and mAUUC for uplift trees on the {\tt marketingAB} dataset}\label{fig:marketingAB_U} 
\end{figure}

\begin{figure}[H]
\begin{center}
\includegraphics[scale=\chartscale]{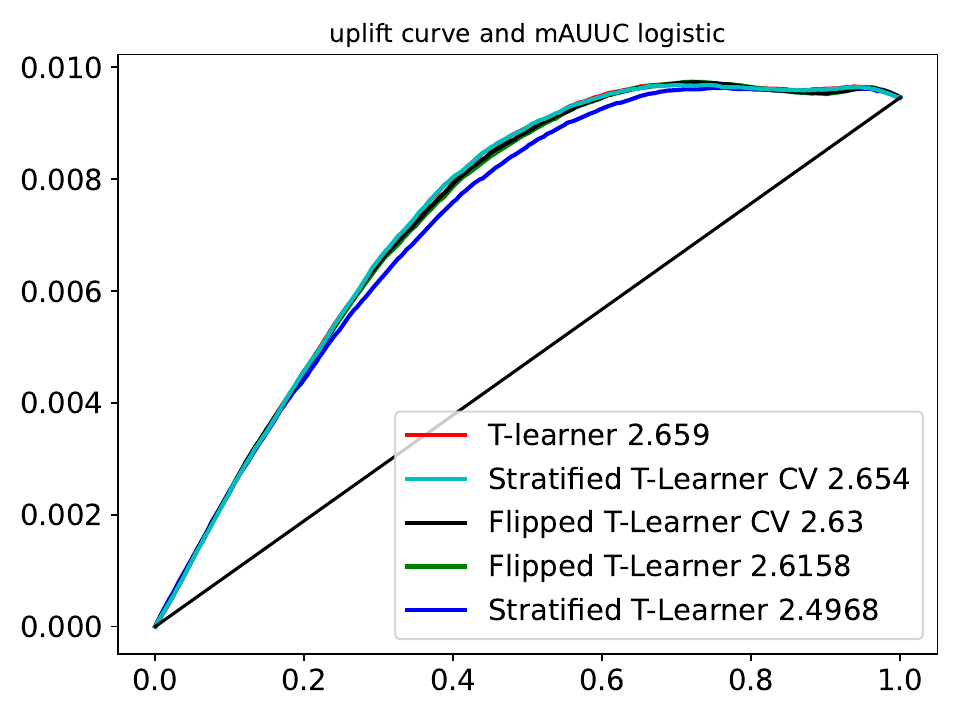}
\includegraphics[scale=\chartscale]{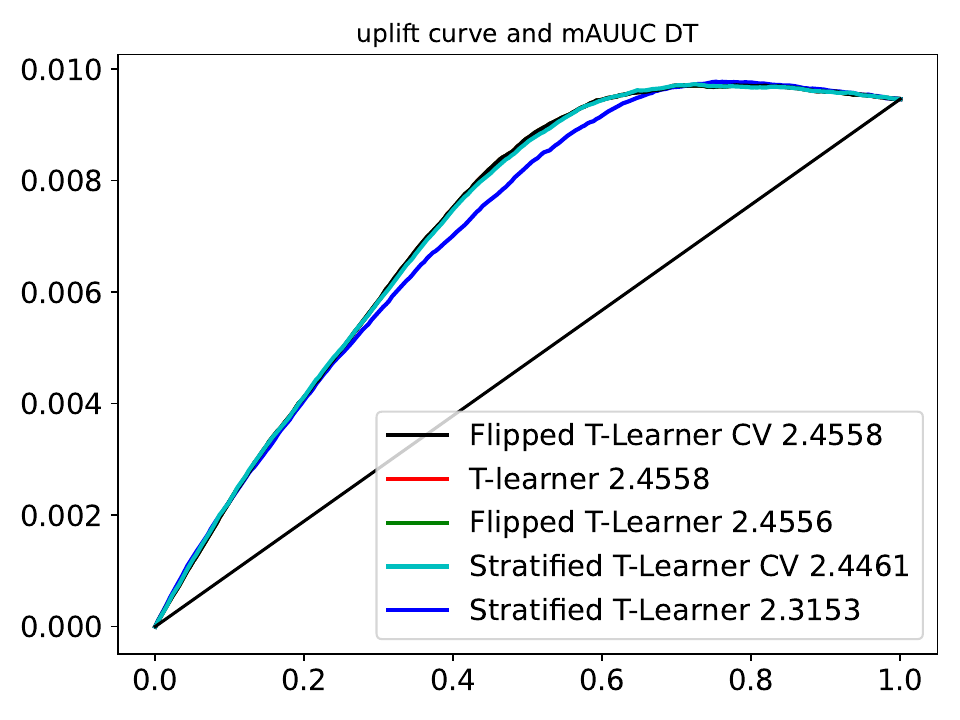}
\includegraphics[scale=\chartscale]{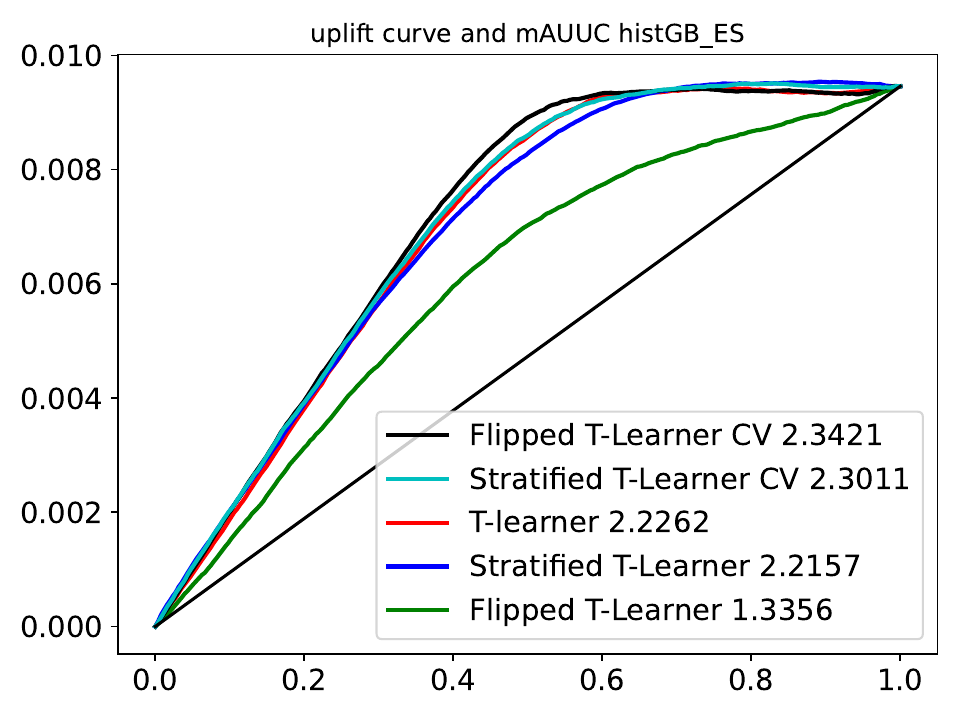}
\includegraphics[scale=\chartscale]{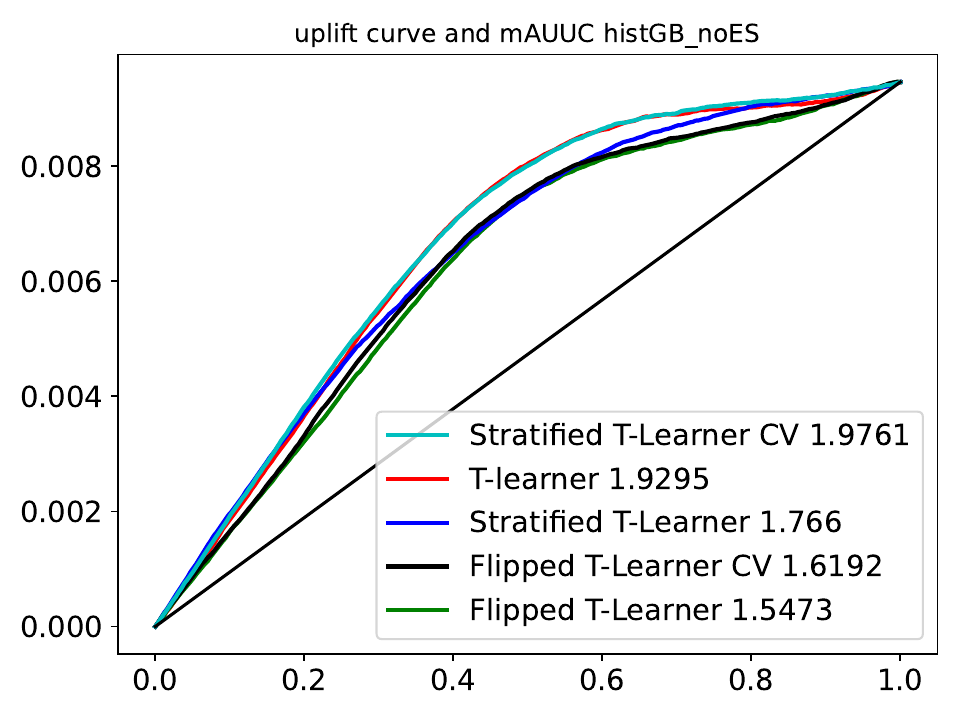}
\end{center}
\caption{Uplift curves and mAUUC for  T-learners on the {\tt Starbucks} dataset for logistic regression, decision tree and gradient boosting used as base learners}\label{fig:starbucks_T} 
\end{figure}

\begin{figure}[H]
\begin{center}
\includegraphics[scale=\chartscale]{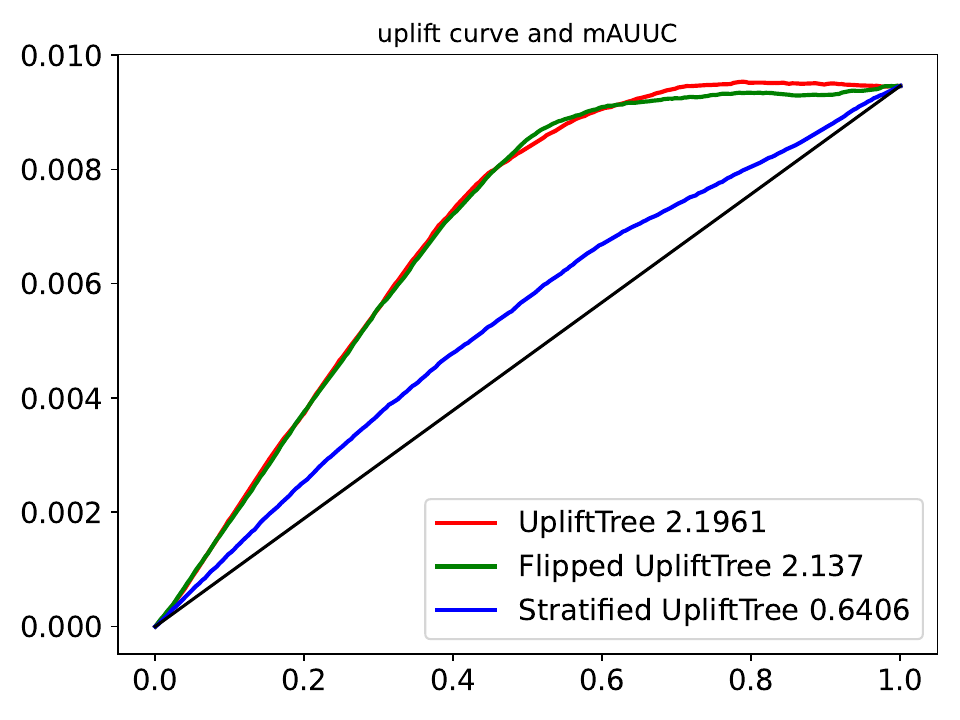}
\end{center}
\caption{Uplift curves and mAUUC for uplift trees on the {\tt Starbucks} dataset}\label{fig:starbucks_U} 
\end{figure}

\subsection{The CVT model}

This subsection shows uplift curves for CVT methods for the {\tt Gusto} (Figure~\ref{fig:Gusto}), {\tt Hillstrom-conversion} (Figure~\ref{fig:hillstrom_W_C}), {\tt Hillstrom-visit} (Figure~\ref{fig:hillstrom_W_V}), {\tt marketingAB} (Figure~\ref{fig:marketingAB}),  and {\tt Starbucks} (Figure~\ref{fig:starbucks}) datasets.

\begin{figure}[H]
\begin{center}
\includegraphics[scale=\chartscale]{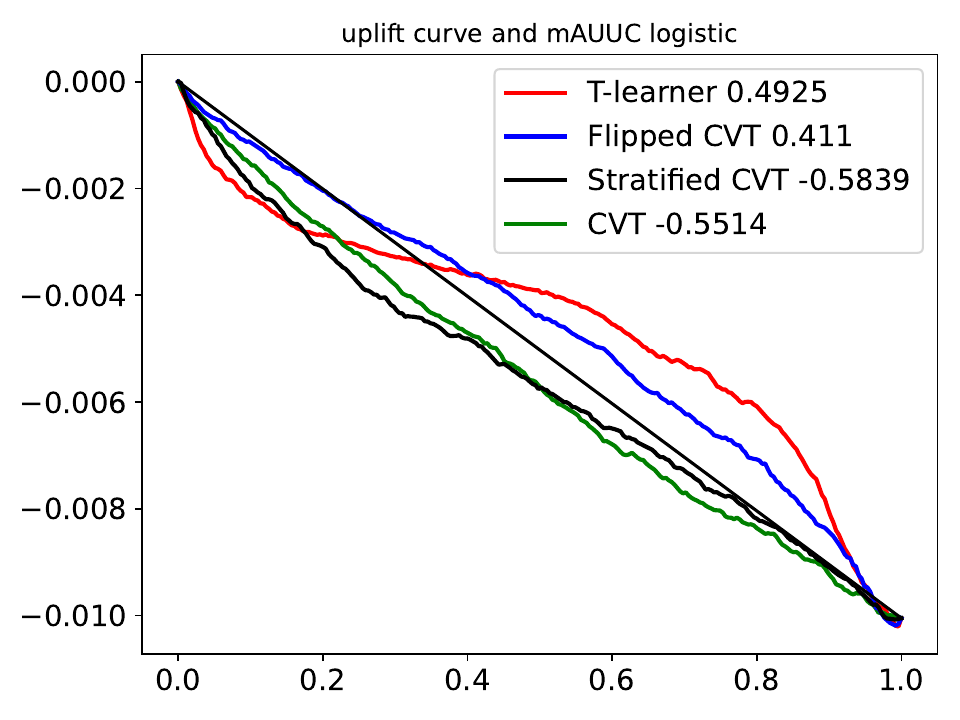}
\includegraphics[scale=\chartscale]{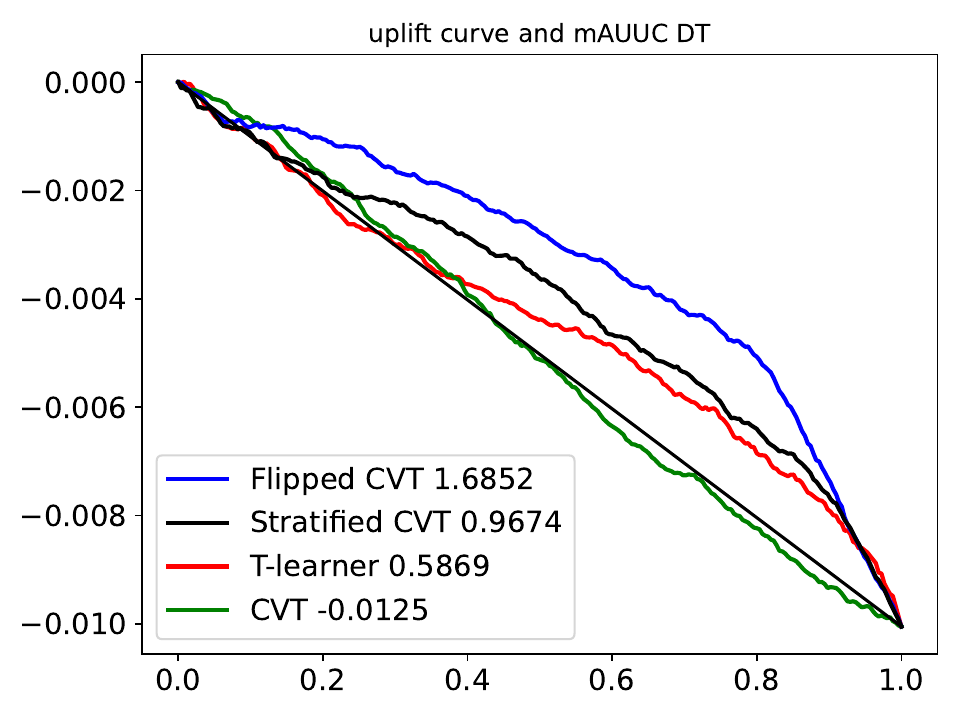}
\includegraphics[scale=\chartscale]{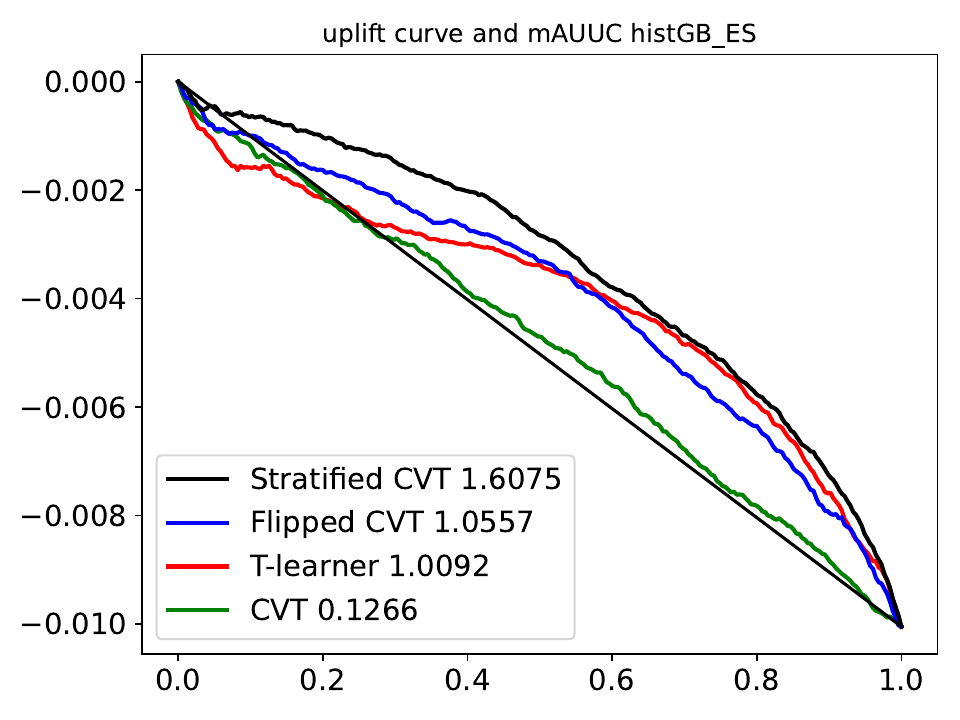}
\includegraphics[scale=\chartscale]{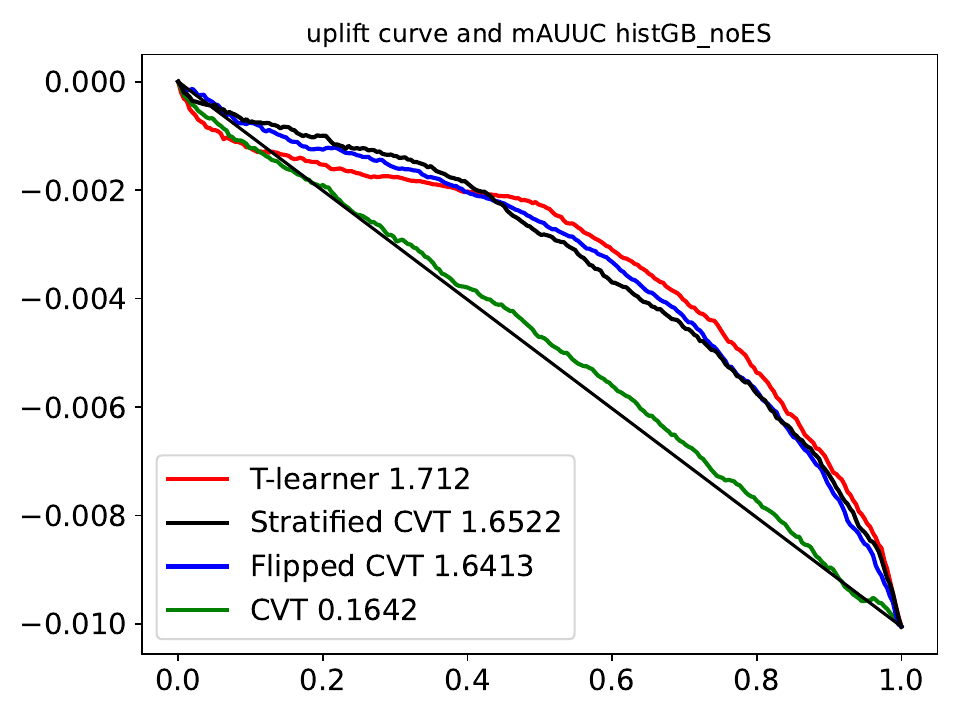}
\end{center}
\caption{Uplift curves and mAUUC for different CVT-learner methods and T-learner on {\tt Gusto} dataset for logistic regression, decision tree and HistGradientboosting used as base learners}\label{fig:Gusto} 
\end{figure}

\begin{figure}[H]
\begin{center}
\includegraphics[scale=\chartscale]{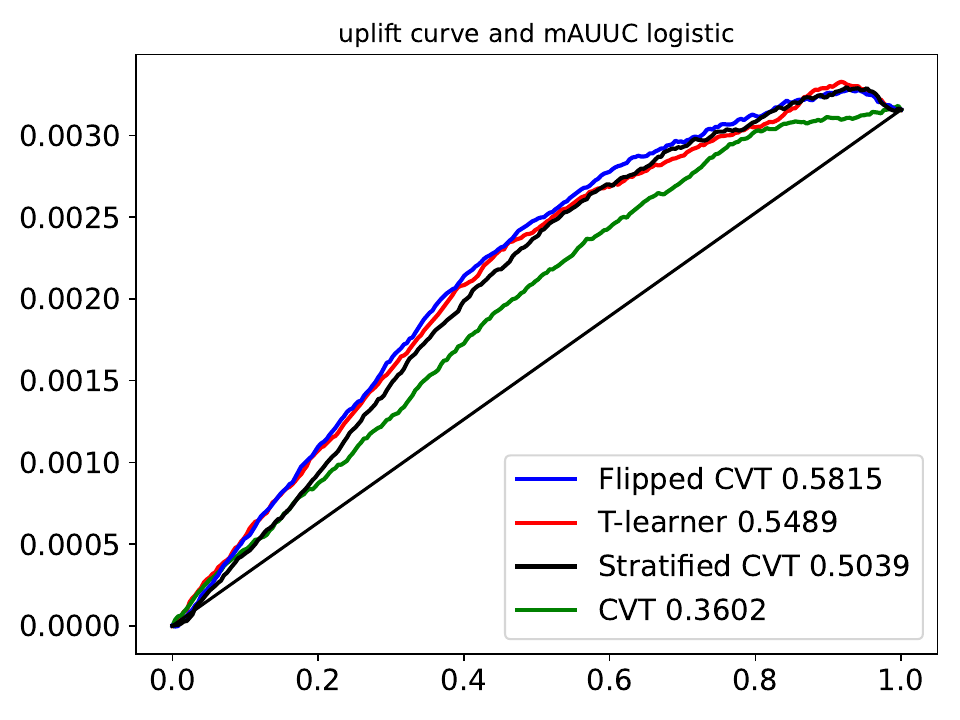}
\includegraphics[scale=\chartscale]{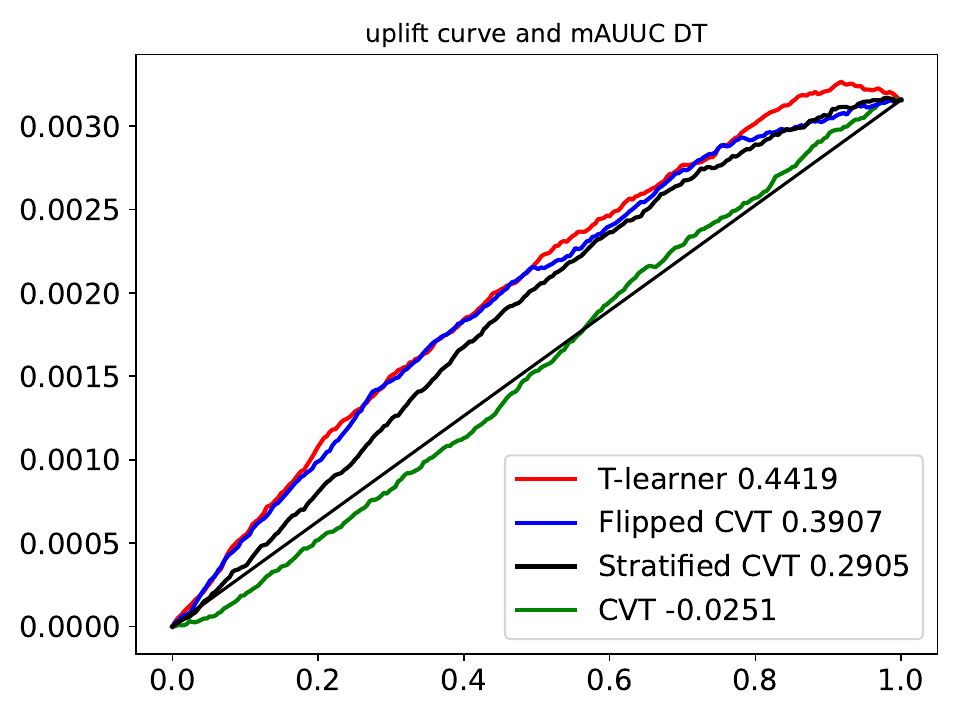}
\includegraphics[scale=\chartscale]{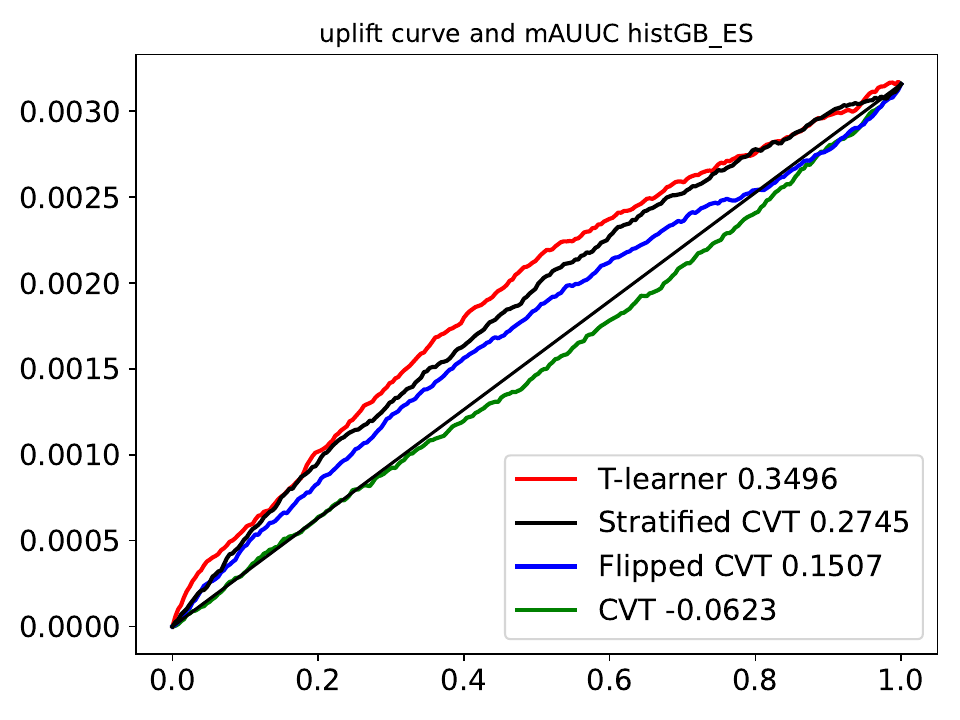}
\includegraphics[scale=\chartscale]{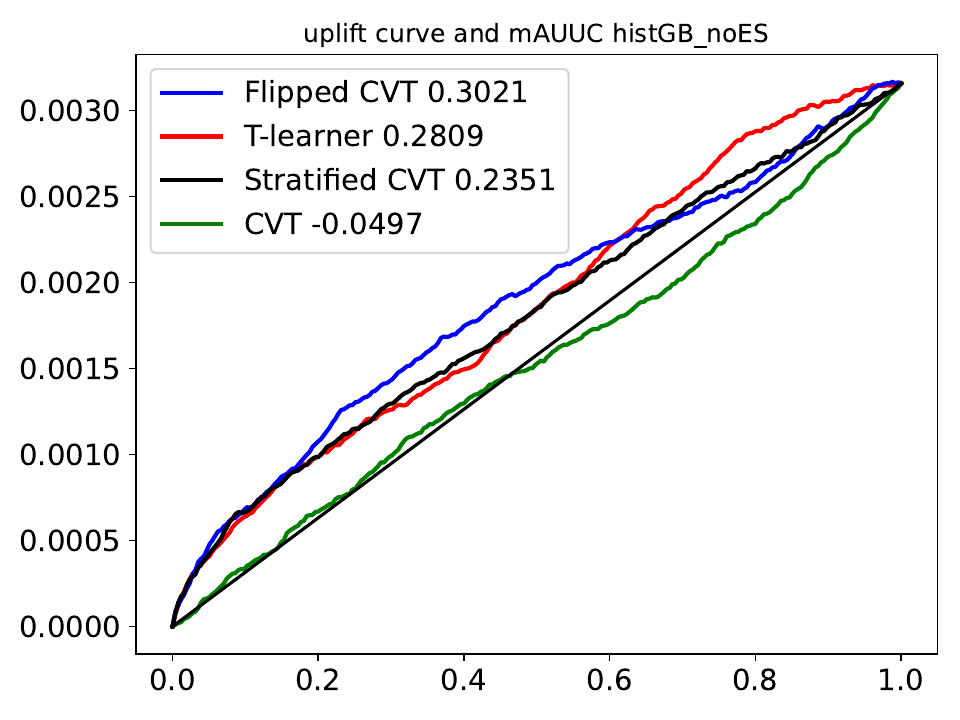}
\end{center}
\caption{Uplift curves and mAUUC for different CVT methods and T-learner on the {\tt Hillstrom-conversion} dataset for logistic regression, decision tree and gradient boosting used as base learners}\label{fig:hillstrom_W_C} 
\end{figure}

\begin{figure}[H]
\begin{center}
\includegraphics[scale=\chartscale]{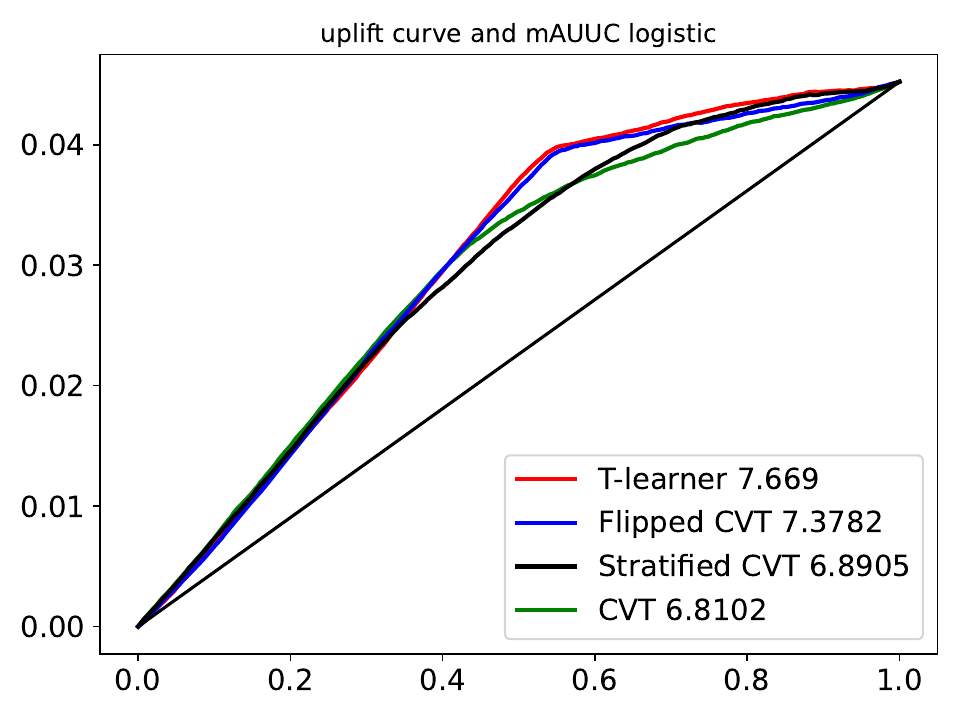}
\includegraphics[scale=\chartscale]{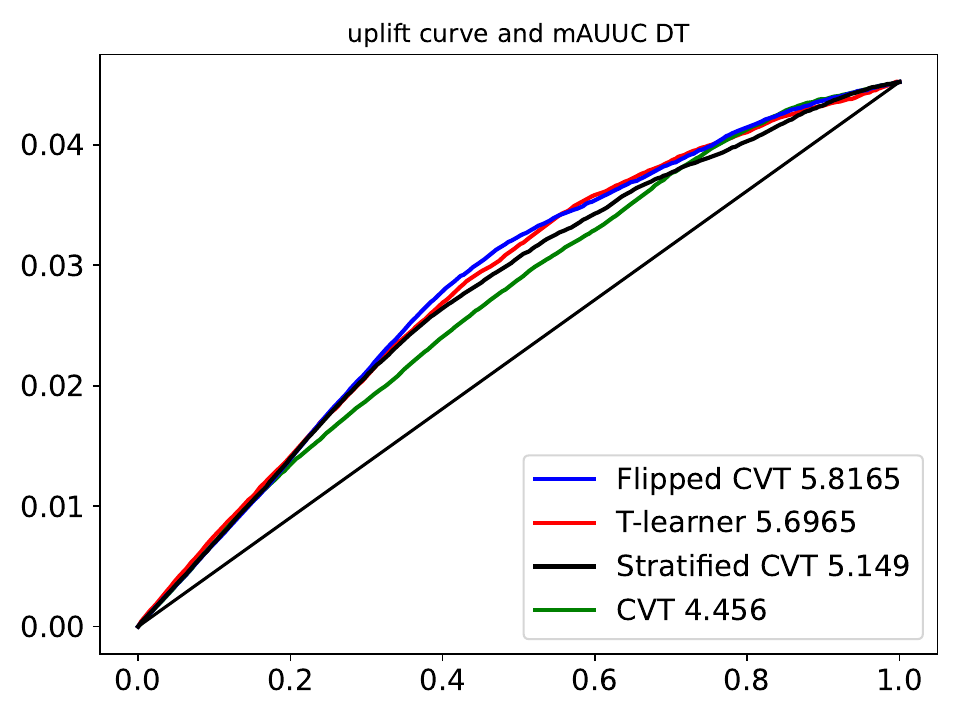}
\includegraphics[scale=\chartscale]{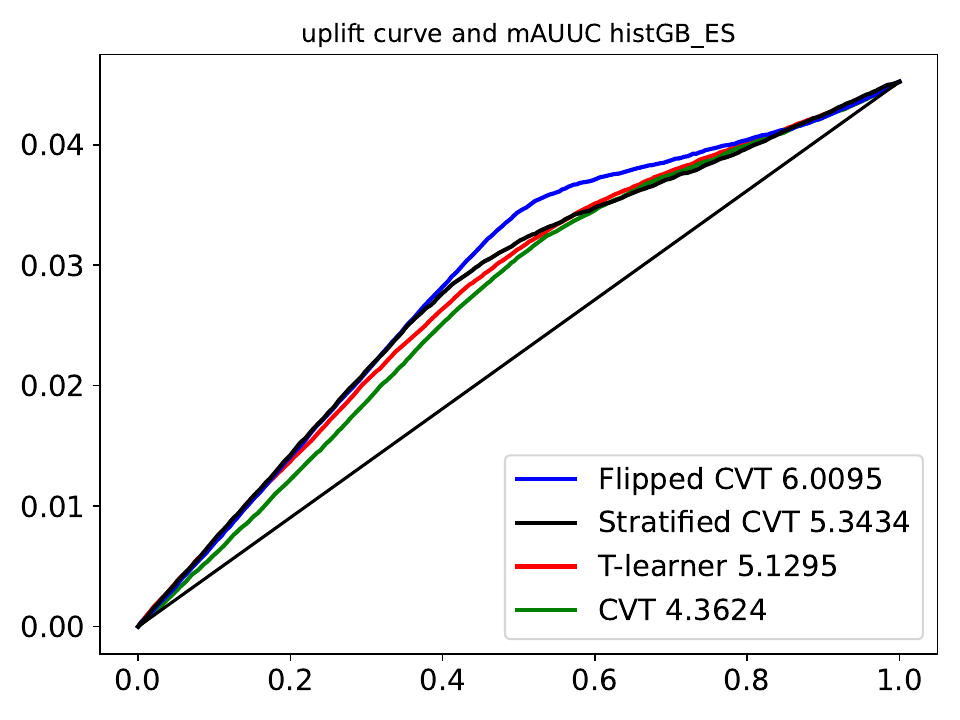}
\includegraphics[scale=\chartscale]{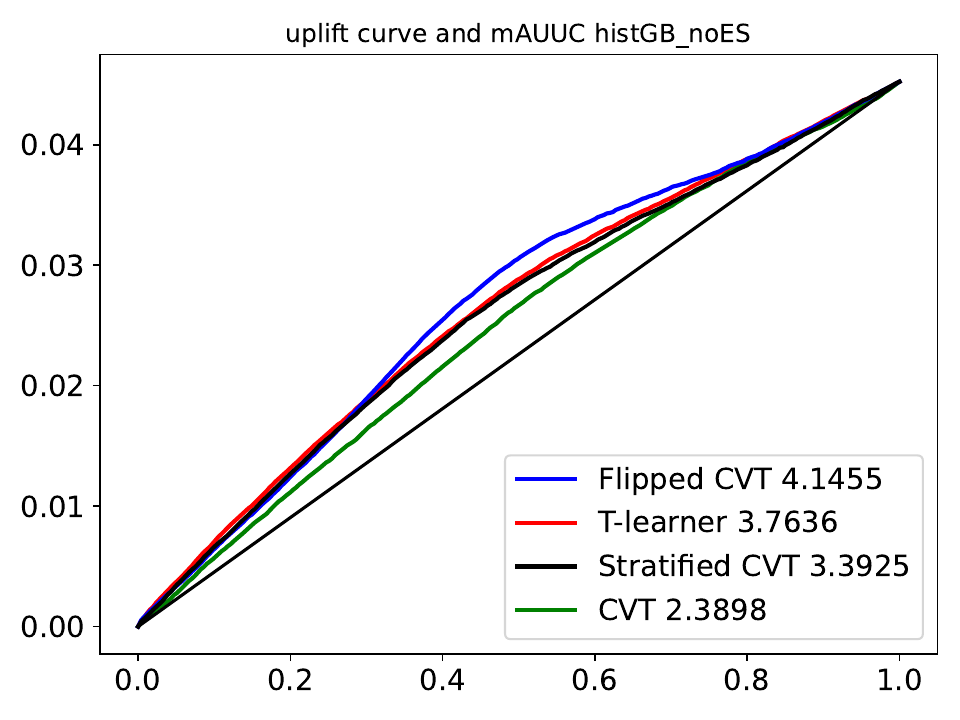}
\end{center}
\caption{Uplift curves and mAUUC for different CVT methods and T-learner on the {\tt Hillstrom-visit} dataset for logistic regression, decision tree and gradient boosting used as base learners}\label{fig:hillstrom_W_V} 
\end{figure}

\begin{figure}[H]
\begin{center}
\includegraphics[scale=\chartscale]{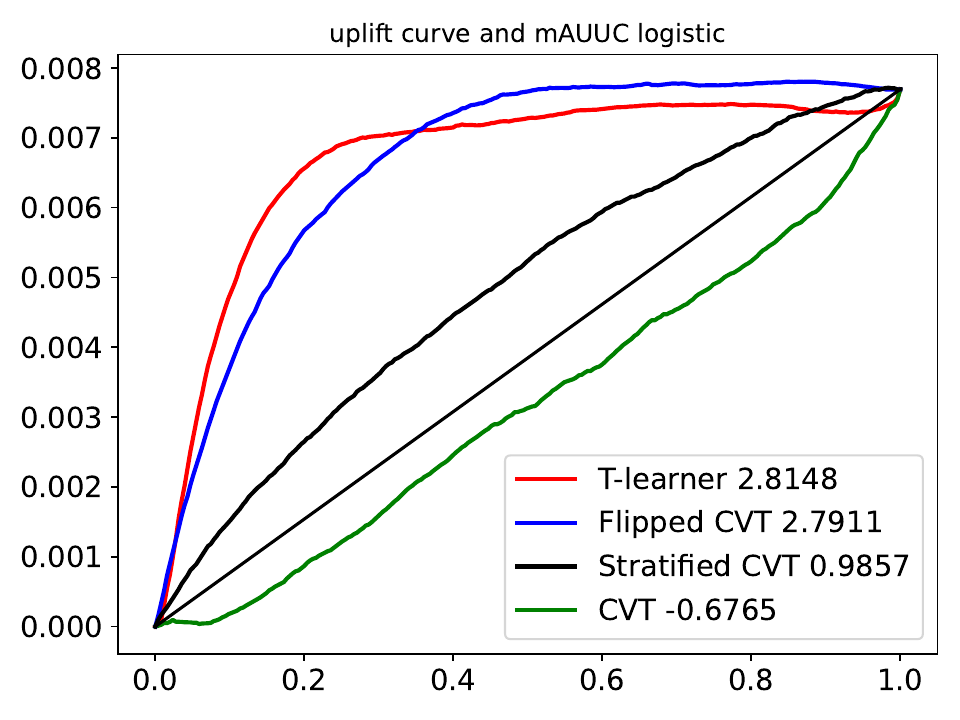}
\includegraphics[scale=\chartscale]{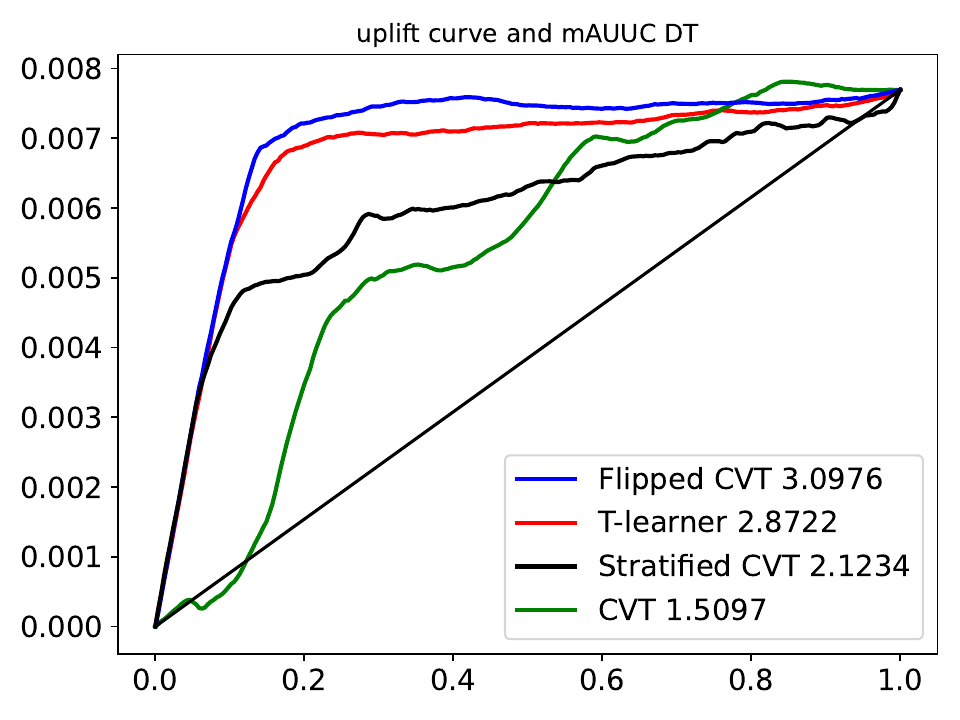}
\includegraphics[scale=\chartscale]{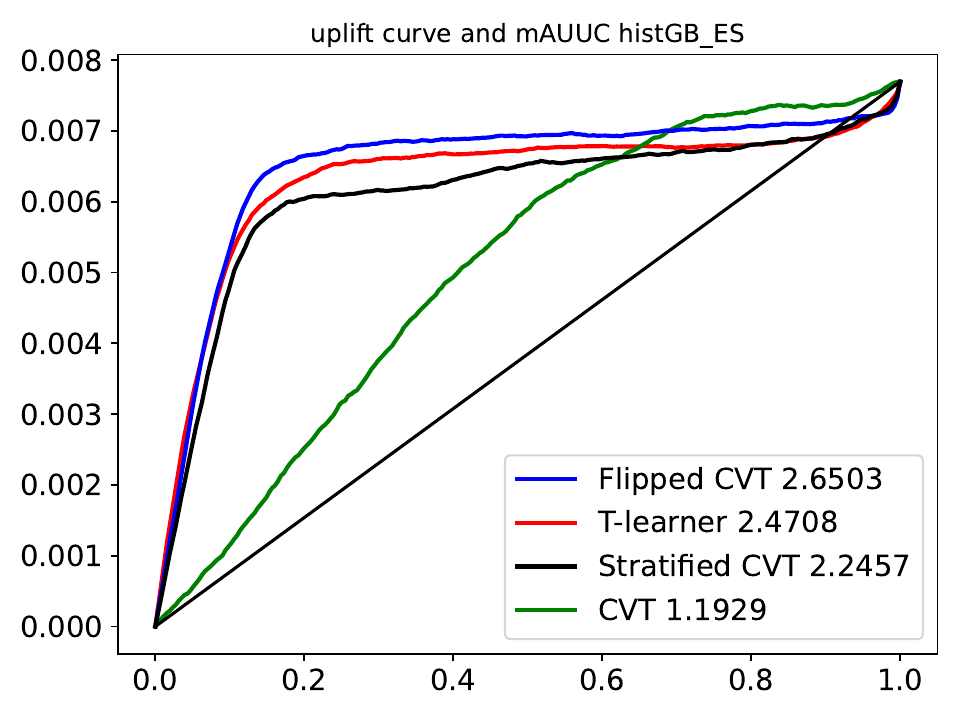}
\includegraphics[scale=\chartscale]{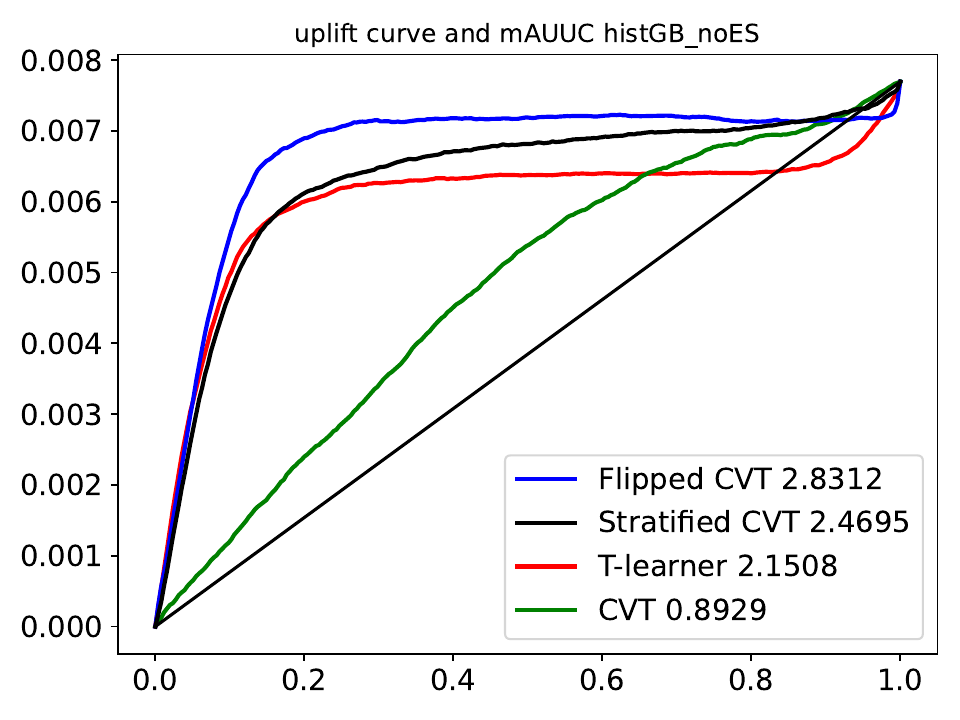}
\end{center}
\caption{Uplift curves and mAUUC for different CVT-learner methods and T-learner on the {\tt marketingAB} dataset for logistic regression, decision tree and gradient boosting used as base learners}\label{fig:marketingAB} 
\end{figure}

\begin{figure}[H]
\begin{center}
\includegraphics[scale=\chartscale]{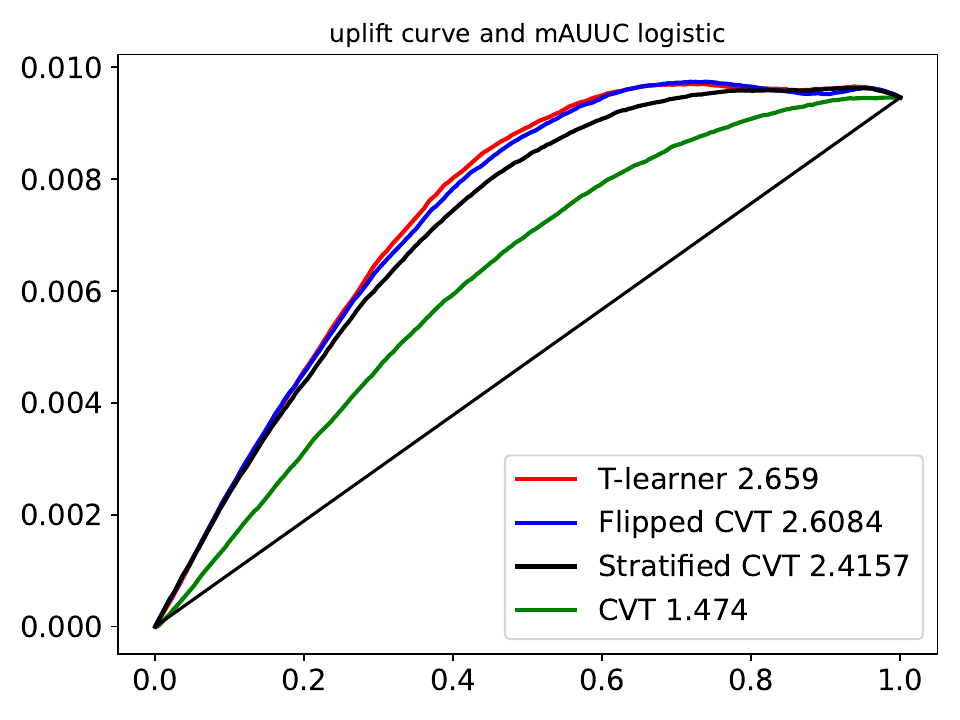}
\includegraphics[scale=\chartscale]{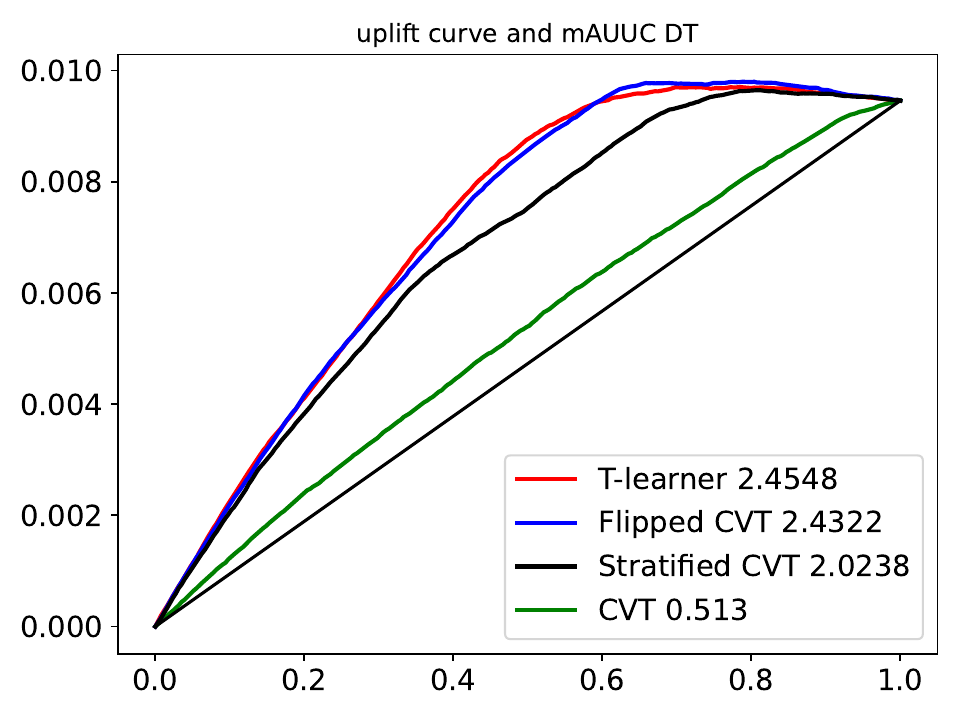}
\includegraphics[scale=\chartscale]{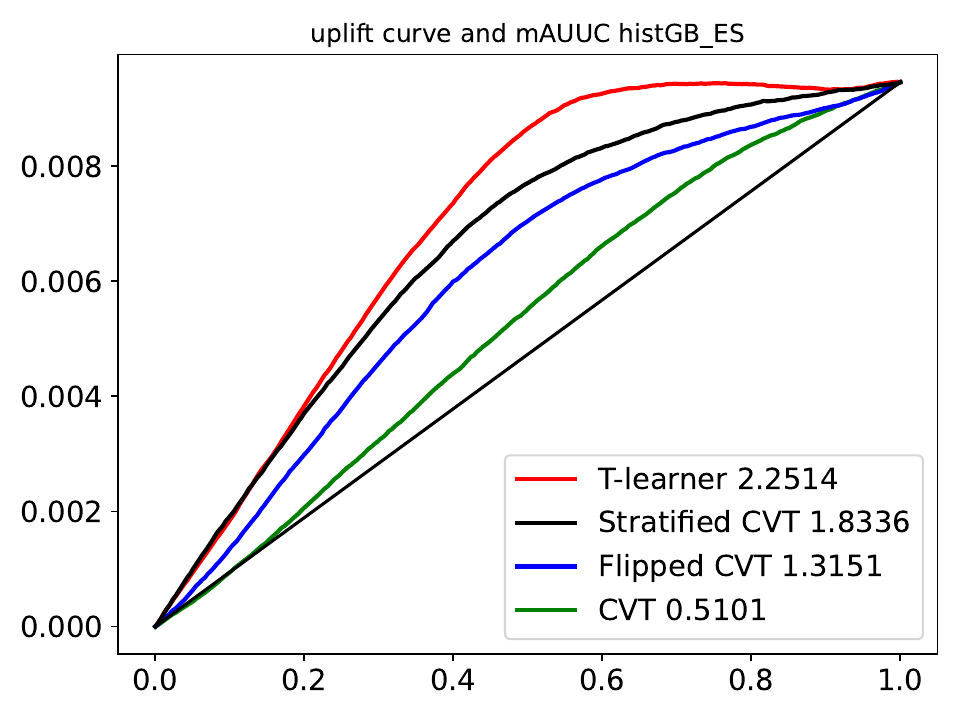}
\includegraphics[scale=\chartscale]{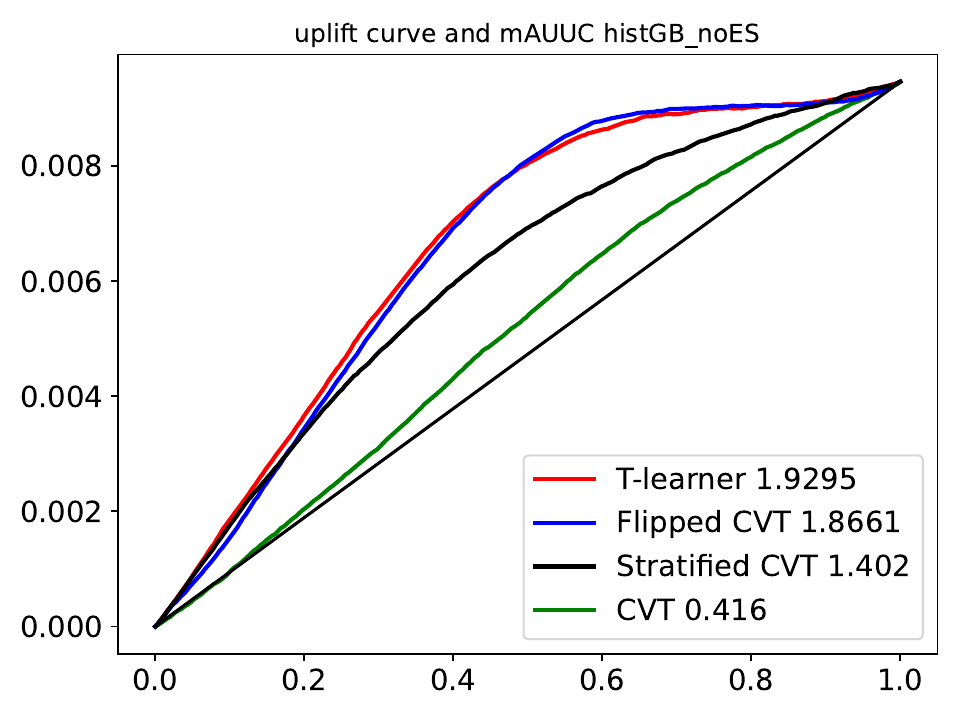}
\end{center}
\caption{Uplift curves and mAUUC for different CVT methods and T-learner on {\tt Starbucks} dataset for logistic regression, decision tree and gradient boosting used as base learners}\label{fig:starbucks} 
\end{figure}

\section{Experiments on standard classification problems}\label{sec:a-classif}

While the main topic of the paper is imbalance in the uplift
modeling/HTE estimation setting, the proposed class flipping method is
also applicable to standard classification
(Section~\ref{sec:imbal-classif}).  Summary of experimental results
with standard classification have been shown in
Section~\ref{sec:standard-classif}.  Here we provide additional
details about the experiments.

Both flipping and majority class undersampling were implemented by
reweighting the data records (see Equation~\ref{eq:under-weights} in
Section~\ref{sec:imbal-classif}), and the correction rate was chosen
to completely balance the classes.

We used 32 standard benchmark datasets and an artificial dataset
generated using {\tt scikit-learn} generator with class separation
parameter equal to $1$, making the problem moderately difficult.  We
additionally downsampled the minority class in some of the benchmark
datasets to make them more imbalanced.  Many of the datasets came from
the {\tt imbalanced-learn} package.  Table~\ref{tab:classif-datasets}
lists all the datasets together with the numbers of all records and
records in the minority class (after undersampling).  The
undersampling rate for the minority class used to increase imbalance
is shown in the last column.

\begin{table}
    \begin{tabular}{|l|c|c|c|}
      \hline
      Dataset & \# minority records & \# all records& sampling rate \\
      \hline        
      Artificial     & 20 & 1000 & --- \\
      Breast cancer  & 5 & 217 & 0.015 \\
      Diabetes       & 5 & 505  & 0.02 \\
      Oil            & 8 & 904  & 0.2 \\
      Mammography    & 26 & 10949 & 0.1 \\
      Abalone        & 6 & 695 & 0.15 \\
      Vowel          & 9 & 909 & 0.1 \\
      Vehicle        & 10 & 657 & 0.05 \\
      Ionosphere     & 10 & 235 & 0.08 \\
      Ecoli          & 35 & 336 & --- \\
      Optical\_digits  & 554 & 5620 & ---  \\
      Satimage       & 626 & 6435  & ---  \\
      Pen\_digits     & 1055 & 10992  & --- \\
      Sick\_euthyroid    & 293 & 3163 & ---  \\
      Spectrometer        & 45 & 531 & ---  \\
      Car\_eval\_34          & 134 & 1728 & ---  \\
      Isolet        & 600 & 7797 & ---  \\
      Us\_crime     & 150 & 1994 & ---  \\
      Yeast\_ml8     & 178 & 2417 & --- \\
      Scene  & 177 & 2407 & ---  \\
      Libras\_move       & 24 & 360  & ---  \\
      Thyroid\_sick            & 231 & 3772  & ---  \\
      Coil\_2000     & 586 & 9822 & ---  \\
      Arrhythmia        & 25 & 452 & ---  \\
      Solar\_flare\_m0          & 68 & 1389 & ---  \\
      Car\_eval\_4        & 65 & 1728 & ---  \\
      Wine\_quality    & 183 & 4898 & ---  \\
      Letter\_img            & 734 & 20000  & ---  \\
      Yeast\_me2   & 51 & 1484 & ---  \\
      Webpage        & 981 & 34780 & ---  \\
      Ozone\_level          & 73 & 2536 & ---  \\
      Protein\_homo       & 1296 & 145751 & ---  \\
      Abalone\_19       & 32 & 4177 & ---  \\
\hline
    \end{tabular}
  \caption{Datasets used for imbalanced classification experiment}\label{tab:classif-datasets}
\end{table}

All classifiers were evaluated using 5-fold stratified
cross-validation.  The procedure has been repeated 100 times for
stability. For each algorithm and imbalance correction method we give
average test set AUROC together with standard deviation.  Performance
without class imbalance correction is also included.  We used the
classifiers which were earlier used as base learners for metamodels,
see Section~\ref{sec:exper}.

Table~\ref{tab:classif-imbal} presents the results. See
Section~\ref{sec:standard-classif} for a summary and discussion.

\begin{table}
\scriptsize
\begin{tabular}{l|l|l|l|l}
Model                         & Method   & Artificial                      & Breast & Diabetes                                  \\ \hline
\multirow{3}{*}{Logistic}     & None     & 0.8459 $\pm$ 0.11          & 0.9969 $\pm$ 0.01          & 0.6690 $\pm$ 0.27                  \\
                              & Flipping & \textbf{0.8688 $\pm$ 0.10} & 0.9939 $\pm$ 0.01          & 0.6975 $\pm$ 0.26                  \\
                              & Under    & 0.8346 $\pm$ 0.11          & \textbf{0.9987 $\pm$ 0.01} & \textbf{0.7295 $\pm$ 0.20}   \\ \hline
\multirow{3}{*}{DT}           & None     & 0.8436 $\pm$ 0.10          & \textbf{0.8845 $\pm$ 0.20} & 0.7884 $\pm$ 0.25                    \\
                              & Flipping & 0.8515 $\pm$ 0.11          & 0.8785 $\pm$ 0.25          & \textbf{0.8041 $\pm$ 0.26}          \\
                              & Under    & \textbf{0.8569 $\pm$ 0.10} & 0.8145 $\pm$ 0.24          & 0.4945 $\pm$ 0.13           \\ \hline
\multirow{3}{*}{HistGB\_ES}   & None     & \textbf{0.8786 $\pm$ 0.10} & \textbf{0.9854 $\pm$ 0.03} & 0.7659 $\pm$ 0.23                    \\
                              & Flipping & 0.8338 $\pm$ 0.12          & 0.9310 $\pm$ 0.16           & 0.5989 $\pm$ 0.30                    \\
                              & Under    & 0.8449 $\pm$ 0.11          & 0.9847 $\pm$ 0.03          & \textbf{0.7737 $\pm$ 0.23}  \\ \hline
\multirow{3}{*}{HistGB\_noES} & None     & 0.8744 $\pm$ 0.11          & \textbf{0.9867 $\pm$ 0.03} & 0.7761 $\pm$ 0.20           \\
                              & Flipping & \textbf{0.8785 $\pm$ 0.11} & 0.9770 $\pm$ 0.09           & 0.6814 $\pm$ 0.33           \\
                              & Under    & 0.8632 $\pm$ 0.11          & 0.9864 $\pm$ 0.03          & \textbf{0.7823 $\pm$ 0.21}       
\end{tabular}

\vspace*{1 cm}

\begin{tabular}{l|l|l|l|l}
Model                         & Method   & Oil 			& Mammography 		 & Abalone                                   \\ \hline
\multirow{3}{*}{Logistic}     & None    & 0.9746 $\pm$ 0.04          & 0.9463 $\pm$ 0.06   & 0.5775 $\pm$ 0.23          \\
                              & Flipping & \textbf{0.9912 $\pm$ 0.01} & 0.9195 $\pm$ 0.08 & \textbf{0.7948 $\pm$ 0.11}  \\
                              & Under   & 0.9757 $\pm$ 0.04          & \textbf{0.9481 $\pm$ 0.06}  & 0.5714 $\pm$ 0.23          \\ \hline
\multirow{3}{*}{DT}           & None  & 0.8261 $\pm$ 0.19          & 0.8747 $\pm$ 0.10            & 0.5261 $\pm$ 0.20          \\
                              & Flipping & \textbf{0.8274 $\pm$ 0.19} & 0.8753 $\pm$ 0.10 & \textbf{0.5430 $\pm$ 0.27}           \\
                              & Under     & 0.7566 $\pm$ 0.21          & \textbf{0.9037 $\pm$ 0.09} & 0.4700 $\pm$ 0.02           \\ \hline
\multirow{3}{*}{HistGB\_ES}   & None     & \textbf{0.9307 $\pm$ 0.14} & 0.6587 $\pm$ 0.24          & 0.4768 $\pm$ 0.27         \\
                              & Flipping  & 0.7884 $\pm$ 0.22          & 0.7842 $\pm$ 0.15 & \textbf{0.5089 $\pm$ 0.28}          \\
                              & Under   & 0.8332 $\pm$ 0.23          & \textbf{0.8046 $\pm$ 0.13} & 0.4466 $\pm$ 0.24          \\ \hline
\multirow{3}{*}{HistGB\_noES} & None    & \textbf{0.9614 $\pm$ 0.07} & 0.6784 $\pm$ 0.19           & 0.5217 $\pm$ 0.23         \\
                              & Flipping  & 0.8570 $\pm$ 0.26           & \textbf{0.7938 $\pm$ 0.15} & \textbf{0.6422 $\pm$ 0.21} \\
                              & Under  & 0.9401 $\pm$ 0.11          & 0.7607 $\pm$ 0.12   & 0.5057 $\pm$ 0.21         
\end{tabular}

\vspace*{1 cm}

\begin{tabular}{l|l|l|l|l}
Model                         & Method                       & Vowel                      & Vehicle                    & Inosphere                  \\ \hline
\multirow{3}{*}{Logistic}     & None              & \textbf{0.9857 $\pm$ 0.02}  & 0.9731 $\pm$ 0.02          & 0.6741 $\pm$ 0.26          \\
                              & Flipping  & 0.9853 $\pm$ 0.01          & \textbf{0.9846 $\pm$ 0.02} & \textbf{0.7030 $\pm$ 0.30}  \\
                              & Under              & 0.9664 $\pm$ 0.02          & 0.9322 $\pm$ 0.05          & 0.6912 $\pm$ 0.26          \\ \hline
\multirow{3}{*}{DT}           & None               & \textbf{0.9639 $\pm$ 0.06} & 0.7130 $\pm$ 0.16           & 0.8101 $\pm$ 0.17          \\
                              & Flipping    & \textbf{0.9639 $\pm$ 0.06} & \textbf{0.7279 $\pm$ 0.19} & 0.7702 $\pm$ 0.18          \\
                              & Under               & 0.9343 $\pm$ 0.11          & 0.5796 $\pm$ 0.12           & \textbf{0.8425 $\pm$ 0.17} \\ \hline
\multirow{3}{*}{HistGB\_ES}   & None               & \textbf{0.9936 $\pm$ 0.02} & \textbf{0.8377 $\pm$ 0.14} & \textbf{0.8901 $\pm$ 0.16} \\
                              & Flipping  & 0.9014 $\pm$ 0.17          & 0.6664 $\pm$ 0.21          & 0.8396 $\pm$ 0.19          \\
                              & Under              & 0.9791 $\pm$ 0.07          & 0.7380 $\pm$ 0.21           & 0.8551 $\pm$ 0.17          \\ \hline
\multirow{3}{*}{HistGB\_noES} & None              & \textbf{0.9965 $\pm$ 0.01} & \textbf{0.8626 $\pm$ 0.12} & 0.8924 $\pm$ 0.16          \\
                              & Flipping  & 0.9919 $\pm$ 0.04          & 0.8242 $\pm$ 0.18          & \textbf{0.9168 $\pm$ 0.16} \\
                              & Under              & 0.9955 $\pm$ 0.01          & 0.8173 $\pm$ 0.17          & 0.8862 $\pm$ 0.16         
\end{tabular}

\vspace*{1 cm}

\begin{tabular}{l|l|l|l|l}
Model                         & Method                       & Ecoli                      & Optical\_digits                    & Satimage \\ \hline
\multirow{3}{*}{Logistic}     & None               & 0.9220 $\pm$ 0.06          & 0.8157 $\pm$ 0.06          & \textbf{0.7698 $\pm$ 0.01}          \\
                              & Flipping           & \textbf{0.9310 $\pm$ 0.05}          & \textbf{0.8542 $\pm$ 0.04}          & 0.7519 $\pm$ 0.02  \\
                              & Under              & 0.9182 $\pm$ 0.06          & 0.8294 $\pm$ 0.05          & 0.7646 $\pm$ 0.01          \\ \hline
\multirow{3}{*}{DT}           & None               & 0.8780 $\pm$ 0.08          & \textbf{0.7554 $\pm$ 0.09}          & 0.9065 $\pm$ 0.01          \\
                              & Flipping           & 0.8846 $\pm$ 0.08          & 0.7476 $\pm$ 0.10          & 0.9069 $\pm$ 0.01          \\
                              & Under              & \textbf{0.8978 $\pm$ 0.07}          & 0.7365 $\pm$ 0.09          & \textbf{0.9114 $\pm$ 0.01} \\ \hline
\multirow{3}{*}{HistGB\_ES}   & None               & \textbf{0.9299 $\pm$ 0.05}          & \textbf{0.6804 $\pm$ 0.12}          & \textbf{0.9604 $\pm$ 0.01} \\
                              & Flipping           & 0.8997 $\pm$ 0.08          & 0.5979 $\pm$ 0.14          & 0.9376 $\pm$ 0.01          \\
                              & Under              & 0.9244 $\pm$ 0.06          & 0.6487 $\pm$ 0.12          & 0.9561 $\pm$ 0.01          \\ \hline
\multirow{3}{*}{HistGB\_noES} & None               & \textbf{0.9237 $\pm$ 0.05}          & \textbf{0.7780 $\pm$ 0.11}          & \textbf{0.9644 $\pm$ 0.01}          \\
                              & Flipping           & 0.9207 $\pm$ 0.07          & 0.6441 $\pm$ 0.14          & 0.9529 $\pm$ 0.01 \\
                              & Under              & 0.9216 $\pm$ 0.05          & 0.6960 $\pm$ 0.11          & 0.9634 $\pm$ 0.01         
\end{tabular}
\end{table}

\begin{table}
\scriptsize
\begin{tabular}{l|l|l|l|l}
Model                         & Method           & pen\_digits                      & sick\_euthyroid                    & spectrometer\\ \hline
\multirow{3}{*}{Logistic}     & None               & 0.9764 $\pm$ 0.0044          & \textbf{0.9562 $\pm$ 0.02}          & 0.9490 $\pm$ 0.05          \\
                              & Flipping           & 0.9598 $\pm$ 0.0070           & 0.8830 $\pm$ 0.02          & \textbf{0.9855 $\pm$ 0.01}  \\
                              & Under              & \textbf{0.9767 $\pm$ 0.0035}          & 0.9529 $\pm$ 0.01          & 0.9570 $\pm$ 0.05     \\ \hline
\multirow{3}{*}{DT}           & None               & 0.9117 $\pm$ 0.0136          & 0.9634 $\pm$ 0.01          & 0.8913 $\pm$ 0.07          \\
                              & Flipping           & 0.9117 $\pm$ 0.0137          & \textbf{0.9635 $\pm$ 0.01}          & 0.8807 $\pm$ 0.08          \\
                              & Under              & \textbf{0.9563 $\pm$ 0.0076}          & 0.9610 $\pm$ 0.01          & \textbf{0.8933 $\pm$ 0.07} \\ \hline
\multirow{3}{*}{HistGB\_ES}   & None               & \textbf{0.9996 $\pm$ 0.0008}          & \textbf{0.9852 $\pm$ 0.01}          & \textbf{0.9745 $\pm$ 0.03} \\
                              & Flipping           & 0.9988 $\pm$ 0.0016          & 0.9710 $\pm$ 0.02          & 0.9381 $\pm$ 0.06          \\
                              & Under              & 0.9994 $\pm$ 0.0010           & 0.9850 $\pm$ 0.01          & 0.9648 $\pm$ 0.04  \\ \hline
\multirow{3}{*}{HistGB\_noES} & None               & \textbf{0.9998 $\pm$ 0.0003}          & 0.9856 $\pm$ 0.01          & \textbf{0.9797 $\pm$ 0.02}          \\
                              & Flipping           & 0.9997 $\pm$ 0.0005          & 0.9797 $\pm$ 0.01          & 0.9679 $\pm$ 0.04 \\
                              & Under              & \textbf{0.9998 $\pm$ 0.0003}          & \textbf{0.9858 $\pm$ 0.01}          & 0.9787 $\pm$ 0.03         
\end{tabular}

\vspace*{1 cm}

\begin{tabular}{l|l|l|l|l}
Model                         & Method           & car\_eval\_34                      & isolet                    & us\_crime\\ \hline
\multirow{3}{*}{Logistic}     & None               & \textbf{0.9980 $\pm$ 0.001}          & 0.9873 $\pm$ 0.004          & \textbf{0.9174 $\pm$ 0.02        }  \\
                              & Flipping           & 0.9891 $\pm$ 0.005           & 0.9859 $\pm$ 0.005          & 0.9093 $\pm$ 0.03  \\
                              & Under              & 0.9979 $\pm$ 0.001          & \textbf{0.9889 $\pm$ 0.004}          & 0.9152 $\pm$ 0.02     \\ \hline
\multirow{3}{*}{DT}           & None               & 0.9229 $\pm$ 0.020          & 0.9090 $\pm$ 0.010          & 0.8784 $\pm$ 0.03          \\
                              & Flipping           & 0.9229 $\pm$ 0.020          & 0.9102 $\pm$ 0.010          & 0.8779 $\pm$ 0.03          \\
                              & Under              & \textbf{0.9818 $\pm$ 0.010}          & \textbf{0.9349 $\pm$ 0.010}          & \textbf{0.8844 $\pm$ 0.03} \\ \hline
\multirow{3}{*}{HistGB\_ES}   & None               & \textbf{0.9995 $\pm$ 0.001}          & \textbf{0.9952 $\pm$ 0.002}          & \textbf{0.9141 $\pm$ 0.02} \\
                              & Flipping           & 0.9941 $\pm$ 0.007          & 0.9853 $\pm$ 0.006          & 0.8626 $\pm$ 0.04          \\
                              & Under              & 0.9993 $\pm$ 0.001           & 0.9946 $\pm$ 0.003          & 0.9049 $\pm$ 0.03  \\ \hline
\multirow{3}{*}{HistGB\_noES} & None               & \textbf{0.9997 $\pm$ 0.001}          & \textbf{0.9960 $\pm$ 0.002}          & \textbf{0.9178 $\pm$ 0.02}          \\
                              & Flipping           & 0.9983 $\pm$ 0.002          & 0.9937 $\pm$ 0.003          & 0.8881 $\pm$ 0.04 \\
                              & Under              & 0.9996 $\pm$ 0.001          & \textbf{0.9960 $\pm$ 0.002}          & 0.9165 $\pm$ 0.02         
\end{tabular}

\vspace*{1 cm}

\begin{tabular}{l|l|l|l|l}
Model                         & Method           & yeast\_ml8                      & scene                   & libras\_move \\ \hline
\multirow{3}{*}{Logistic}     & None               & 0.5646 $\pm$ 0.04          & 0.7085 $\pm$ 0.04          & 0.9425 $\pm$ 0.09          \\
                              & Flipping           & \textbf{0.5661 $\pm$ 0.04}           & \textbf{0.7161 $\pm$ 0.04}          & 0.9143 $\pm$ 0.12  \\
                              & Under              & 0.5592 $\pm$ 0.04          & 0.7145 $\pm$ 0.04          & \textbf{0.9489 $\pm$ 0.08}     \\ \hline
\multirow{3}{*}{DT}           & None               & 0.5455 $\pm$ 0.05         & 0.6952 $\pm$ 0.04          & 0.8372 $\pm$ 0.11          \\
                              & Flipping           & \textbf{0.5458 $\pm$ 0.05}          &0.6975 $\pm$ 0.04          & 0.8374 $\pm$ 0.11          \\
                              & Under              & 0.5445 $\pm$ 0.05          & \textbf{0.7004 $\pm$ 0.04}          & \textbf{0.8650 $\pm$ 0.10} \\ \hline
\multirow{3}{*}{HistGB\_ES}   & None               & \textbf{0.5697 $\pm$ 0.04}          & \textbf{0.7839 $\pm$ 0.04}          & \textbf{0.9469 $\pm$ 0.07} \\
                              & Flipping           & 0.5518 $\pm$ 0.05          & 0.7156 $\pm$ 0.05          & 0.8826 $\pm$ 0.11          \\
                              & Under              & 0.5620 $\pm$ 0.05           & 0.7551 $\pm$ 0.04          & 0.9375 $\pm$ 0.08  \\ \hline
\multirow{3}{*}{HistGB\_noES} & None               & \textbf{0.5878 $\pm$ 0.04}          & \textbf{0.8117 $\pm$ 0.03}          & 0.9569 $\pm$ 0.06          \\
                              & Flipping           & 0.5597 $\pm$ 0.05          & 0.7660 $\pm$ 0.04          & 0.9408 $\pm$ 0.08 \\
                              & Under              & 0.5832 $\pm$ 0.04         & 0.8060 $\pm$ 0.03          & \textbf{0.9600 $\pm$ 0.06}         
\end{tabular}

\vspace*{1 cm}

\begin{tabular}{l|l|l|l|l}
Model                         & Method           & thyroid\_sick                      & coil\_2000                   & arrhythmia \\ \hline
\multirow{3}{*}{Logistic}     & None               & \textbf{0.9488 $\pm$ 0.017}          & \textbf{0.7364 $\pm$ 0.02}          & \textbf{0.8940 $\pm$ 0.06}          \\
                              & Flipping           & 0.9060 $\pm$ 0.019           & 0.7357 $\pm$ 0.02          & 0.8478 $\pm$ 0.06  \\
                              & Under              & 0.9458 $\pm$ 0.016          & 0.7352 $\pm$ 0.02          & 0.8863 $\pm$ 0.07     \\ \hline
\multirow{3}{*}{DT}           & None               & 0.9731 $\pm$ 0.010         & 0.7294 $\pm$ 0.02          & 0.9391 $\pm$ 0.07          \\
                              & Flipping           & 0.9730 $\pm$ 0.011          &0.7294 $\pm$ 0.02          & 0.9396 $\pm$ 0.07          \\
                              & Under              & \textbf{0.9770 $\pm$ 0.010}          & \textbf{0.7317 $\pm$ 0.02}          & \textbf{0.9446 $\pm$ 0.06} \\ \hline
\multirow{3}{*}{HistGB\_ES}   & None               & \textbf{0.9962 $\pm$ 0.005}          & \textbf{0.7492 $\pm$ 0.02}          & \textbf{0.9696 $\pm$ 0.05} \\
                              & Flipping           & 0.9852 $\pm$ 0.014          & 0.7161 $\pm$ 0.02          & 0.9471 $\pm$ 0.07          \\
                              & Under              & 0.9941 $\pm$ 0.009           &  0.7460 $\pm$ 0.02          & 0.9653 $\pm$ 0.06  \\ \hline
\multirow{3}{*}{HistGB\_noES} & None               & \textbf{0.9976 $\pm$ 0.002}          & \textbf{0.7367 $\pm$ 0.02}          & \textbf{0.9776 $\pm$ 0.03}          \\
                              & Flipping           & 0.9916 $\pm$ 0.010          & 0.7214 $\pm$ 0.02          & 0.9526 $\pm$ 0.07 \\
                              & Under              & 0.9974 $\pm$ 0.003         & 0.7272 $\pm$ 0.02          & \textbf{0.9766 $\pm$ 0.03}         
\end{tabular}
\end{table}

\begin{table}
\scriptsize
\begin{tabular}{l|l|l|l|l}
Model                         & Method           & solar\_flare\_m0                      & car\_eval\_4                   & wine\_quality \\ \hline
\multirow{3}{*}{Logistic}     & None               & 0.7709 $\pm$ 0.06          & \textbf{0.9981 $\pm$ 0.0016}          & 0.7827 $\pm$ 0.04          \\
                              & Flipping           & \textbf{0.7746 $\pm$ 0.06}           & 0.9954 $\pm$ 0.0036         & \textbf{0.7840 $\pm$ 0.04}  \\
                              & Under              & 0.7666 $\pm$ 0.06          & \textbf{0.9981 $\pm$ 0.0015}          & 0.7810 $\pm$ 0.04     \\ \hline
\multirow{3}{*}{DT}           & None               & \textbf{0.7518 $\pm$ 0.06}         & 0.9411 $\pm$ 0.0134          & 0.7953 $\pm$ 0.04          \\
                              & Flipping           & 0.7513 $\pm$ 0.06          &0.9411 $\pm$ 0.0134         & \textbf{0.7954 $\pm$ 0.04}          \\
                              & Under              & 0.7469 $\pm$ 0.07          & \textbf{0.9720 $\pm$ 0.0098}         & 0.7850 $\pm$ 0.04 \\ \hline
\multirow{3}{*}{HistGB\_ES}   & None               & \textbf{0.7700 $\pm$ 0.06}          & \textbf{1.0 $\pm$ 0.0002}         & \textbf{0.8592 $\pm$ 0.03} \\
                              & Flipping           & 0.7085 $\pm$ 0.08          & 0.9961 $\pm$ 0.0107          & 0.7853 $\pm$ 0.05          \\
                              & Under              & 0.7543 $\pm$ 0.06           &  \textbf{1.0 $\pm$ 0.0002}          & 0.8523 $\pm$ 0.03  \\ \hline
\multirow{3}{*}{HistGB\_noES} & None               & \textbf{0.7556 $\pm$ 0.06}          & \textbf{1.0 $\pm$ 0.0001}          & \textbf{0.8592 $\pm$ 0.03}          \\
                              & Flipping           & 0.7099 $\pm$ 0.08          & 0.9998 $\pm$ 0.0008        & 0.8204 $\pm$ 0.05 \\
                              & Under              & 0.7446 $\pm$ 0.06         & \textbf{1.0 $\pm$ 0.0001}          & 0.8537 $\pm$ 0.03         
\end{tabular}

\vspace*{1 cm}

\begin{tabular}{l|l|l|l|l}
Model                         & Method           & letter\_img                      & yeast\_me2                   & webpage \\ \hline
\multirow{3}{*}{Logistic}     & None               & 0.9867 $\pm$ 0.003          & 0.8719 $\pm$ 0.06          & 0.9726 $\pm$ 0.006         \\
                              & Flipping           & 0.9725 $\pm$ 0.004           & \textbf{0.8780 $\pm$ 0.06}         & 0.9623 $\pm$ 0.009  \\
                              & Under              & \textbf{0.9880 $\pm$ 0.002}          & 0.8742 $\pm$ 0.06          & \textbf{0.9768  $\pm$ 0.005}     \\ \hline
\multirow{3}{*}{DT}           & None               & 0.9452 $\pm$ 0.005         & 0.8924 $\pm$ 0.05         & 0.7904 $\pm$ 0.013          \\
                              & Flipping           & 0.9452 $\pm$ 0.005          &0.8897 $\pm$ 0.06        & 0.7904 $\pm$ 0.013          \\
                              & Under              & \textbf{0.9725 $\pm$ 0.004}          & \textbf{0.9090 $\pm$ 0.04}        & \textbf{0.7968 $\pm$ 0.017} \\ \hline
\multirow{3}{*}{HistGB\_ES}   & None               & \textbf{0.9997 $\pm$ 0.001}          & \textbf{0.9197 $\pm$ 0.04}         & \textbf{0.9788 $\pm$ 0.005} \\
                              & Flipping           & 0.9954 $\pm$ 0.004          & 0.8287 $\pm$ 0.09          & 0.9499 $\pm$ 0.011          \\
                              & Under              & 0.9995 $\pm$ 0.001           &  0.8878 $\pm$ 0.07          & 0.9768 $\pm$ 0.005  \\ \hline
\multirow{3}{*}{HistGB\_noES} & None               & \textbf{0.9998 $\pm$ 0.001}         & \textbf{0.9144 $\pm$ 0.04}          & \textbf{0.9800 $\pm$ 0.005}          \\
                              & Flipping           & 0.9989 $\pm$ 0.002          & 0.8509 $\pm$ 0.08        & 0.9700 $\pm$ 0.008 \\
                              & Under              & \textbf{0.9998 $\pm$ 0.001}         & 0.8995 $\pm$ 0.05          & 0.9786 $\pm$ 0.005         
\end{tabular}

\vspace*{1 cm}

\begin{tabular}{l|l|l|l|l}
Model                         & Method           & ozone\_level                      & protein\_homo                   & abalone\_19 \\ \hline
\multirow{3}{*}{Logistic}     & None               & 0.8924 $\pm$ 0.04          & 0.9910 $\pm$ 0.002          & 0.8131 $\pm$ 0.06         \\
                              & Flipping           & 0.8921 $\pm$ 0.04           & 0.9716 $\pm$ 0.005         & \textbf{0.8517 $\pm$ 0.04}  \\
                              & Under              & \textbf{0.8934 $\pm$ 0.04}          & \textbf{0.9922 $\pm$ 0.002}          & 0.8303  $\pm$ 0.06     \\ \hline
\multirow{3}{*}{DT}           & None               & 0.7832 $\pm$ 0.06         & 0.9466 $\pm$ 0.006         & \textbf{0.7351 $\pm$ 0.10}          \\
                              & Flipping           & 0.7804 $\pm$ 0.07          &0.9463 $\pm$ 0.006        & 0.7248 $\pm$ 0.11          \\
                              & Under              & \textbf{0.8215 $\pm$ 0.06}          & \textbf{0.9605 $\pm$ 0.006}        & 0.7208 $\pm$ 0.10 \\ \hline
\multirow{3}{*}{HistGB\_ES}   & None               & \textbf{0.8701 $\pm$ 0.05}          & 0.9921 $\pm$ 0.003         & \textbf{0.7148 $\pm$ 0.11} \\
                              & Flipping           & 0.7898  $\pm$ 0.08          & 0.9669 $\pm$ 0.010          & 0.5878 $\pm$ 0.13          \\
                              & Under              & 0.8362 $\pm$ 0.07           &  \textbf{0.9924 $\pm$ 0.002}          & 0.6437 $\pm$ 0.14  \\ \hline
\multirow{3}{*}{HistGB\_noES} & None               & \textbf{0.9002 $\pm$ 0.04}         & 0.9935 $\pm$ 0.003          & \textbf{0.7748 $\pm$ 0.08}          \\
                              & Flipping           & 0.8496 $\pm$ 0.06          & 0.9800 $\pm$ 0.006        & 0.6327 $\pm$ 0.14 \\
                              & Under              & 0.8985 $\pm$ 0.04         & \textbf{0.9942 $\pm$ 0.002}          & 0.7124 $\pm$ 0.11         
\end{tabular}

  \caption{Effects of different class imbalance corrections on classifier
    performance (Area Under ROC curve).  The Method column gives the method used to correct
    class imbalance: undersampling (Under), flipping or none.}
  \label{tab:classif-imbal}
\end{table}

\end{appendices}

\end{document}